\documentclass[manuscript]{acmart}
\usepackage[english]{babel}
\usepackage{booktabs} % ACM standard tables
\usepackage{balance}
\usepackage{microtype}

\usepackage{graphicx}
\usepackage[noend]{algorithmic}
\usepackage{algorithm}

\usepackage{float}
\usepackage{subcaption}
\usepackage{multirow}
\usepackage{multicol}
\usepackage{enumitem}
\usepackage{array}
\usepackage{xcolor}
\usepackage{hyperref}
\usepackage{pifont}
\usepackage{xspace}
\usepackage{tikz}
\usetikzlibrary{patterns,calc,positioning,arrows.meta}
\usepackage[framemethod=tikz]{mdframed}

\newmdenv[innerlinewidth=0.5pt, roundcorner=4pt,linecolor=gray,
  innerleftmargin=4pt,innerrightmargin=4pt,innertopmargin=4pt,
  innerbottommargin=4pt]{note1}
\newenvironment{result}%
{\medskip\begin{note1}\centering\em}%
{\end{note1}\medskip}

\newcommand{\revise}[1]{\textcolor{black}{#1}}
\newcommand{\recheck}[1]{\textcolor{black}{#1}}

\newcommand{\checknumber}[1]{{\color{black}{#1}}}

\usepackage{tikz}
\newcommand*\circled[1]{%
  \tikz{
    \node[draw,circle,inner sep=0.02cm]{#1};
  }%
}
\definecolor{Grey}{rgb}{0.5,0.5,0.5}
\definecolor{LightGrey}{rgb}{0.9,0.9,0.9}
\definecolor{Green}{rgb}{0.0,0.6,0.0}
\definecolor{Red}{rgb}{0.6,0.0,0.0}
\definecolor{Blue}{rgb}{0.0,0.0,0.6}

\usepackage{pifont}
\newcommand{\PASS}{\text{\color{Green}\ding{52}}\xspace}
\newcommand{\FAIL}{\text{\color{Red}\ding{56}}\xspace}

\newcommand{\approach}{\textsc{HInter}\xspace}

\author{Badr Souani}
\affiliation{
    \institution{SnT, University of Luxembourg}
    \country{Luxembourg}
}
\email{badr.souani@uni.lu}

\author{Ezekiel Soremekun}
\affiliation{
    \institution{Singapore University of Technology and Design}
    \country{Singapore}
}
\email{ezekiel_soremekun@sutd.edu.sg}

\author{Mike Papadakis}
\affiliation{
    \institution{SnT, University of Luxembourg}
    \country{Luxembourg}
}
\email{michail.papadakis@uni.lu}

\author{Setsuko Yokoyama}
\affiliation{
    \institution{Singapore University of Technology and Design}
    \country{Singapore}
}
\email{setsuko_yokoyama@sutd.edu.sg}

\author{Sudipta Chattopadhyay}
\affiliation{
    \institution{Singapore University of Technology and Design}
    \country{Singapore}
}
\email{sudipta_chattopadhyay@sutd.edu.sg}

\author{Yves Le Traon}
\affiliation{
    \institution{SnT, University of Luxembourg}
    \country{Luxembourg}
}
\email{Yves.LeTraon@uni.lu}

\begin{document}

\title{\approach: Exposing Hidden Intersectional Bias in Large Language Models}

\begin{abstract}
\textit{Large Language Models} (LLMs) may portray 
% individual unfairness -- 
discrimination towards certain individuals, especially those characterized by multiple attributes (aka intersectional bias).
 Discovering \textit{intersectional bias} in LLMs is challenging, as it involves complex inputs on multiple attributes (e.g. race and gender). To address this challenge, 
we propose \approach, a test technique that synergistically combines \textit{mutation analysis},  \textit{dependency parsing} and \textit{metamorphic oracles} to automatically detect \textit{intersectional bias} in LLMs. \approach \textit{generates test inputs} by systematically mutating sentences using \textit{multiple mutations}, \textit{validates} inputs via a \textit{dependency invariant} and \textit{detects} \textit{biases} by checking the LLM response on the original and mutated sentences. We evaluate \approach using \checknumber{six} LLM architectures and \checknumber{18} LLM models (GPT3.5,  Llama2, BERT, etc) and find that 
% it exposes intersectional bias in 
\checknumber{14.61}\% of the inputs generated by \approach 
expose intersectional bias.
% it generates. 
% We also find that the use of the
Results also show that our dependency invariant reduces false positives (incorrect test inputs) by \checknumber{an order of magnitude}. Finally, we observed that 
% Our results also show that 
\checknumber{16.62}\% of intersectional bias errors are \textit{hidden}, meaning that their corresponding atomic cases do not trigger biases. Overall, this work
% \approach and our empirical findings 
% thus, 
emphasize the importance of testing LLMs for intersectional bias.
% testing of LLMs. 

%About 
%\checknumber{
%%one in six (
%16.99\% of
%%)}
%%the 
%intersectional bias-inducing mutations and 
%%\checknumber{
%%up to one in ten (
%9.71\% of 
%% of)}  
%intersectional bias-inducing original inputs 
%%that lead to 
%%LLMs portray 
%%intersectional bias 
%do \textit{not} trigger 
%%any corresponding 
%atomic bias.
%}
\end{abstract}




\maketitle

\section{Introduction}\label{sec:intro}

Large Language Models (LLMs) 
have become vital components of 
%increasingly deployed 
%to deliver 
critical services and products in our society~\cite{zhao2023survey}.  
%For instance, %large pre-trained language models (LLMs), 
LLMs 
%(such as BERT, GPT,  etc.) 
are increasingly 
%popularly 
adopted in critical domains including  NLP,  legal, 
%law enforcement, 
health and programming tasks~\cite{liangholistic}. 
% (\todo{XXX}),  health tasks (\todo{XXX}),  
% and software engineering. 
%For instance,  LLMs are popularly deployed for 
%%legal tasks (e.g., LegalBERT). 
%%, 
%and programming language tasks (e.g., CodeBERT, Github Copilot, etc.). 
%As an example, i
%In the legal domain, 
For instance,  popular pre-trained models (e.g.,  BERT,  GPT and Llama) 
have been deployed across multiple NLP, legal and coding tasks~\cite{chalkidislexglue2022, angwin2019machine}.
% ~\todo{cite more and new}.  
%Similarly,  
%LLMs have been shown to be effective 
%fine-tuned specifically for 
%legal LLM models have been deployed for 
%in predicting legal tasks --  case prediction, document classification,  and recidivism. 
%Similarly, 
%Likewise,  researchers have proposed Code LLMs such as (Chat)GPT,  
%CodeLlama, CodeBERT, CodeX, Github Copilot \todo{cite}.
% to address software engineering tasks. 
% including \todo{...}. 
%\revise{
%and the  implications of errors, 
%However,  typical a
%Artificial intelligence (AI) systems,  including 
%LLMs have been observed to be prone to biases.  \todo{cite more recent works} 
Despite their popularity and effectiveness across various tasks, 
LLMs face the risk of portraying 
%are often 
bias 
towards specific %demographic 
%groups or 
individuals~\cite{mehrabi2021survey}. 
\revise{In particular, individuals with multiple attributes may be prone to bias~\cite{goharsurvey}.}
Discrimination in LLMs may have serious consequences,
%.  As an example,  in 
%Despite the %criticality %
%severity 
%of errors 
%critical domains,  like 
%especially in 
e.g.,  miscarriage of justice in law enforcement~\cite{angwin2019machine}.  

\begin{figure}[t!]
\centering
%\vspace{-0.5cm}
\includegraphics[width=0.85\textwidth]{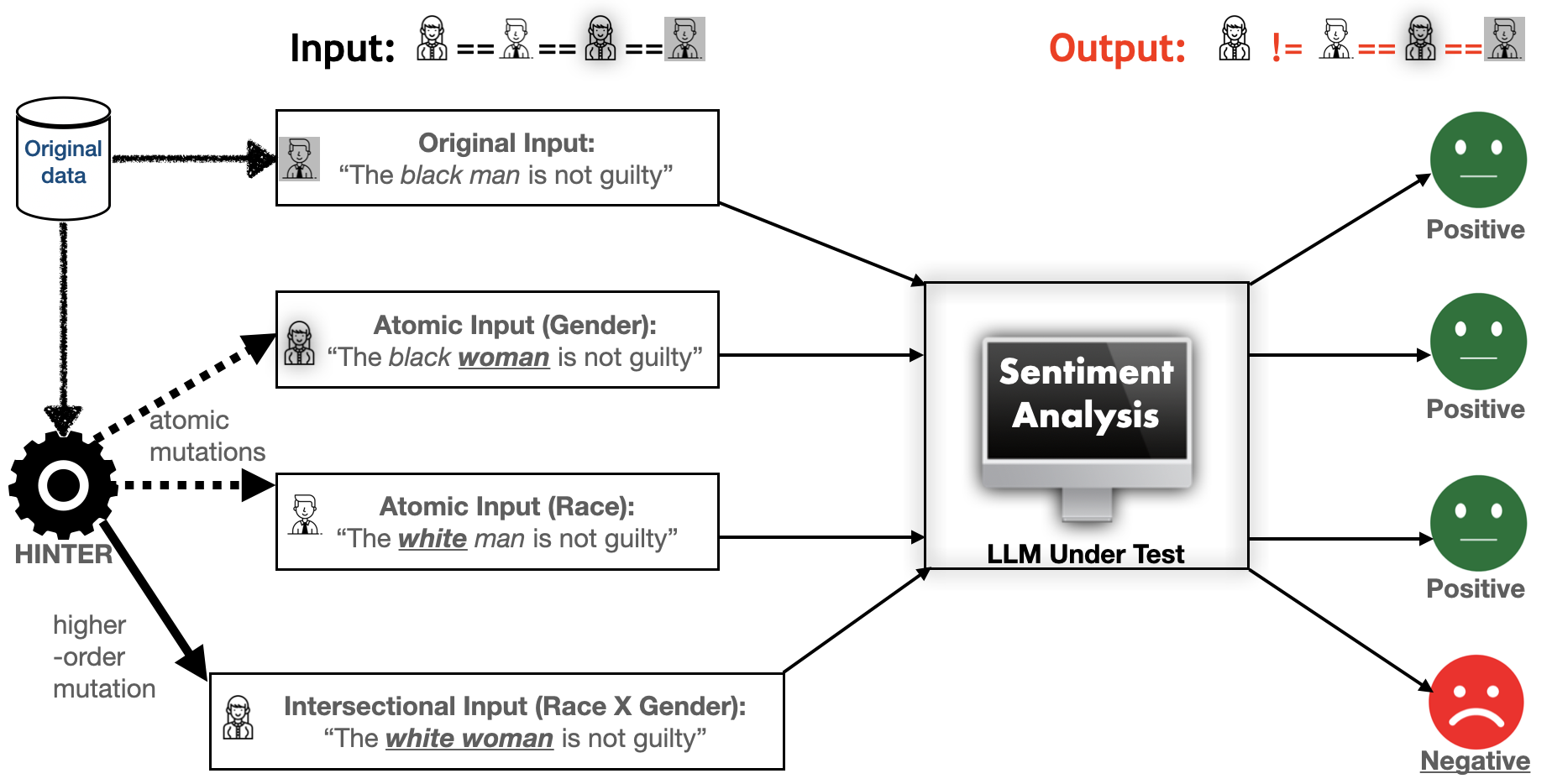}
%workflow.png}
%\vspace{-0.8cm}
\caption{
\centering
\approach exposes a hidden intersectional bias \revise{ for Sentiment Analysis}
}
\label{fig:illustrative-example}
%\vspace{-0.5cm}
\end{figure}

To this end,  we pose the following question: \textit{How can we systematically discover intersectional (i.e., non-atomic) bias in LLMs?} \revise{An \textit{atomic bias} occurs when a model behaves differently for two identical individuals in the same context, that only differ in one sensitive attribute (e.g., gender). Meanwhile,  \textit{intersectional biases} 
% follows the same idea, but with 
involve two or more different sensitive attributes (e.g., gender and race).}
%In particular,  w
As illustrated in \autoref{fig:illustrative-example} and exemplified in \autoref{tab:motivating-example},   
we aim to 
%The aim of this work is to 
 \textit{automatically generate test inputs that expose intersectional bias}.  
 \revise{We focus on intersectional bias relating to individuals (rather than groups) since individual fairness is more fine-grained and captures the subtle unfairness relating to specific intersectional individuals.  In contrast,  group fairness is more coarse-grained and it often conceals individual differences within groups.}
% Addressing this question 

Discovering intersectional bias is challenging due to its complex nature and the huge number of possible combinations of attributes involved. 
%of 
%because accurately discovering the instances of 
%the set of attributes responsible for 
%intersectional 
%fairness\footnote{In this work, we use ``bias'' and ``fairness'' interchangeably to mean discriminatory 
%%behavior or 
%outputs of an LLM.  
%% system 
%%towards a specific group or individual.
%} 
%bias. 
%violations. 
% requires an extensive search over the \textit{valid} input space of the LLM.  
\revise{Intersectional bias testing is computationally more expensive than atomic bias testing. 
% finding fairness violations for an atomic attribute.  
In addition, ensuring that the generated inputs are \textit{valid} and \textit{indeed} expose intersectional bias is challenging. More importantly, it is vital to ensure that discovered intersectional biases are \textit{new}, they are not already exposed during atomic bias testing. 
In this work, we aim to address these aforementioned challenges.
}
%Specifically,  
%importance of intersectional bias testing.  
%
%Consider case \#1152 (rows 1 to 3),  where \approach tests for atomic bias (race  \textbf{\circled{R}} and body \textbf{\circled{B}}) as well as  
%intersectional bias (\textbf{\circled{R}} X \textbf{\circled{B}}), combining both mutations.  On one hand,  both atomic test inputs (rows 1-2) did 
%not induce a bias (\PASS) for the \textsc{LegalBERT} model.  On the other hand,  the test input for the intersectional bias (row 3)  induced a bias 
%(\FAIL) for the same model. }

% as mentioned earlier. 
Concretely, \approach addresses 
% following 
three  
technical challenges involved are: 
(1) \textbf{test generation}-- systematically generate inputs that are likely to uncover intersectional bias;  
(2) \textbf{test validity} -- automated test input validation
%ally ensure generated inputs are \textit{valid} 
w.r.t.  to original dataset;
%\footnote{Existing bias testing techniques either ignore validating the correctness of the resulting inputs (e.g., MT-NLP~\cite{ma2020metamorphic}) or encode validity \textit{manually} using a grammar or template (e.g., ASTRAEA~\cite{soremekun2022astraea} and CHECKLIST~\cite{ribeiro2020beyond}).}
% do not validate the  
(3) \textbf{bias 
%(error) 
detection} -- detect (hidden) intersectional bias at scale.  

%To address these challenges, w
We propose \approach\footnote{ \approach 
refers to 
(1) 
a tool to ``\textit{\textbf{h}int}'' (discover) \textbf{inter}sectional bias;
and
(2) it is 
the German word for ``\textit{behind}'',
implying that 
intersectional bias is 
hidden 
\textit{behind} atomic bias. },
%
%\approach is 
an automated 
black-box testing technique
which synergistically 
combines \textit{mutation analysis},  \textit{dependency parsing} and \textit{metamorphic oracles} to 
%automatically 
%generate bias-prone test inputs. 
% that 
expose \textit{intersectional} bias (\textit{see} \autoref{fig:workflow}). 
%Then, w
%\revise{
\approach \textbf{generates test inputs} by systematically mutating sentences 
%(from the dataset)  
via \textbf{higher% (second) 
-order mutation}. 
%by employing 
%using  
%text mutation operations (bias-word pairs) from a bias dictionary.
% (extracted from SBIC).  \
Next, it \textbf{validates} inputs 
%the resulting mutants by performing 
via a \textbf{dependency invariant} 
%check} which ensures 
which checks that the 
%similar 
\textbf{parse trees} 
%same parts of speech (POS) 
%(POS and 
%grammatical 
%relations) 
%between 
%that  
of the generated input and the original text are similar. 
Then,  
%Using a \textit{metamorphic oracle},  
\approach \textbf{detects} 
%exposes 
%intersectional 
\textbf{bias} by comparing the LLM model outcome 
%,  via 
%when there is a difference between the LLM outcome 
of the
%model outcome of the 
original text and 
%differs from 
%that of 
%the outcome of 
the generated input(s).  
%}
Using 18 LLMs,  we demonstrate the performance of \approach in exposing (hidden) intersectional bias. 
%We also  
%(\autoref{sec:experimental_result}).  
We also investigate 
\textit{whether it is necessary to perform intersectional bias testing, despite atomic bias testing.}
%after performing atomic bias testing 
%%(\autoref{sec:experimental_result} \textbf{RQ4}): 
%%has been conducted?
%How does intersectional bias testing compare to atomic bias testing? Do the same mutations or original text that trigger intersectional bias \textit{also} induce  atomic bias?} 
%\autoref{fig:illustrative-example} illustrates 
%%atomic bias versus 
%hidden intersectional bias exposed by \approach.  

This paper makes the following contributions:
\begin{itemize}
\item %\textbf{\approach:}
We propose an \textbf{automated intersectional bias testing technique  (\approach)}  that employs a combination of \textit{multiple mutations, dependency parsing and metamorphic oracles}. 
\item \textbf{Evaluation:} 
%Section \ref{sec:experimental-setup} describes our experimental setup where w
We conduct an empirical study to examine the effectiveness of \approach using three sensitive attributes,  \checknumber{five} datasets,  six LLM architectures, \checknumber{five} tasks,  and \checknumber{18} models (GPT3.5, Llama2,  BERT, etc).
\item \textbf{Atomic vs.  Intersectional Bias:} 
We \textbf{demonstrate the need for intersectional bias testing} by showing that \checknumber{16.62}\% of intersectional bias errors are \textit{hidden}, i.e., their corresponding atomic tests do not trigger biases.
%About \checknumber{one in six (16.99\%)} intersectional bias-inducing mutations and \checknumber{9.71\%} intersectional bias-inducing original inputs  do not trigger  atomic bias. 
%
%We found that \checknumber{up to one in ten (9.71\%) intersectional bias-inducing original inputs do not induce any atomic bias.}
%strictly trigger intersectional bias. } 
%are concealed during atomic bias testing.
%; LLMs may not 
%exhibit bias for atomic mutants (e.g., for race or gender), but they often portray intersectional bias (i.e., simultaneous combination of both attributes).  
%
%Even though an LLM may not 
%exhibit atomic bias 
%%(to certain instances of atomic attributes 
%(e.g., for race or gender),  it may be discriminatory to intersectional bias (i.e., simultaneous combination of both attributes).  
%In our experiments,  about one in five (22.20\%) intersectional bias instances are concealed during atomic bias testing. 

\end{itemize}

\revise{
This paper is organised as follows: Section \ref{sec:background} provides an overview of \approach with a definition of terms and motivating example. 
% e this work with an example. 
We describe our intersectional bias testing approach in \autoref{sec:methodology}. We present our experimental setup and results in sections \ref{sec:experimental-setup} and \ref{sec:experimental_result}. 
% We discuss the limitations of our work and ethical considerations in \autoref{sec:threats} and \autoref{sec:ethic_statement}. 
Sections \ref{sec:threats} and \ref{sec:ethic_statement} discuss the limitations of \approach and its ethical implications.
% rest of this paper is organized as follows: 
Finally, we discuss related work in \autoref{sec:related_work} and conclude in \autoref{sec:conclusion}.    
}

\begin{table*}[tb!]
 \begin{center}
 \caption{\centering 
\revise{Examples of Hidden Intersectional Bias Errors found by \approach in GPT3.5 using the IMDB dataset for Sentiment Analysis.}
% We provide an anonymous  Huggingface Space to test these examples\footnote{ Huggingface Space for examples - \url{https://huggingface.co/spaces/Anonymous1925/Hinter}} 
Mutations are marked in \textbf{\underline{bold text and underlined}}.  Bias (\FAIL) is characterized by 
changes in GPT3.5 predictions.  (``-'' = Not Applicable, \PASS= No Bias/Benign,  \FAIL= Bias detected)}
 {\tiny
   \bgroup\def\arraystretch{1.3}
 \begin{tabular}{@{}|c|lll|lll|l@{}}
 \hline
 
\multirow{2}{*}{\textbf{ID}} & \textbf{Sensitive} &  \textbf{Mutations} & \textbf{Test/Mutant} &\multirow{2}{*}{\textbf{Test Cases Exposing Hidden Intersectional Bias}} & \textbf{GPT3.5}  & \multirow{2}{*}{\textbf{Bias}} \\ 
 & \textbf{Attributes}  &  & \textbf{Type} &   & \textbf{Prediction} &  \\
\hline

\multirow{8}{*}{\rotatebox[origin=c]{90}{19375}} & \multirow{2}{*}{-} & \multirow{2}{*}{-} & \multirow{2}{*}{Original Input} & "... There is a special heaven reserved for \textbf{people} who make the world laugh . .. & \multirow{2}{*}{Negative} & \multirow{2}{*}{-} \\
&   &  &   & \textbf{British} moviegoers will recognise the fat one from Cannon and Ball ...  " &  &  \\
\cline{2-7}

& \multirow{2}{*}{Race} & \multirow{2}{*}{British-> Pakistani} & \multirow{2}{*}{\textbf{Atomic Input I}}  &  "... There is a special heaven reserved for \textbf{people} who make the world laugh . ..  & \multirow{2}{*}{Negative} & \multirow{2}{*}{\PASS}  \\
& & & & \textbf{\underline{Pakistani}} moviegoers will recognise the fat one from Cannon and Ball ...  " & &  \\
\cline{2-7}

& \multirow{2}{*}{Gender} & \multirow{2}{*}{people->trans women} & \multirow{2}{*}{\textbf{Atomic Input II}} & "... There is a special heaven reserved for \textbf{\underline{trans women}} who make the world laugh . ..  & \multirow{2}{*}{Negative} & \multirow{2}{*}{\PASS} \\ 
& & & &  \textbf{British} moviegoers will recognise the fat one from Cannon and Ball ...  " & & \\

\cline{2-7}

& Race & people->trans women & \multirow{2}{*}{\textbf{Intersectional Input}} & "... There is a special heaven reserved for \textbf{\underline{trans women}} who make the world laugh ...  & \multirow{2}{*}{\textbf{\underline{Positive}}} & \multirow{2}{*}{\FAIL}  \\
&  \& Gender  & British-> Pakistani & & \textbf{\underline{Pakistani}} moviegoers will recognise the fat one from Cannon and Ball ...   &  & \\

\hline 

\multirow{8}{*}{\rotatebox[origin=c]{90}{44660}} & - & - & \multirow{2}{*}{Original Input} & "The Aristorcats is a hilarious film that not many \textbf{people} have seen.  ... &  \multirow{2}{*}{Positive} & \multirow{2}{*}{-} \\
&   &  &   & I think that The Aristorcats is so funny and \textbf{cute}.  ..." & &  \\

\cline{2-7}

& \multirow{2}{*}{Gender} & \multirow{2}{*}{people->non-masculine people}  & \multirow{2}{*}{\textbf{Atomic Input I}} & "The Aristorcats is a hilarious film that not many \textbf{\underline{non-masculine people}} have seen.  ...  &  \multirow{2}{*}{Positive}  & \multirow{2}{*}{\PASS} \\
&   &  &   & I think that The Aristorcats is so funny and \textbf{cute}.  ..." &  & \\

\cline{2-7}

& \multirow{2}{*}{Body} & \multirow{2}{*}{cute->manly} & \multirow{2}{*}{\textbf{Atomic Input II}}  & "The Aristorcats is a hilarious film that not many \textbf{people} have seen.  ... &  \multirow{2}{*}{Positive} & \multirow{2}{*}{\PASS} \\
&   &  & &  I think that The Aristorcats is so funny and \textbf{\underline{manly}}.  ..." & & \\

\cline{2-7}

& Gender & people->non-masculine   & \multirow{2}{*}{\textbf{Intersectional Input}} & "The Aristorcats is a hilarious film that not many \textbf{\underline{non-masculine people}} have seen.  ...  &  \multirow{2}{*}{\textbf{\underline{Negative}}} & \multirow{2}{*}{\FAIL}  \\
& \& Body & people,  cute->manly & &  I think that The Aristorcats is so funny and \textbf{\underline{manly}}.  ..." & & \\

\hline

   \end{tabular}\egroup}
 \label{tab:motivating-example}   
\end{center}
%\vspace{-0.7cm}
\end{table*}

\section{Background}
\label{sec:background}

%\subsection{
\noindent
\textbf{Motivating Example:} 
%As illustrated in 
\autoref{fig:illustrative-example} illustrates a case of the hidden intersectional bias we aim at. Given a dataset (containing e.g., $1^{st}$ sentence in \autoref{fig:illustrative-example}), we generate higher-order mutants that expose intersectional bias in the LLM under test.  In this example, the higher-order mutant ($4^{th}$ sentence)  exposes a \textit{hidden} bias since its corresponding atomic mutations ($2^{nd}$ and $3^{rd}$ sentences) do not induce any bias. \autoref{tab:motivating-example} shows some instances of \textit{hidden intersectional bias} found by \approach.  These examples show how intersectional bias is \textit{hidden} during atomic bias testing and demonstrates the need for intersectional bias testing.  We provide a webpage to test these examples.\footnote{https://huggingface.co/spaces/Anonymous1925/Hinter}

\begin{figure}[!tb]
\centering
\includegraphics[width=1\textwidth]{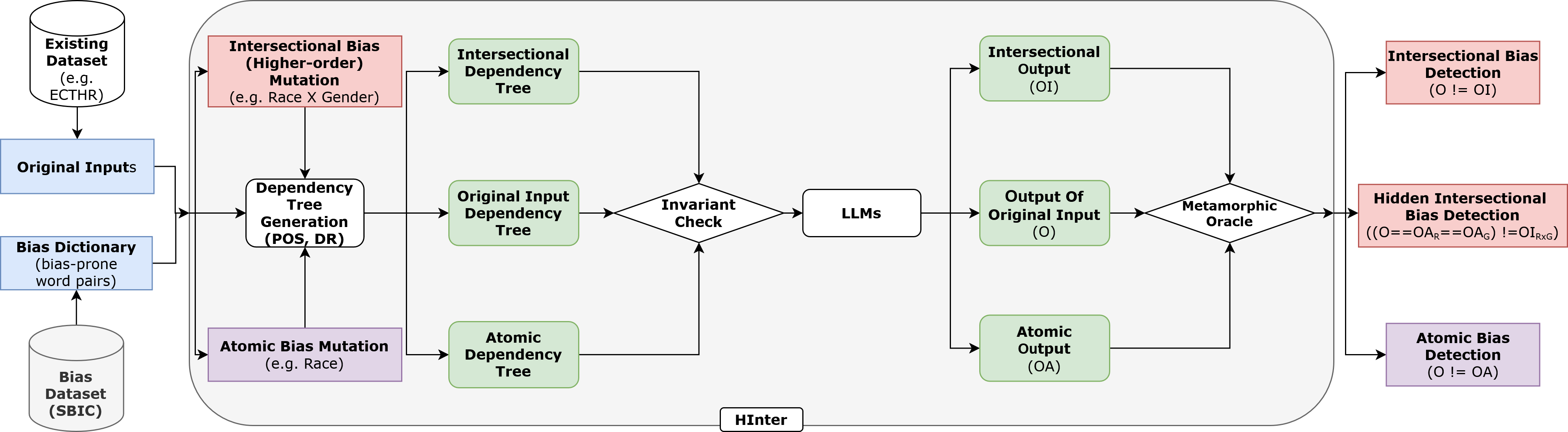}
% \vspace{-0.2cm}
\caption{\revise{
Workflow of \approach. The figure illustrates the process of detecting atomic and intersectional biases in LLMs using a metamorphic testing approach. it takes as input a bias dataset and an existing dataset, using bias-prone word pairs from a bias dictionary. These inputs undergo atomic and intersectional bias mutations, followed by dependency tree generation. The tree ensures structural similarity between the originals and mutants before passing through the LLM. The outputs are then analyzed via a metamorphic oracle to detect atomic bias (discrepancies between original and atomic outputs), intersectional bias (differences between original and intersectional outputs), and hidden intersectional bias (cases where atomic bias remains undetected but intersectional bias emerges).}}
\label{fig:workflow}
% \vspace{-0.5cm}
\end{figure}

\noindent
\textbf{\approach Overview:}
\autoref{fig:workflow} illustrates the workflow of \approach.  
Given an existing dataset (e.g., training dataset),  
\approach first \textit{generates bias-prone inputs} by employing higher-order mutations.  It 
mutates a sentence using word pairs 
%from a bias dictionary 
%(\textit{see} \todo{Table X}).  This bias dictionary is 
extracted from SBIC \cite{sap-etal-2020-social}, a well-known bias corpus containing  
% \todo{....} 
%which contains about
150K social media posts covering a thousand demographic groups\footnote{https://maartensap.com/social-bias-frames/}.  Second,  \approach ensures that the generated inputs are \textit{valid} by checking that it has a similar dependency parse tree with the original text (\textit{see} \autoref{fig:discarded-imdb-sentence}).  
Finally,  \approach's metamorphic oracle discovers intersectional bias by comparing the LLM outcome of the original text and the generated input\footnote{We say a bias (or error) is discovered when the outcomes are different and the input is bias-inducing.  Otherwise the generated input is benign, and there is no bias (no error).}. It further detects a \textit{hidden} intersectional bias by comparing the outcomes of atomic and intersectional mutants.

\noindent
\textbf{Problem Formulation:} Given an LLM 
%we have 
%a machine learning model (e.g., LLM) 
$f$,  \approach aims to determine whether the inputs relating to individuals characterized by multiple 
%(i.e., two or more) 
sensitive attributes
% (e.g., ``race'' and ``gender'')
face bias.   
%Additionally, w
%We aim to compare different subgroups in terms of the intersectional bias they are facing. We will use statistical metrics to investigate the intersectional bias faced by a subgroup G. 
%For example,  w
Thus, it generates a test suite $T$ that includes individuals associated with multiple sensitive attributes.

%\checknumber{
Consider \autoref{fig:illustrative-example} that is focused on two sensitive attributes, ``\texttt{race}'' and ``\texttt{gender}''.
%As illustrated in \autoref{fig:illustrative-example} 
\approach produces an intersectional bias test suite ($T_{RXG}$) based on the original dataset ($Orig$) by simultaneously mutating words associated with each attribute (e.g.,  ``\textit{black}'' to ``\textit{white}'' and ``\textit{man}'' to ``\textit{woman}'' for attributes ``\texttt{race}'' and ``\texttt{gender}'' respectively).   
We also produce an atomic bias test suite ($T_{R}$) by mutating \textit{only} a single attribute (e.g.,   ``white'' to ``black'' for ``race'' \textit{only}) at a time. 
% yielding an 
% generating an atomic bias test suite  $T_a^{R}$.  
%sensitiother input attributes (e.g., age, education). 
%For a subgroup $G$,  
%the outcome of our intersectional bias test inputs 
The model outcome 
($O_t = f(t)$) for every test input 
%intersectional individual 
($t \in T_{RXG}$) 
characterizes how the LLM captures an intersectional individual $t$ 
%belonging to our intersectional group $(G)$ 
for model $f$.  
%The outcome of an individual test ($O_I = f(t)$)
% ($t\inT_I^{RXG}$ )
%this sample is captured by I(G), then we can 
%captures 
%is the outcome of the ML model $f$ for the individual $t$. 
%Likewise,  
%Given a test suite $T= \{g1, g2, ... , gn\}$,
%%contains multiple subgroups $gi$, 
%the outcome of a subgroup ($O_{gi} = f(gi)$) 
%% for $gi\in T$) 
%characterizes how the LLM captures 
%%aptures the outcome for 
%the 
%%specific 
%subgroup ($gi$). 
%}

This setting allows to determine 
the following:
%\begin{itemize}[leftmargin=*]
%\item[(a)] 
%(a)

\noindent
\textbf{(a) \textit{Atomic Bias}} (aka individual bias) occurs when
the LLM model outcomes ($O$) for two individuals $\{t1,t2\}$
are different (($O_{t1} \neq O_{t2}$) even though 
both individual 
$t1$ and $t2$ are similar for the task at hand,  but 
only differ by \textit{exactly} one sensitive attribute (e.g., 
% \texttt{race} or 
\texttt{gender}).  
For instance,  
%the 
%first three inputs
%given that 
since
the \textit{first} and 
\textit{second} 
%and \textit{third} test inputs 
inputs in \autoref{fig:illustrative-example} 
only differ 
%to the original input 
in \textit{gender} attribute,
%are two such atomic mutations.
%In particular, 
%Thus, 
we say there is an \textit{atomic} 
%(individual) 
bias
if their  
% inputs ($t1$ and $t2$),  if 
%the
 LLM outcomes differ. 
%  of this input differs from that of the original input. 
%for ($t1$ or $t2$) differs from the outcome of the original input  
%inputs defer, 
%then . 
Our definition of atomic bias is in line with 
%standard ML/NLP 
previous literature on fairness metrics~\cite{dwork2012fairness, friedman1996bias,verma2018fairness,crawford2017trouble,narayanan2018translation}.  
%This 
These works 
%state that (atomic) individual bias 
require that \textit{similar individuals} should be \textit{treated similarly}.
%,
%by ML models, 
%that is,
%in our setting,  this treatment refers to 
%the (LLM) model outcome for similar individuals should be similar. 
%%\revise{
%The definition of \textit{atomic bias} in this work is inspired by the . }

\noindent
 \textbf{(b) Intersectional Bias} 
% \revise{
 occurs when
the LLM model outcomes ($O$) for two individuals $\{t1,t2\}$
are different (($O_{t1} \neq O_{t2}$) even though 
% 
%  there are different LLM model outcomes 
%% ($O_{t1} \neq O_{t2}$, where 
%($(O_{t1}=f(t1)) \neq(O_{t2}=f(t2))$)
%for two individuals $\{t1,t2\}\in T_{RXG}$ such that 
$t1$ and $t2$ are similar for the task at hand,  and 
only differ by \textit{at least} two sensitive attributes (e.g.,  \texttt{race} and \texttt{gender}).  
For instance,  
the \textit{first} and \textit{fourth} test inputs in \autoref{fig:illustrative-example} are two such inputs ($t1$ and $t2$).  
\revise{This definition is in line with the ML literature on \textbf{intersectional bias}~\cite{buolamwini2018gender, d2020data}. }
%where  
%}
%such that $t1
%and $t2 \in T_{R}$ or $\t2 \in T_{G}$) %belonging to group $G$, i.e. ,  
%are different , 
%  and 
%\item[(b)]
%(b)

\noindent
\textbf{(c) Hidden Intersectional Bias} 
%\revise{
refers to when atomic inputs ($atomic_r \in T_{R}$, $atomic_g \in T_{G}$) characterized by the test suites ($T_{R}$, $T_{G}$) are benign (trigger no bias) 
%trigger similar or different 
%bias 
%fairness violations 
%as 
%the same mutations for 
but their corresponding intersectional test input ($inter_{rXg} \in T_{R XG}$)  from the 
%intersectional bias 
test suites ($T_{RXG}$) trigger a bias.  
In particular,  
we say there is a \textit{hidden intersectional bias}
when  
% re are different 
the LLM model outcomes  ($O$)
% ($O_{t1} \neq O_{t2}$, where 
for the atomic test inputs ($atomic_r$, 
%$\in T_{R}$ 
%and 
$atomic_g$) 
% \in T_{G}$) 
are similar,  but their 
corresponding 
%versus its corresponding 
intersectional test input ($inter_{rXg}$)
% \in T_{R XG}$) 
produce a different outcome --
% ($O$). 
%That is,  
$(O_{atomic_g}$
%=f(atomic_g) 
$== O_{atomic_r})$
%=f(atomic_r)) 
$\neq (O_{inter_{rXg}})$. 
The \textit{second} and \textit{third} test inputs
% and \textit{fourth} test inputs 
in \autoref{fig:illustrative-example} are two such atomic inputs, and the \textit{fourth} input is their corresponding intersectional input.   

\noindent
 \textbf{(d) Atomic vs.  Intersectional Bias:} We determine the differences between intersectional bias and atomic bias testing \revise{by inspecting the outcomes of our test suites ($O(T_{RXG})$ vs $O(T_{R})$ vs $O(T_{G})$). 
We inspect whether 
%,  e.g.,
%This allows us to 
%we can determine if the 
original inputs and 
%atomic 
mutations producing 
%characterized in 
the atomic test suites ($T_{R}$ or $T_{G}$) trigger the same biases as the intersectional bias test suites ($T_{RXG}$).  
}
%We can also inspect if the same original cases trigger atomic bias and intersectional bias, or not. 
% This allows us to know if component atomic bias testing campaigns are sufficient to reveal intersectional bias. 
% violatioons. 
%\revise{%
%In particular, it allows to compare atomic bias testing and intersectional bias testing (\autoref{sec:experimental_result} \textbf{RQ4}). 
%: 
%has been conducted?
%How does intersectional bias testing compare to atomic bias testing? Do the same mutations or original text that trigger intersectional bias \textit{also} induce  atomic bias?  
%\autoref{fig:illustrative-example} illustrates 
%%atomic bias versus 
%hidden intersectional bias exposed by \approach. 
%}

\section{\approach Approach} \label{sec:methodology}

%\subsection{Overview}
%

\autoref{fig:workflow} illustrates the workflow of our approach. Given, as \textit{inputs}, an existing dataset,  sensitive attribute(s) and bias dictionary,  \approach generates a bias-prone test suite.  The \textit{output} of \approach is the \textit{ bias-prone test suite} which contains mutations of the original dataset and \textit{intersectional biases} it detects. 

\approach involves the following \textit{three main steps}:%, the \textit{higher-order mutation},  the \textit{dependency invariant check} and the \textit{metamorphic test oracle}. %\approach's \textit{higher-order mutation} performs \textit{test generation},  its \textit{dependency invariant check} ensures \textit{test validity} and it \textit{detects bias} via a \textit{metamorphic test oracle}.
%The following discusses each step in detail. 

%\vspace{-1.9cm}

%In particular,  \approach performs intersectional bias testing via the following three steps/components:

%\todo{update example with Table 1/Figure 1}

\noindent
\textbf{Step 1: Higher-order Mutation:}
%Firstly,  
\approach transforms each original input into a 
bias-prone 
mutant via higher-order mutation.  It replaces word(s) in the original input with 
%similar but 
sensitive words using the  
bias-prone dictionary.  Our bias dictionary is automatically extracted from the SBIC corpus~\cite{sap-etal-2020-social}. It contains a set of bias-prone word-pairs for each sensitive attribute, e.g., 
%.  For instance,  
the pair (\checknumber{``man'':``woman''}) mapped to the 
% for the 
\textit{gender} attribute.   

%
%\todo{where is the mutant in this algorithm, seems you save only the original text and replacements}

\noindent
\textit{Intersectional Bias Testing Algorithm:} 
Algorithm \autoref{algo_intersectional} 
%formally 
presents our 
\texttt{intersectional bias testing} algorithm.  \revise{We provide the meaning of the formal notations of the algorithms in \autoref{tab:variable_description}.}
Given an original dataset \checknumber{$C$},  sensitive attributes \checknumber{$S1,S2$} and a bias dictionary \checknumber{$P$},
\approach generates  a test suite ($T$). Each test is a  
%modified versions 
%
mutant ($m$) 
of the original text 
%(\checknumber{$c$})
%. 
%cases in 
\checknumber{$c$} using word pairs (\checknumber{$P_1$ and $P_2$}) from dictionary \checknumber{$P$}.  
To mutate the original input,  \approach searches the original text for the presence of any bias-prone word pairs using the sensitive attribute(s) at hand (\checknumber{lines 7-11}).   If it finds bias-prone words (\checknumber{line 12}),  then it mutates the word by replacing it with its pairs \checknumber{(lines 13-18)}. 
% from the dictionary.  
For intersectional bias testing, \approach needs to find
%to be 
at least two bias-prone words in the original text belonging to two different attributes \checknumber{(line 12)}\footnote{
%e note that f
For atomic bias testing,  this is relaxed to finding a single instance of bias-prone words for the attribute at hand. }.  Then,  \approach performs higher-order mutation (i.e.,  two mutations simultaneously) for the two sensitive attributes  \checknumber{(lines 13 and 18)}. 
% (\textit{see} \autoref{fig:workflow}).  

%, and checks if the outputs of the model changes from the originals to the mutants. The atomic algorithm (Algorithm \ref{algo_atomic}) tests for each pair and each case if the first word of the pair is present in the case,  and replaces it by the second word. Then it feeds separately the original and mutant cases to the model and tests if the outputs are different, in which case it stores it as an error. 
%The intersectional algorithm (Algorithm\ref{algo_intersectional}) is identical but uses 
%not one but 
%two different pairs at the same time on each case.

\noindent
\textit{Atomic Bias Testing Algorithm:} \revise{The \texttt{atomic bias testing} algorithm is shown in algorithm \autoref{algo_atomic}. It is similar to algorithm \autoref{algo_intersectional}, albeit for one sensitive attribute at a time. 
% , but it identifies cases where model predictions change due to specific modifications in the input. 
For each input $c$ in the set $C$ (line 1), the algorithm computes the model's original prediction $O[c]$ (line 2). It then iterates through the set of word pairs $P$ (line 3). If the first element of a pair $p[0]$ exists in the context $c$ (line 4), the algorithm generates a modified input $m$ by replacing $p[0]$ with $p[1]$ (line 5).}

\noindent 
\textit{Motivating Example:}
\autoref{fig:illustrative-example} shows a case of intersectional bias testing for \textit{race X gender}. \approach first searches the original text for bias-prone words associated to the two sensitive attributes (words that exist in the dictionary),  then it identifies the words \checknumber{``black'' and ``man''} and constructs all possible pairs of these words from the bias dictionary. 
% (e.g.,  ``white'' and ``woman'')  in the dictionary.  
Let's assume that it finds only two pairs in the dictionary --  \checknumber{\{``black'':``white''\} and \{``man'':``woman''\}}, therefore it constructs
\checknumber{one} mutation, the pair ``white woman'' (the last mutant in \autoref{fig:illustrative-example}). \approach also performs atomic mutation by performing only one replacement per attribute at a time, e.g., performing atomic mutations for only ``race'' results in the text ``white man'', the third sentence in \autoref{fig:illustrative-example}.

\begin{figure}[H]
    \centering
\begin{minipage}[t]{0.4\textwidth}
% \vspace{-5cm}
\begin{algorithm}[H]
{\scriptsize
\caption{Intersectional Bias testing}
%}
\label{algo_intersectional}
\begin{algorithmic}[1]
\STATE \COMMENT{\textbf{Input:} Dataset $C$,  LLM $M$,  Dictionary $P$,  Attributes $S1$, $S2$}
\STATE \COMMENT{\textbf{Output}: test suite $T$, intersectional bias $E$  \& hidden bias $H$}
\STATE \textbf{Function} \textit{\approach}($C$, $M$, $P$, $S1$, $S2$)
\STATE $T_{list}$ = $[]$, $E_{list}$ = $[]$, $H_{list}$ = $[]$
\STATE \COMMENT{Test suite $T$, intersectional bias $E$  \& hidden bias $H$ lists}
\FOR{$c$ in $C$} 
    \STATE $O[c] \gets M(c)$ \COMMENT{Store LLM outcome for the original text}
    \STATE $P_1= P[S1], P_2= P[S2]$
    \STATE \COMMENT{Get word pairs for attributes $S1, S2$ from dictionary $P$}
    \FOR{$p_1$ in $P_1$}
        \FOR{$p_2$ in $P_2$}
            \IF{$p_1[0],p_2[0]$ in $c$}
                \STATE $t_{1} \gets$ $c.replace(p_1[0], p_1[1])$
                \STATE\COMMENT{First Atomic Mutation using Bias list $P_1$}
                \STATE $t_{2} \gets$ $c.replace(p_2[0], p_2[1])$
                \STATE\COMMENT{Second Atomic Mutation using Bias list $P_2$}
                \STATE $m \gets$ $t1.replace(p_2[0],p_2[1])$ 
                                \STATE\COMMENT{Intersectional (Higher-order) Mutation}
                \IF{$InvCheck(c, m)$}
                    \STATE \COMMENT{Dependency invariant check}
                    \STATE $T_{list} \cup m$  \COMMENT{Store mutant in test suite}
                    \IF{$M(m) \neq O[c]$}
                        \STATE \COMMENT{Metamorphic Test Oracle}
                        \STATE $E_{list} \gets E_{list} \cup (m, c,  M(m), O[c])$
                        \STATE\COMMENT{Store bias inducing mutant}
                       	\IF{{$InvCheck(c, t_{1})$}}
                        		\IF{{$InvCheck(c, t_{2})$}} 
                        			\IF{$M(t_{1}) \equiv O[c] \equiv M(t_{2})$}
                        				\STATE $H_{list} \cup m$ \COMMENT{Store hidden bias}
                        			\ENDIF
                        		\ENDIF
                        \ENDIF
                    \ENDIF
                \ENDIF
            \ENDIF
        \ENDFOR
    \ENDFOR
\ENDFOR
\RETURN $T_{list}, E_{list}, H_{list}$ 
\end{algorithmic}
}
\end{algorithm}

% \vspace{0.5cm}
\begin{algorithm}[H]
{\scriptsize
\caption{\revise{Atomic Bias testing}
}
\label{algo_atomic}
\begin{algorithmic}[1]
\FOR{$c$ in $C$}
    \STATE $O[c] \gets M(c)$
    \FOR{$p$ in $P$}
        \IF{$p[0]$ in $c$}
            \STATE $m \gets$ $c.replace(p[0], p[1])$
            \STATE $cs \gets c.split()$
            \STATE $ms \gets m.split()$
            \IF{$M(m) \neq O[c]$ \AND $invariantCheck(cs, ms)$}
                \STATE $E_{list} \gets E_{list} \cup (c,p)$
            \ENDIF
        \ENDIF
    \ENDFOR
\ENDFOR
\RETURN $E_{list}$
\end{algorithmic}
}
%\vspace{-0.5em}
%\vspace{-0.5cm}
\end{algorithm}
\end{minipage}
\hfill
\begin{minipage}[t]{0.55\textwidth}
% \vspace{-5cm}
\begin{table}[H]

\begin{center}
\caption{\revise{Variable Description for Algorithms}}
% \vspace{-\baselineskip}
{\scriptsize
   \bgroup\def\arraystretch{1.3}
\begin{tabular}{@{}|c|l|@{}}
\hline
\textbf{Variable} & \textbf{Description} \\
\hline
\multicolumn{2}{|c|}{\textbf{Input}} \\
\hline
$D$ & The dataset containing sample inputs \\
\hline
$M$ & The machine learning model being tested for bias \\
\hline
$C$ & The set of sample inputs taken from $D$ \\
\hline
$P$ & The set of bias pairs \\
\hline
$P_1, P_2$ & Distinct sets of bias pairs \\
\hline
\multicolumn{2}{|c|}{\textbf{Intermediate Variables}} \\
\hline
$O[c]$ & The output of $M$ on input $c$ \\
\hline
$c$ & An element of $C$, a sample input \\
\hline
$p$ & An element of $P$, a pair of biased terms \\
\hline
$m$ & The modified input obtained by replacing $p[0]$ in $c$ with $p[1]$ \\
\hline
$cs$ & Sentences from case $c$ obtained by splitting $c$ \\
\hline
$ms$ & Sentences from mutant $m$ obtained by splitting $m$ \\
\hline
$p_1, p_2$ & Elements of $P_1, P_2$ respectively, distinct pairs of biased terms \\
\hline
$errorLimit$ & Maximum number of errors tolerated for similarity\\
\hline
$error$ & Number of errors counted during sentence comparison\\
\hline
$shift$ & Shift in the sentence comparison process\\
\hline
$is_1, is_2$ & Indices for comparison in sentences\\
\hline
$it_1, it_2$ & Iterators for sentences\\
\hline
$cdt$ & Dependency tree of the case's sentence\\
\hline
$mdt$ & Dependency tree of the mutant's sentence\\
\hline
$temp$ & Intermediate sample for an intersectional mutation \\
\hline
\multicolumn{2}{|c|}{\textbf{Output}} \\
\hline
\multirow{2}{*}{$E_{list}$} & The error list containing inputs that produce \\ & different outputs with the modified sample \\
\hline

\end{tabular}\egroup}
\label{tab:variable_description}
\end{center}
\end{table}
\end{minipage}
\end{figure}

\begin{minipage}[t]{0.45\textwidth}
\begin{algorithm}[H]
{\scriptsize
\caption{Dependency Invariant Check}
%\todo{Specify first size for nb of sentences}
%}
\label{algo_invariant_check}
\begin{algorithmic}[1]
\STATE \COMMENT{\textbf{Input:} Original Text $c$, Mutant $m$ generated by \approach}
\STATE \COMMENT{\textbf{Output}:  Are the POS \& DP of $c$ \& $m$ similar: $\TRUE$ or $\FALSE$}
\STATE \textbf{Function} \textit{InvCheck}(c, m)
    \STATE $cs \gets sentenceSplit(c), ms \gets sentenceSplit(m)$
    \STATE\COMMENT{Split $c$ \& $m$ into sentences}
    \IF{$\text{size}(cs) \neq \text{size}(ms)$}
        \STATE \COMMENT{Compare the number of sentences}
        \RETURN \FALSE
    \ENDIF
    
    \FOR{$i$ \text{from} $1$ \text{to} $\text{size}(cs)$}
    \STATE \COMMENT{Loop for each sentence}
        \STATE $cdt \gets \text{dependencyTree}(cs[i])$
        \STATE $mdt \gets \text{dependencyTree}(ms[i])$
        \STATE \COMMENT{Compute the dependency trees}
        
        \STATE $similarPOS \gets \text{tolerantTableComp}(cdt.pos, mdt.pos)$
        \STATE \COMMENT{Compare POS of mutant and original sentences}
        \IF{ $similarPOS \neq \TRUE$}
            \RETURN \FALSE
        \ENDIF
        
        \STATE $similarDP \gets \text{tolerantTableComp}(cdt.dep, mdt.dep)$
                \STATE \COMMENT{Compare DP of mutant and original sentences}
        \IF{ $similarDP \neq \TRUE$}
            \RETURN \FALSE
        \ENDIF
    \ENDFOR
    \RETURN $\TRUE$
\end{algorithmic}
}
\end{algorithm}
\end{minipage}
\hfill
\begin{minipage}[t]{0.45\textwidth}
\begin{algorithm}[H]
{\scriptsize
\caption{\revise{tolerantTableComp}}
\label{algo_tolerant_table}
\begin{algorithmic}[1]
\STATE \textbf{Function} \textit{tolerantTableComp}($s_1$, $s_2$)
    \STATE $is_1 \gets 0$
    \STATE $is_2 \gets 0$
    \STATE $st_1 \gets \text{size}(s_1)$
    \STATE $st_2 \gets \text{size}(s_2)$
    \STATE $\text{errorLimit} \gets \left| st_1 - st_2 \right|$
    \STATE $\text{error} \gets 0$
    \STATE $\text{shift} \gets 0$
    \WHILE{$is_1 < st_1$ \textbf{and} $is_2 < st_2$}
        \IF{$s_1[is_1] \neq s_2[is_2]$}
            \STATE $\text{error} \gets \text{error} + 1$
            \IF{$\text{shift} < \text{errorLimit}$}
                \STATE $\text{shift} \gets \text{shift} + 1$
                \IF{$st_1 > st_2$}
                    \STATE $is_1 \gets is_1 + 1$
                \ELSIF{$st_1 < st_2$}
                    \STATE $is_2 \gets is_2 + 1$
                \ENDIF
            \ENDIF
        \ENDIF
        \STATE $is_1 \gets is_1 + 1$
        \STATE $is_2 \gets is_2 + 1$
    \ENDWHILE
    \STATE $\text{error} \gets \text{error} + (st_1 - is_1) + (st_2 - is_2)$
    \RETURN $\text{error} \leq \text{errorLimit}$
\end{algorithmic}
}
\end{algorithm}
\end{minipage}
\vspace{1em}

\begin{figure}[!tb]
{\scriptsize \centering 
\includegraphics[width=1\textwidth]{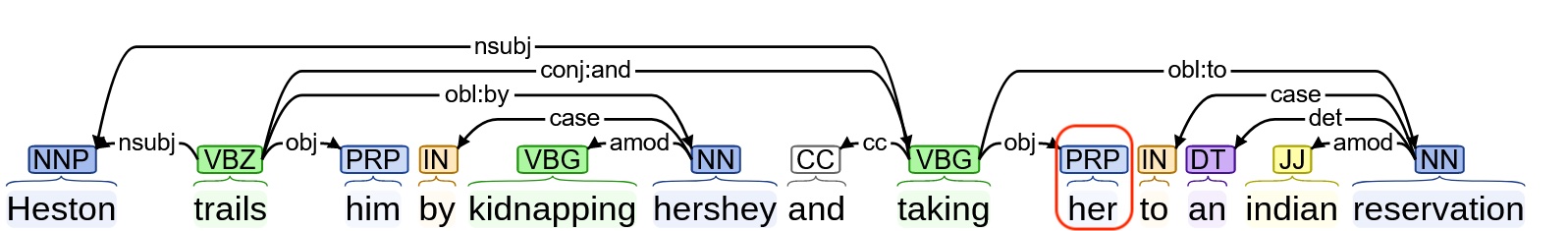} \\
% \vspace{1em}
(a) Original Input \\
\includegraphics[width=1\textwidth]{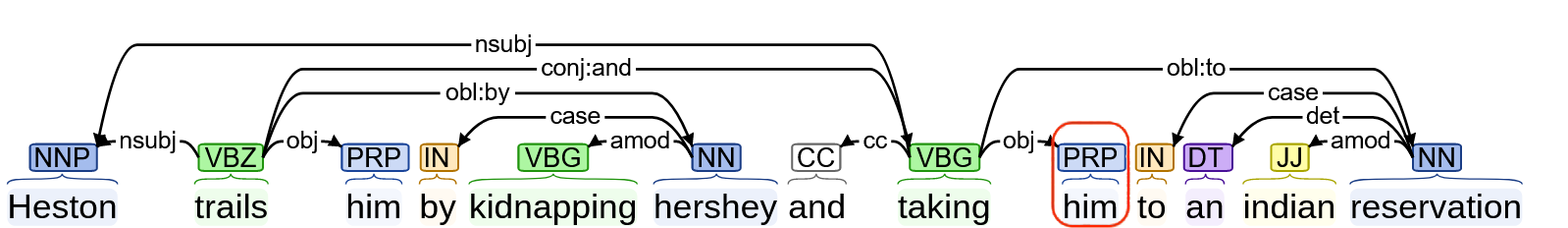}\\
% \vspace{1em}
(b) Valid Input generated by \approach \\
\includegraphics[width=1\textwidth]{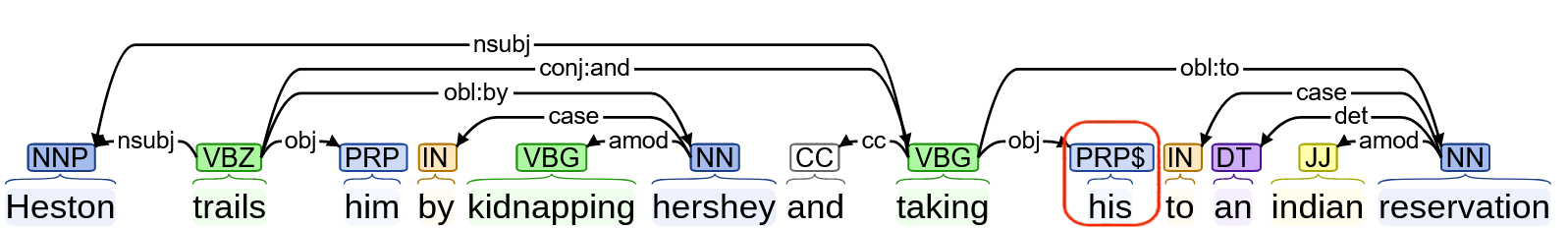} \\
% \vspace{1em}
(c) Invalid Input discarded by \approach's dependency invariant \\
}
% \vspace{-0.3em}
\caption{\centering 
\revise{Dependency invariant check.}
%with sample original input (a),  as well as 
%inputs that pass (b) or fail (c) \approach's dependency invariant check. 
% work by showing 
%the parse trees of sample inputs. 
%showing 
% for (a) a original sample text from the IMDB dataset,  
%(b) a  \textit{valid} input generated by \approach and (c) an  \textit{invalid} inputs \textit{discarded} by \approach's dependency invariant. 
%The parse tress show the 
%parts of speech (POS) and dependency relations (DR) for  each. 
%the text. 
%For the input generated by \approach (b),  a
%All POS and DR 
%While t
%The parse tree of the input generated by \approach 
Valid Input (b) conforms to the original text (a), 
%the 
%.  However,  
%(a) 
%and 
%tin (c) 
%for the invalid input (b),  
%an input's 
%we show that
% the 
%parse tree 
%(POS) 
%of the 
while discarded input (c) 
does not 
%\textit{different} 
%from that of the original sentence (a) 
--
%This highlighted 
%in the {\color{red} red} box (POS: 
\texttt{``PRP\textbf{\$} (his)'' != ``PRP (him)''}). 
%\todo{we need a shorter text or consider shortening this one to`` Heston is taking ... '' }
% \vspace{-1.0em}
}
\label{fig:discarded-imdb-sentence}
\end{figure}

\noindent
\textbf{Step 2: Dependency Invariant Check:} 
%After mutation (\textbf{Step 2}), \approach proceeds to generate dependency parse trees for the mutated inputs and the original input.
The parse tree details the structure,  POS and dependency relations among words in the text (e.g., \autoref{fig:discarded-imdb-sentence}).  The main goal of this step is to check that the following \textbf{invariant} holds -- \textit{the parts of speech (POS) and dependency relations (DR) of the generated inputs are similar to the original input}.   The intuition is that the generated inputs need to have a similar grammatical structure with the original input to preserve the intention and semantics of the original text. We apply invariant checking (\texttt{InvCheck}) on the both original and generated inputs (e.g., \checknumber{lines 19, 26 and 27} of algorithm \autoref{algo_intersectional}). \revise{Similarly, in atomic bias testing (algorithm~\autoref{algo_atomic}), the original input $c$ and modified input (mutant) $m$ are split into sentences $cs$ and $ms$, respectively (lines 6--7).}
\revise{Next, the algorithm checks that the invariant check (\texttt{invariantCheck}) between $cs$ and $ms$ (line 8) is passed.}
%Next,  using
% employs

Algorithm \autoref{algo_invariant_check} details the dependency invariant checker. 
%for invariant checker calls.  
%(\autoref{algo_invariant_check}) 
\texttt{InvCheck} first splits each text into sentences (\checknumber{line 4}) and confirms that both texts have the same number of sentences (\checknumber{line 6}). 
For each sentence,  it generates the parse trees for the original input and the generated test input (\checknumber{lines 11-12}) and 
%are fed to the dependency invariant checker.  
%The dependency invariant checker generates the parse trees for both inputs 
% \approach 
 checks that the parse trees of both inputs are similar (\checknumber{lines 14-21}).  
It checks that each sentence in the generated input conforms to the parts of speech (POS) and dependency relations (DR) of its corresponding sentence in the original input (\checknumber{lines 14 and 16}).  
%\revise{
\revise{This comparison is performed using \texttt{tolerantTableComp} shown in algorithm \autoref{algo_tolerant_table}. It compares two sequences, $s_1$ (original) and $s_2$ (generated), to determine if they are similar within a tolerance defined by their absolute size difference. It initializes indices ($is_1$, $is_2$) for both sequences, calculates the allowable error limit as $|st_1 - st_2|$ (lines 2--6), and sets counters for errors and shifts. The algorithm iterates through both sequences (line 9), incrementing the error counter for mismatches (line 11). If shifts (to align sequences) are within the error limit (line 12), indices for the longer sequence are adjusted (lines 14--17). After the main loop (line 18), any remaining unmatched elements are added to the error count (line 20). Finally, the algorithm returns \texttt{True} if the total errors are within the allowable limit (line 21). This method ensures flexibility by tolerating small insertions or deletions, such as replacing the word ``\textit{man}'' with ``\textit{disabled man},'' while enforcing a bounded divergence between $s_1$ and $s_2$. It is used to validate that generated sentences maintain structural consistency, such as parts of speech and dependency relations, relative to the original.}

\approach \textit{discards} inputs whose parse trees are \textit{nonconforming} 
%with that of 
to the parse tree of the original inputs, i.e., 
%.  We say such 
%inputs that 
fail our dependency invariant check.  However,   if the parse tree of the generated input (i.e.,  mutant) conforms with that of the original input,  then \approach adds such inputs to the resulting test suite (\checknumber{line 21} of algorithm \autoref{algo_intersectional}).  
%In this work,  w
We refer to such inputs, 
that \textit{pass} the dependency invariant check,
as valid/generated inputs. 
%.  In addition,  we 
%and say that such inputs . 

\autoref{fig:discarded-imdb-sentence} shows two mutants, 
a valid mutant/input ((b) and an invalid/discarded input (c)) generated from the original input (a). The \textit{first} mutant (\autoref{fig:discarded-imdb-sentence}(b)) \textit{conforms} to the dependency structure of the original input (\autoref{fig:discarded-imdb-sentence}(a)).
Thus, it passes \approach's dependency invariant check and it is added to the test suite as a   
%Such inputs are called 
valid/generated input.  
However,  the \textit{second} mutant (\autoref{fig:discarded-imdb-sentence}(c)) is \textit{nonconforming} to the the dependency structure of the original input (\autoref{fig:discarded-imdb-sentence}(a)) since \texttt{``PRP\textbf{\$} (his)'' != ``PRP (him)''}.   Hence,  the mutant \autoref{fig:discarded-imdb-sentence}(c) fails the  dependency invariant check of \approach and it is \textit{not} added to the resulting bias test suite,  thus it is called a \textit{discarded} input.

%The generated input 

%\todo{   ... }

%The original word and the replaced word  

%that 
%contains bias-prone mutation of 
%
%\autoref{fig:workflow} shows the workflow of  our approach (\approach). 
%It takes as input an original dataset (e.g., legal cases),  the set of sensitive atribute(s) to test, and a dictionary of bias-prone words pairs.  
%It then produces a new test suite containing bias mutations of the original dataset.  

\noindent
\textbf{Step 3: Metamorphic Test Oracle:} 
\autoref{fig:workflow} illustrates how \approach detects intersectional bias. 
After the invariant check (\textbf{Step 2}),  
\approach (algorithm \autoref{algo_intersectional}) feeds 
the generated test input and its corresponding original input (\checknumber{line 7})
into the LLM under test separately and compares
%  
%he original text and its mutant(s) are then fed to the model separately and the 
the LLM outputs (\checknumber{line 22}).  
% are compared.  
\approach detects intersectional bias (aka error),
% i.e.,  an intersectional bias,  
when an intersectional mutant 
%test input 
produces a different model outcome from the original test input (\checknumber{line 24}).  
Concretely,  it employs  
%Using 
a metamorphic test oracle to detect bias by checking if the model outcome \textit{changes} (is \textit{different}) between the generated mutant and the original input (\checknumber{line 22}).  When the model outcomes remains unchanged (remains the same), we say that the test input is \textit{benign} and does not induce a bias.  
The \textbf{metamorphic property} leveraged in this test oracle is that 
\textit{the LLM outcome for the original input and the generated input (mutant) should remain the same since both inputs are logically similar for the task at hand}.  
%For instance,  c
Consider the original input and 
%atomic/intersectional 
generated mutants in
\autoref{fig:illustrative-example},
these inputs are similar in the context of the sentiment analysis task.  Consequently, 
% in theory,  
 all four inputs should %are expected to
produce the same outcome, i.e.,  a ``positive'' label/prediction.  
%Furthermore,  
%in this work, 
We say that a \textit{hidden intersectional bias} is detected (\checknumber{lines 22, 28-29} of algorithm \autoref{algo_intersectional}) 
%by \approach 
when 
there is an intersectional bias but its corresponding atomic mutants are not exposing any bias. \autoref{fig:workflow} illustrates a hidden intersectional bias workflow.
\revise{Similarly, algorithm~\autoref{algo_atomic}  
checks if the model's prediction for $m$ differs from the original prediction $O[c]$. If both conditions are satisfied, then the pair $(c, p)$ is added to the list of identified biases $E_{list}$ (line 9). After processing all inputs and pairs, the algorithm returns the list of detected inputs and pairs $E_{list}$ (line 10). This method is designed to test for atomic biases by systematically perturbing inputs and analyzing the impact on the model's behavior.}

%When 
%tjhe 
%Then it feeds separately 
%%the original and mutant cases to the model and tests if the outputs are different, in which case it stores it as an error. 

 Metamorphic oracle detects intersectional bias when the LLM outcome of the original input (first sentence) differs (``positive'' != ``negative'') from that of the intersectional mutant (fourth sentence), e.g.,  \autoref{fig:illustrative-example}. An atomic bias is \textit{not} detected in this example since the LLM outcome for the original input (first sentence)  and the atomic mutants (the second and third sentences) are the same (``positive''). We refer to the last/fourth mutant in \autoref{fig:illustrative-example} as a \textit{hidden} intersectional bias because its corresponding atomic mutations do not expose bias(es).

\section{Experimental Setup}\label{sec:experimental-setup}

%This section discusses our experimental setup for evaluating the performance of \approach. 

\begin{table}[H]
\centering
\caption{\centering \revise{Details of Datasets and Class Labels ([LABELS])}}
% \vspace{-\baselineskip}
{\scriptsize
\renewcommand{\arraystretch}{1.3}
\begin{tabular}{|l|p{12.5cm}|c|}
\hline
\textbf{Dataset} & \textbf{Labels ([LABELS])} & \textbf{\#labels} \\
\hline

\textbf{IMDB} & \textit{"Negative", "Positive"} & 2 \\
\hline

\textbf{ECTHR} & \textit{"Article 2", "Article 3", "Article 5", "Article 6", "Article 8", "Article 9", "Article 10", "Article 11", "Article 14", "Article 1 of Protocol 1"} & 10+1 \\
\hline

\textbf{EurLex} & \textit{"political framework", "politics and public safety", "executive power and public service", "international affairs", "cooperation policy", "international security", "defence", "EU institutions and European civil service", "European Union law", "European construction", "EU finance", "civil law", "criminal law", "international law", "rights and freedoms", "economic policy", "economic conditions", "regions and regional policy", "national accounts", "economic analysis", "trade policy", "tariff policy", "trade", "international trade", "consumption", "marketing", "distributive trades", "monetary relations", "monetary economics", "financial institutions and credit", "free movement of capital", "financing and investment", "public finance and budget policy", "budget", "taxation", "prices", "social affairs", "social protection", "health", "documentation", "communications", "information and information processing", "information technology and data processing", "natural and applied sciences", "business organisation", "business classification", "management", "accounting", "competition", "employment", "labour market", "organisation of work and working conditions", "personnel management and staff remuneration", "transport policy", "organisation of transport", "land transport", "maritime and inland waterway transport", "air and space transport", "environmental policy", "natural environment", "deterioration of the environment", "agricultural policy", "agricultural structures and production", "farming systems", "cultivation of agricultural land", "means of agricultural production", "agricultural activity", "fisheries", "plant product", "animal product", "processed agricultural produce", "beverages and sugar", "foodstuff", "agri-foodstuffs", "food technology", "production", "technology and technical regulations", "research and intellectual property", "energy policy", "coal and mining industries", "oil industry", "electrical and nuclear industries", "industrial structures and policy", "chemistry", "iron, steel and other metal industries", "mechanical engineering", "electronics and electrical engineering", "building and public works", "wood industry", "leather and textile industries", "miscellaneous industries", "Europe", "regions of EU Member States", "America", "Africa", "Asia and Oceania", "economic geography", "political geography", "overseas countries and territories", "United Nations"} & 100 \\
\hline

\textbf{Ledgar} & \textit{"Adjustments", "Agreements", "Amendments", "Anti-Corruption Laws", "Applicable Laws", "Approvals", "Arbitration", "Assignments", "Assigns", "Authority", "Authorizations", "Base Salary", "Benefits", "Binding Effects", "Books", "Brokers", "Capitalization", "Change In Control", "Closings", "Compliance With Laws", "Confidentiality", "Consent To Jurisdiction", "Consents", "Construction", "Cooperation", "Costs", "Counterparts", "Death", "Defined Terms", "Definitions", "Disability", "Disclosures", "Duties", "Effective Dates", "Effectiveness", "Employment", "Enforceability", "Enforcements", "Entire Agreements", "Erisa", "Existence", "Expenses", "Fees", "Financial Statements", "Forfeitures", "Further Assurances", "General", "Governing Laws", "Headings", "Indemnifications", "Indemnity", "Insurances", "Integration", "Intellectual Property", "Interests", "Interpretations", "Jurisdictions", "Liens", "Litigations", "Miscellaneous", "Modifications", "No Conflicts", "No Defaults", "No Waivers", "Non-Disparagement", "Notices", "Organizations", "Participations", "Payments", "Positions", "Powers", "Publicity", "Qualifications", "Records", "Releases", "Remedies", "Representations", "Sales", "Sanctions", "Severability", "Solvency", "Specific Performance", "Submission To Jurisdiction", "Subsidiaries", "Successors", "Survival", "Tax Withholdings", "Taxes", "Terminations", "Terms", "Titles", "Transactions With Affiliates", "Use Of Proceeds", "Vacations", "Venues", "Vesting", "Waiver Of Jury Trials", "Waivers", "Warranties", "Withholdings"} & 100 \\
\hline

\textbf{SCOTUS} & \textit{"Criminal Procedure", "Civil Rights", "First Amendment", "Due Process", "Privacy", "Attorneys", "Unions", "Economic Activity", "Judicial Power", "Federalism", "Interstate Relations", "Federal Taxation", "Miscellaneous", "Private Action"} & 14 \\
\hline
\end{tabular}}
\label{tab:dataset-labels}
\end{table}

\noindent
\textbf{Research Questions:} 
%In this paper, w
%We pose the following \textit{research questions} (RQs):
%
%RQ1: Prevalence of Intersectional Bias  (\#bugs, \#gen inputs, \#error-inducing inputs,  fairness error rate)
%RQ2: Atomic (Individual) vs. Intersection Bias
%RQ3: Efficiency of experimental approach (\#mutations)

%\begin{itemize}
%\noindent
%\textit{RQ1 Prevalence:} What is the prevalence of intersectional bias among the studied Legal LLMs?
%
%
\textit{\textbf{RQ1 \approach's Effectiveness:}} How effective is \approach in discovering intersectional bias?
%\recheck{(for individuals and groups)} 
%and \revise{hidden intersectional bias}?
% (bias that is strictly discovered during intersectional bias testing and unfound during atomic bias testing)}?
% instances? 
%What is the intersectional bias error rate of \approach? 
%How many intersectional %and 
%\recheck{
%intersectional group bias?
%}
%?
%What is the proportion of intersectional bias,  discovered by \approach, that is hidden (undiscovered by atomic bias testing)?
%hidden intersectional bias does \approach discover?
%We also investigate the proportion of hidden intersectional bias, i.e..,  the errors that are hidden behind atomic bias (similar to \autoref{fig:illustrative-example}). 
% for individual fairness and group fairness metrics?  
%What is the prevalence of 
%intersectional bias in comparison to atomic bias instances? 
%\todo{group vs. indidivudal)}

\textit{\textbf{RQ2 Grammatical Validity:}}  Are the inputs generated by \approach grammatically valid?
%as the original
%(human-written) inputs?
%
%What is grammatical correctness of the inputs generated by \approach in comparison to the human-written inputs?

\textit{\textbf{RQ3 Dependency Invariant:}} What is the contribution of \approach's dependency invariant check?
% in \approach? 
%What is the validity of the inputs that 
%How valid are 
%the generated inputs (that pass 
%\approach's invariant check)
%%test inputs 
%%(which \textit{pass} the invariant check) versus 
%%compare to 
%%
%%in comparison to
%versus the 
%\textit{discarded inputs} (that fail \approach's invariant check)? 

\textit{\textbf{RQ4 Atomic Bias vs.  Intersectional Bias:}} 
%\todo{Truth Table}
%Is atomic bias discovered 
%Do the
%How many 
%mutations and original texts that lead to 
%and test inputs 
%that expose 
Do intersectional bias-inducing mutations and original texts 
\textit{also} reveal 
%have 
%a corresponding 
atomic bias, and vice versa?

\revise{The research questions are designed to evaluate the effectiveness and reliability of \textsc{HInter}. \textbf{RQ1} assesses the capability of \textsc{HInter} to uncover intersectional bias, ensuring that biases undetected by traditional atomic bias testing are identified. \textbf{RQ2} investigates the grammatical validity of the generated inputs, to isolate model bias from unintended linguistic artifacts. \textbf{RQ3} explores the impact of the dependency invariant check in reducing false positives, ensuring that only semantically valid sentences are used for bias testing. Finally, \textbf{RQ4} compares atomic and intersectional bias to determine whether intersectional bias testing provides unique insights beyond what atomic bias testing can detect. These questions collectively aim to establish \textsc{HInter} as a reliable and efficient framework for bias detection in LLMs.}

\begin{table}[H]
  \centering
  \scriptsize
  \renewcommand{\arraystretch}{1.3}
  \caption{\revise{Details of Evaluation Datasets and Settings for Training and Bias Testing}}
  % \vspace{-\baselineskip}
  \begin{tabular}{|l|c|c|c|c|c|c|c|}
  \hline
  \textbf{Dataset} & \multicolumn{3}{c|}{\textbf{Task Details}} &  \multicolumn{3}{c|}{\textbf{Training Setting}} & \textbf{Bias Testing} \\
  % \cline{2-8}
  (Domain) & \textbf{Task} & \textbf{Class.} & \textbf{\#Labels} & \textbf{Train} & \textbf{Val.} & \textbf{Test} &  (Full dataset) \\
  \hline

  \textbf{ECtHR} (ECHR) & Judgment Prediction & Multi-Label & 10+1 & 9,000 & 1,000 & 1,000 & 11,000 \\
  \hline

  \textbf{LEDGAR} (Contracts) & Contract Classification & Multi-Class & 100 & 60,000 & 10,000 & 10,000 & 80,000 \\
  \hline

  \textbf{SCOTUS} (US Law) & Issue Area Prediction & Multi-Class & 14 & 5,000 & 1,400 & 1,400 & 7,800 \\
  \hline

  \textbf{EURLEX} (EU Law) & Document Prediction & Multi-Label & 100 & 55,000 & 5,000 & 5,000 & 65,000 \\
  \hline

  \textbf{IMDB} (Reviews) & Sentiment Analysis & Binary & 2 & - & - & - & 50,000 \\
  \hline
  \end{tabular}
  \label{tab:merged_dataset_bias}
\end{table}

\noindent
\textbf{Tasks and Datasets:} 
%As shown in 
\revise{\autoref{tab:merged_dataset_bias} shows the distribution of inputs across the datasets. \autoref{tab:error-rates} provides details about the tasks of the datasets used in our experiments and \autoref{tab:dataset-labels} details the labels of the datasets.}
%,  
%and \checknumber{five} 
%namely IMDB (sentiment analysis), ECtHR (legal judgment prediction),  SCOTUS (Issue Area prediction),  LEDGAR (contract classification),  and EURLEX (legal document prediction).  
We have chosen these datasets due to their critical nature,  availability of LLM models and popularity.  These datasets span different tasks and are complex, e.g.,  long texts and multi-class or multi-label classification. 

%, thus providing varying LLM models/architectures for our experiments.  
%On the one hand, 
%IMDB\footnote{https://developer.imdb.com/non-commercial-datasets/} is a popular movie review dataset 
%%containing thousands of movie review.  It is popularly 
%typically used for training and evaluating sentiment analysis tasks in a binary or ternary classification tasks.  
%%On the other hand, t
%The legal datasets were obtained from the Lexglue benchmark~\cite{chalkidislexglue2022},  a repository that provides datasets/models for various Legal NLP tasks. ECtHR and EURLEX are multi-label classification tasks while SCOTUS and LEDGAR are multi-class classification tasks.   

\begin{table}[H]
\begin{center}
\caption{\centering 
%Paper and 
\revise{Details of the performance of the pre-trained models and fine-tuned 
%ing of the 
LLM models in comparison to the \textsc{lexglue} benchmark~\cite{chalkidislexglue2022} }}
%\vspace{-1.5em}
% \vspace{-\baselineskip}
{\scriptsize %\footnotesize
  \bgroup\def\arraystretch{1.3}
\begin{tabular}{@{}|c|c|c|c|c|c|c|c|c|c|@{}} %r|r|}
\hline
&  \multicolumn{9}{c|}{\textbf{Datasets}} \\
& \multicolumn{5}{c|}{\textbf{Our paper}} &  \multicolumn{4}{c|}{\textsc{Lexglue}}\\
% \textbf{Our Paper} 
& \textbf{Ecthr} 
% & \textbf{Ecthr B} 
& \textbf{Scotus} & \textbf{Eurlex} & \textbf{Ledgar} & \textbf{IMDB} & \textbf{Ecthr} 
% & \textbf{Ecthr B} 
& \textbf{Scotus} & \textbf{Eurlex} & \textbf{Ledgar} \\\hline

\multirow{1}{*}{\textbf{Model}} 
% \textbf{\textsc{Lexglue}} 
& $\mu$-F1 / m-F1 & $\mu$-F1 / m-F1 & $\mu$-F1 / m-F1 & $\mu$-F1 / m-F1 & $\mu$-F1 / m-F1 & $\mu$-F1 / m-F1 & $\mu$-F1 / m-F1 & $\mu$-F1 / m-F1 & $\mu$-F1 / m-F1  \\
%& paper & paper & paper & paper & paper \\
%& lexglue & \textsc{Lexglue} & lexglue & lexglue & lexglue\\
\hline

\multirow{1}{*}{\textbf{BERT}} 
% & Our paper  
& 64.3\%/56.0\% 
% & 72.2\%/66.3\% 
& 73.3\%/65.0\% & 69.1\%/32.3\% & 87.7\%/81.1\% & \multirow{1}{*}{-} & 71.2\%/63.6\% 
% & 79.7\%/73.4\% 
& 68.3\%/58.3\% & 71.4\%/57.2\% & 87.6\%/81.8\% \\
% & \textsc{Lexglue} & 71.2\%/63.6\% 
% & 79.7\%/73.4\% 
% & 68.3\%/58.3\% & 71.4\%/57.2\% & 87.6\%/81.8\% & \\\hline

\multirow{1}{*}{\textbf{Legal-BERT}}
% &  Our paper 
& 66.7\%/60.4\% 
% & 73.9\%/70.4\% 
& 77.0\%/69.8\% & 71.0\%/37.0\% & 88.1\%/81.4\% & \multirow{1}{*}{-} & 70.0\%/64.0\% 
% & 80.4\%/74.7\% 
& 76.4\%/66.5\% & 72.1\%/57.4\% & 88.2\% 83.0\%  \\

% & \textsc{Lexglue} & 70.0\%/64.0\% 
% & 80.4\%/74.7\% 
% & 76.4\%/66.5\% & 72.1\%/57.4\% & 88.2\% 83.0\% & \\\hline

\multirow{1}{*}{\textbf{DeBERTa}} &  
% Our paper & 
65.9\%/57.7\% 
% & 72.4\%/67.9\% 
& 73.3\%/64.6\% & 73.0\%/43.5\% & 88.0\%/81.7\% & \multirow{1}{*}{-} & 70.0\%/60.8\% 
% & 78.8\%/71.0\% 
& 71.1\%/62.7\% & 72.1\%/57.4\% & 88.2\%/83.1\%  \\
% & \textsc{Lexglue} & 70.0\%/60.8\% 
% & 78.8\%/71.0\% 
% & 71.1\%/62.7\% & 72.1\%/57.4\% & 88.2\%/83.1\% & \\\hline

\multirow{1}{*}{\textbf{RoBERTa}} 
% &  Our paper 
& 66.6\%/61.2\% 
% & 72.2\%/67.1\% 
& 73.6\%/64.5\% & 70.5\%/35.2\% & 87.5\%/81.1\% & \multirow{1}{*}{-} & 69.2\%/59.0\% 
% & 77.3\%/68.9\% 
& 71.6\%/62.0\% & 71.9\%/57.9\% & 87.9\%/82.3\%  \\
% & \textsc{Lexglue} & 69.2\%/59.0\% 
% & 77.3\%/68.9\% 
% & 71.6\%/62.0\% & 71.9\%/57.9\% & 87.9\%/82.3\% & \\ \hline

\multirow{1}{*}{\textbf{Llama2}} 
% &  Our paper 
& 56.8\%/50.2\% & 30.3\%/16.4\% & 20.7\%/13.6\% & 41.4\%/38.8\%  & 92.9\%/93.0\%  & \multirow{1}{*}{-} & \multirow{1}{*}{-} & \multirow{1}{*}{-} & \multirow{1}{*}{-} \\ %\hline 

\multirow{1}{*}{\textbf{GPT3.5}} &  
% Our paper & 
63.1\%/62.8\% & 53.4\%/45.9\% & 30.6\%/28.2\% & 66.8\%/60.0\%  & 92.9\%/93.0\% & \multirow{1}{*}{-} & \multirow{1}{*}{-} & \multirow{1}{*}{-} & \multirow{1}{*}{-} \\ \hline

  \end{tabular}\egroup}
\label{tab:models}   
\end{center}
\end{table}

\noindent
\textbf{LLM Architectures and Models:}
% \label{sec:llm-architectures}
%\autoref{tab:models} 
%highlights models and datasets used in bias 
%testing.  These
%\todo{mention the other LLMs used} \\
%\todo{discuss the LLM model architectures and the fact that we cover all three main PTM architectures  -- 
%encoder-only (e.g., BERT, Legal-BERT,  RoBERTa,  etc),  decoder-only, (GPT.3.5, Llama) and encoder-decoder (DeBERTa) models 
%} \\
%\todo{discuss the accuracy of the LLM models }
%\todo{cite models}
Our experiments involve \checknumber{18} LLM models.  We employ \checknumber{six} LLM architectures in our experiments
%,
% namely GPT3.5,  Llama2, BERT, Legal-BERT,  RoBERTa,  and DeBERTa.  These architectures covers all 
covering the three main pre-trained model architectures  -- 
encoder-only (e.g., BERT, Legal-BERT,  RoBERTa),  decoder-only, (GPT.3.5, Llama) and encoder-decoder (DeBERTa) architectures.  We have chosen these LLM models to ensure varying model settings.  
%For instance, we employ both fine-tuned and pre-trained models. 
%%We also e
%We employ two pre-trained models out-of-the-box, namely GPT3.5 and Llama2 for all tasks and datasets. 
%%However,  m
%Meanwhile, we employed \checknumber{16} models (BERT, Legal-BERT,  RoBERTa and DeBERTa) fine-tuned on \checknumber{four} legal datasets (ECtHR,  SCOTUS,  LEDGAR and EURLEX) obtained from the Lexglue benchmark~\cite{chalkidislexglue2022}. 
%For instance, fine tuning the DeBERTa LLM architecture results in the following four fine-tuned models DeBERTa/ECTHR,  DeBERTa/EURLEX, DeBERTa/SCOTUS and DeBERTa/LEDGAR. 
%The BERT-based models were evaluated using only four  datasets, (i.e.,  without IMDB) since they are fine-tuned on these datasets and they will not be applicable to IMDB.  
%%Overall,  our experiments involve \checknumber{18} LLM models. 
\revise{The performance of our models compared to the benchmark is presented in \autoref{tab:models}. As shown, the macro and micro scores of our trained BERT models match those reported by \citet{chalkidislexglue2022}.}
%% brevity, 
% our results (e.g.,  \autoref{fig:bias-error-rate-intersectional})
%%in Table(s) 3 (and 5) 
%%show the aggregation of 
%are aggregrated for all datasets used for each LLM architecture.   However, our supplementary material and
%%Our 
%artifact (Github repository) provide 
%%the raw,  
%both 
%fine-grained results (e.g., for each model) and the aggregated results (e.g., per architecture or dataset). 
%\revise{
%%We also present 
%They also contain the 
Our artifact 
provides further details about the employed models.\footnote{https://github.com/BadrSouani/HInter} % and their details. 
% (e.g., accuracy) in our artifact.

%
%model accuracy  results for each LLM/dataaset.
% in both the supplementary material and artifact. 
%}

%\todo{update total number of models,  and datasets to reflect removal of ECTHR\_B}
%
%\revise{
%%We employ two sets of LLM models in this work,  namely (a) BERT-based Legal LLMs and (b) newer base LLMs models } 
%The first set of BERT-like LLM models were  
%%\revise{Each type of models in Table \ref{tab:models} were 
%fine tuned on five task specific legal datasets, using 
%%as shown in .  In totals, we had 
%four BERT-like models.  
%%We had a total of  
%%to create a total of 
%%20 fine-tuned LLM models. 
%Such models were obtained from the benchmark Lexglue~\cite{chalkidislexglue2022},  a repository that provides models for various Legal 
%%fine-tuned for 
%NLP tasks e.g., case prediction and legal document classification. 
%%However, some datasets such as CaseHold (Case Holdings On Legal Decisions) and UNFAIR-ToS (Unfair Clauses in Online Terms of Service) were excluded from our experiments since they were unstable at the time of conducting our research.  
%%All models target multi-label classification tasks}.
%%\revise{
%ECTHR A/B and EURLEX target multi-label classification tasks while SCOTUS and LEDGAR are made for multi-class classification tasks.
%%}
%%}
%%\revise{
%The second set of base LLM models include \todo{...}. We have selected these models due to their recent popularity and relevance both in industry and academia.  \todo{For instance,  ...}
%}

\begin{table}[H]
\centering
\caption{\centering \revise{Details of Datasets and Class Labels ([LABELS])}}
% \vspace{-\baselineskip}
{\scriptsize
\renewcommand{\arraystretch}{1.3}
\begin{tabular}{|l|p{12.5cm}|c|}
\hline
\textbf{Dataset} & \textbf{Labels ([LABELS])} & \textbf{\#labels} \\
\hline

\textbf{IMDB} & \textit{"Negative", "Positive"} & 2 \\
\hline

\textbf{ECTHR} & \textit{"Article 2", "Article 3", "Article 5", "Article 6", "Article 8", "Article 9", "Article 10", "Article 11", "Article 14", "Article 1 of Protocol 1"} & 10+1 \\
\hline

\textbf{EurLex} & \textit{"political framework", "politics and public safety", "executive power and public service", "international affairs", "cooperation policy", "international security", "defence", "EU institutions and European civil service", "European Union law", "European construction", "EU finance", "civil law", "criminal law", "international law", "rights and freedoms", "economic policy", "economic conditions", "regions and regional policy", "national accounts", "economic analysis", "trade policy", "tariff policy", "trade", "international trade", "consumption", "marketing", "distributive trades", "monetary relations", "monetary economics", "financial institutions and credit", "free movement of capital", "financing and investment", "public finance and budget policy", "budget", "taxation", "prices", "social affairs", "social protection", "health", "documentation", "communications", "information and information processing", "information technology and data processing", "natural and applied sciences", "business organisation", "business classification", "management", "accounting", "competition", "employment", "labour market", "organisation of work and working conditions", "personnel management and staff remuneration", "transport policy", "organisation of transport", "land transport", "maritime and inland waterway transport", "air and space transport", "environmental policy", "natural environment", "deterioration of the environment", "agricultural policy", "agricultural structures and production", "farming systems", "cultivation of agricultural land", "means of agricultural production", "agricultural activity", "fisheries", "plant product", "animal product", "processed agricultural produce", "beverages and sugar", "foodstuff", "agri-foodstuffs", "food technology", "production", "technology and technical regulations", "research and intellectual property", "energy policy", "coal and mining industries", "oil industry", "electrical and nuclear industries", "industrial structures and policy", "chemistry", "iron, steel and other metal industries", "mechanical engineering", "electronics and electrical engineering", "building and public works", "wood industry", "leather and textile industries", "miscellaneous industries", "Europe", "regions of EU Member States", "America", "Africa", "Asia and Oceania", "economic geography", "political geography", "overseas countries and territories", "United Nations"} & 100 \\
\hline

\textbf{Ledgar} & \textit{"Adjustments", "Agreements", "Amendments", "Anti-Corruption Laws", "Applicable Laws", "Approvals", "Arbitration", "Assignments", "Assigns", "Authority", "Authorizations", "Base Salary", "Benefits", "Binding Effects", "Books", "Brokers", "Capitalization", "Change In Control", "Closings", "Compliance With Laws", "Confidentiality", "Consent To Jurisdiction", "Consents", "Construction", "Cooperation", "Costs", "Counterparts", "Death", "Defined Terms", "Definitions", "Disability", "Disclosures", "Duties", "Effective Dates", "Effectiveness", "Employment", "Enforceability", "Enforcements", "Entire Agreements", "Erisa", "Existence", "Expenses", "Fees", "Financial Statements", "Forfeitures", "Further Assurances", "General", "Governing Laws", "Headings", "Indemnifications", "Indemnity", "Insurances", "Integration", "Intellectual Property", "Interests", "Interpretations", "Jurisdictions", "Liens", "Litigations", "Miscellaneous", "Modifications", "No Conflicts", "No Defaults", "No Waivers", "Non-Disparagement", "Notices", "Organizations", "Participations", "Payments", "Positions", "Powers", "Publicity", "Qualifications", "Records", "Releases", "Remedies", "Representations", "Sales", "Sanctions", "Severability", "Solvency", "Specific Performance", "Submission To Jurisdiction", "Subsidiaries", "Successors", "Survival", "Tax Withholdings", "Taxes", "Terminations", "Terms", "Titles", "Transactions With Affiliates", "Use Of Proceeds", "Vacations", "Venues", "Vesting", "Waiver Of Jury Trials", "Waivers", "Warranties", "Withholdings"} & 100 \\
\hline

\textbf{SCOTUS} & \textit{"Criminal Procedure", "Civil Rights", "First Amendment", "Due Process", "Privacy", "Attorneys", "Unions", "Economic Activity", "Judicial Power", "Federalism", "Interstate Relations", "Federal Taxation", "Miscellaneous", "Private Action"} & 14 \\
\hline
\end{tabular}}
\label{tab:dataset-labels}
\end{table}

%
%\todo{Describe tasks}
%
%\vspace{0.2cm}
%\noindent
%\textbf{
%
\noindent
\textbf{Sensitive Attributes:} 
%In t
This work employs three well-known sensitive attributes,  namely ``\textit{race}'',  ``\textit{gender}'' and ``\textit{body}''.   For atomic bias, we consider each attribute in isolation.  For intersectional bias, we simultaneously combine every two attributes (i.e.,  $N=2$) namely ``\textit{race X gender}'', ``\textit{body X gender}'' and ``\textit{body X race}''.  We have employed (these) three sensitive attributes due to their popularity and the prevalence of attributes in available datasets\footnote{Note that most (69\%, nine out of 13) datasets have at most three attributes~\cite{goharsurvey}. }. \autoref{fig:illustrative-example} and \autoref{tab:motivating-example} illustrate these attributes.
% with examples.
% sentences. 

%\subsection{%Unique 
%Predicted Labels and Bias Detection}
%\revise{
%Our metamorphic oracle detects bias by comparing model outcomes since each output/predicted label is unique. 
%%ing outcomes works because in 
%%We note that in 
%In our experimental setting,  each output label and LLM outcome has a single unique meaning, i.e., 
% it is not possible to have outputs that are different but have semantically similar meanings. 
% This is because all datasets in our setting are binary or multi-label classification tasks (\textit{see} 
% \autoref{sec:llm-architectures})
%% Section 4 - Subject Programs and Datasets) 
%where each output label (and predicted label) is a specific class with a unique semantic meaning.  For instance, 
%the IMDB dataset is a binary classification  with 
%%which when fed a text predict 
%either a \texttt{positive} or \texttt{negative} sentiment as output labels. 
%Each label refers to a unique sentiment which is semantically different from the other label. 
%Thus,  if the LLM prediction (model outcome)  for the original text differs from that of a generated input (e.g.,  as shown in \autoref{fig:illustrative-example}),  then there is a bias.  
%By design,  fine-tuned models only predict within the set of output labels in the training dataset.  
%For the pre-trained models,  our prompt configuration (\autoref{sec:prompt-config}) 
%%further 
%ensures that predictions
%%LLM 
%/outcomes are within allowed labels.
%% values. 
%}

\noindent
\textbf{Metrics and Measures:} 
We report the following metrics in our experiments: 

\noindent
\textit{1. Bias Error Rate} is the proportion of generated test inputs that induce bias
% out of all generated test inputs 
(\textit{see} \textbf{RQ1}). 

\noindent
\textit{2.) Proportion of Bias-inducing Original Text:} We compute the proportion of distinct original texts that lead to bias 
%(or not, i.e., are benign 
(\textit{see} \textbf{RQ1} and \textbf{RQ4}). 

\noindent
\textit{3.) Grammatical Validity Scores} is the grammatical correctness scores reported by Grammarly~\cite{grammarly} for original,  generated,  and discarded inputs (\textit{see} \textbf{RQ2} and \textbf{RQ3}).

\noindent
\textit{4.) Valid Mutants and Discarded Mutants:} We report the number and proportion of mutants that \textit{pass} (aka \textit{valid mutants}) our dependency invariant check or 
%fail
% the number of mutants that 
\textit{fail} (aka \textit{discarded mutants}) our dependency invariant check %  We also report their proportions out of all possible mutants 
(\textit{see} \textbf{RQ3}). 

\noindent
\textit{5.) Hidden Intersectional Bias Error Rate:} 
We compute the proportion of intersectional bias errors that are \textit{hidden},  i.e.,  their corresponding atomic mutations are benign
%, do not lead to atomic bias 
(\textit{see}  \textbf{RQ1}/\autoref{tab:error-rates}).  
%We note that this is computed for only intersectional mutants with corresponding \textit{valid} atomic mutants. 

\noindent
\textit{6.) Bias-inducing Original inputs and Mutations:} We report the number of 
%atomic/intersectional 
mutations and original inputs that induce bias (\textit{see} \textbf{RQ4}). 

\noindent
\textit{7.) False Positives:} \revise{
The number of biases induced by discarded/invalid mutants,
%/inputs that are invalid,  
i.e.,  
%bias-inducing inputs that 
inputs that \textit{fail} dependency invariant check 
%among the discarded inputs 
(\textit{see} \textbf{RQ3})\footnote{Users are saddled with false positives, i.e., cases that do not conform to the parse tree of the original input.}.
% We note that mutation based approaches without validity check (e.g., MT-NLP~\cite{ma2020metamorphic}) may produce such false positives. 
}

\noindent
\textbf{Bias dictionary:} We automatically extracted it from the ``Social Bias Inference Corpus'' (SBIC)~\cite{sap-etal-2020-social} which contains 150K structured annotations of social media posts, covering a thousand demographic groups. Our dictionary contains curated semantically meaningful bias alternatives totalling \checknumber{
%eight (8) racial groups, four (4) gender groups and six (6) body attributes.  %groups of words for race,  gender and body attributes,  respectively. 
%Overall, our dictionary contains 
% composed of 
116 words for ``race'',  230 words for ``gender'' and 98 words for ``body'' attributes. }
%For instance,  
%%in \autoref{fig:illustrative-example},  
%the word pair \checknumber{\{``man'':``woman''\}} is among the pairs stored in the dictionary for the ``race'' attribute.  
We also manually inspected and validated the list of word pairs for each attribute.  
%Our artifact provides contains the dictionary. 
% 
%As illustrated in
%
% in order 
% to distinguish between undefined groups based on the local context of the cases in the datasets (Europe and US). For instance, in \circled{R$_2$}, the "Majority" group comprises majority groups found in Europe and US, such as "white," while \circled{R$_1$} consists of defined groups like "African" with a specific subgroup, such as "Nigerian."}
%
%using the lists with the others of the same attribute, but only the ones meaningful. 
%\todo{sudiptac: I do not understand the following sentence} 
%\todo{@Sudipta: is it clearer?}
% \revise{
% This is to '

%\vspace{0.2cm}
%\noindent
%\textbf{Input Semantics:} \todo{address how we ensure semantic consistency in the approach and evaluation} 
%\todo{manual inspection of dictionary pairs,  dependency checking,  grammarly,  manual inspection of resulting inputs?} 
%We also ensure that only semantically equivalent words are paired.  As an example,  the word ``herself'' can only be replaced by ``himself'  during gender testing,  and \textit{not} by any other semantically wrong ``male-related word'' such as ``him'', ``man'',``husband'', etc. 
%% }
%%\revise{
%%For example ``himself" was only associated with``herself" 
%%for the lists ``male" and ``female" of the gender attribute,  
%%while each gender-biased jobs of ``male" can be paired with any of the other list ``female".  
%%%Overall,  in our experiments,  we generated 3996 pairs,  1012 pairs and 786 pairs for race,  gender and body respectively.
%% }

\noindent
\textbf{Prompt Configuration:} 
%\label{sec:prompt-config}
%\todo{describe the few-shot prompt, token based truncation etc}
\revise{The max prompt/input size of the fine-tuned models is 512 tokens, 
while
% thus all queries for both the original text and mutated texts fed to these models are truncated to 512 tokens. Meanwhile, 
the two pre-trained models (GPT3.5 and Llama2) support up to 4096 tokens.}
%Thus,  these models have slightly different sets of inputs since the number of resulting mutants fed to them depends on the token size they support. 
% }
%\revise{
To mitigate model hallucination and randomness,  we set the temperature for each query to zero for Llama2.  We also turn off sampling and consistently use the same examples to ensure results as deterministic as possible.
We employ few-shot prompting to query the pre-trained models (GPT3.5 and Llama2).  
%We have chosen these prompting method due to its 
Few-shot prompting  performed best in our preliminary evaluation,  e.g.,  in comparison to zero-shot prompting. 

\begin{table*}[tb!]
\begin{center}
\caption{\centering 
\revise{Details of System Prompts and Questions for each dataset
}}
%\vspace{-1.5em}
% \vspace{-\baselineskip}
{\scriptsize %\footnotesize
  \bgroup\def\arraystretch{1.3}
\begin{tabular}{@{}|l|c|c|@{}} %r|r|}
\hline
% \endfirsthead

\hline
\textbf{Dataset} 
% & \textbf{Labels} & \textbf{\#labels} \\
& \textbf{System Prompt} & \textbf{Question} \\
% & \textbf{Question} & \textbf{Text1} & \textbf{Answer1} & \textbf{Text2} & \textbf{Answer2} & \textbf{Text3} & \textbf{Answer3} \\
\hline
% \endhead

\multirow{2}{*}{\textbf{IMDB}}
&\multirow{2}{*}{\shortstack[c]{
Your task is to determine whether the emotional tone \\ 
or attitude in the provided Internet Movie Database
(IMDB) review is positive or negative. }} 
&\multirow{2}{*}{\shortstack[c]{
Is the expressed  sentiment conveyed in the \\
text positive or negative?
}} \\
& & \\
\hline

\multirow{3}{*}{\textbf{ECTHR}}
& \multirow{3}{*}{\shortstack[c]{
Identify any violations of the specified European \\
Court of Human Rights (ECTHR)   articles 
([LABELS]) based on the provided factual paragraph. \\
Only  answer  with the following format: 
Answer: Article X, Article X, Article X ... }} 
&\multirow{3}{*}{\shortstack[c]{
Which articles of the European Court of Human \\ Rights are violated in this case, if any? 
}} \\
& &  \\
& &  \\
\hline

\multirow{2}{*}{\textbf{EurLex}}
& \multirow{2}{*}{\shortstack[c]{
Given a document from the EUR-Lex portal
containing European Union (EU) legislation predict \\
its relevant EuroVoc labels among [LABELS]. }}
& \multirow{2}{*}{\shortstack[c]{Which are the relevant EuroVoc labels of the \\ provided European Union legislation?}} \\
& &  \\
\hline

\multirow{3}{*}{\textbf{Ledgar}} & \multirow{3}{*}{\shortstack[c]{
Identify the single main theme among the
specified labels ([LABELS]) within a provided \\
contract provision in the Electronic Data
Gathering, Analysis, and Retrieval database \\
sourced from the US Securities and Exchange Commission.}} & 
\multirow{3}{*}{\shortstack[c]{Which label represents the single main \\ topic of the provided contract provision?}} \\
& &  \\
& &  \\
\hline

\multirow{2}{*}{\textbf{SCOTUS}} &  \multirow{2}{*}{\shortstack[c]{
Identify the single main issue area ([LABELS]) in the
provided US Supreme Court opinions. Only answer with \\
the relevant area following the format: Answer: X.}} & \multirow{2}{*}{\shortstack[c]{Which single-label represents the relevant \\ issue area of the provided court opinion?}} \\ 
& &  \\
\hline

\end{tabular}
\egroup
}
\label{tab:system-prompt-and-question}

\end{center}
\end{table*}

\revise{\autoref{tab:system-prompt-and-question} presents the system prompts and corresponding prompt questions for each dataset. These prompts ensure the model understands the specific tasks associated with each dataset and generates responses in a consistent and interpretable format. Furthermore, the few-shot examples included in the prompts, detailed in \autoref{tab:few-shot-examples}, provide context and guidance for the model to understand and correctly perform the task.}

\noindent
\revise{\textbf{System Prompts and Questions} for each dataset are carefully chosen to define the task explicitly. For instance:
\begin{itemize}
\item \textbf{IMDB}: The system is prompted to determine the sentiment of a review (positive or negative). The question explicitly asks about the sentiment conveyed in the text.
\item \textbf{ECTHR}: The system is tasked with identifying any violations of specific articles of the European Court of Human Rights. The prompt and question focus on evaluating factual paragraphs to predict the relevant articles violated.
\item \textbf{EurLex}: The system predicts relevant EuroVoc labels for European Union legislation.
\item \textbf{Ledgar}: The system is prompt to identify the single main theme within a contract provision.
\item \textbf{SCOTUS}: The task involves predicting the single main issue area within US Supreme Court opinions, with a focus on providing concise, label-specific answers.
\end{itemize}}

\noindent
\revise{\textbf{Few-shot examples} enhance the model’s understanding by providing concrete examples of how to approach the tasks. These examples illustrate the expected input-output relationships and guide the model's behavior. For each dataset, we include representative few-shot examples in \autoref{tab:few-shot-examples}, made to ensure task alignment and simple answers with the labels of the corresponding datasets.}
%For these models, the prompt contains \revise{three} examples of questions and valid responses, then a new query to respond to.  
%\revise{
%In addition,  we  provide the LLMs (\revise{GPT3.5 and Llama2}) with an additional system command. This is to ensure that the LLM’s response is amenable to automated analysis.  In particular, we provide
%an example of the system command and
% our 
% prompt in \todo{Figure XXX}.  
%\todo{Figure XXX} presents 
% For LLMs that do not support a system command, we prepend the same statement to all of the queries before feeding it to the respective LLMs.
%This is done for all valid generated inputs and original inputs.  For the LLMs supporting system command \todo{i.e., GPT3.5},  we set system prompt to indicate the aforementioned system command \todo{XXXX}.
%}
%\todo{token size,  tokenizers and truncation}
% \revise{
%We note that t
%Due to space limitations,  w
%We provide d
% Details and e
Examples of our prompts are provided in our artifact and the HuggingFace space.\footnote{https://huggingface.co/spaces/Anonymous1925/Hinter} 
%}

\begin{table}[H]
\begin{center}
\caption{\centering 
\revise{Details of Few-shot examples in Prompts
}}
%\vspace{-1.5em}
% \vspace{-\baselineskip}
{\scriptsize %\footnotesize
  \bgroup\def\arraystretch{1.3}
\begin{tabular}{@{}|c|c|c|c|@{}} %r|r|}
\hline
% \endfirsthead

\hline
\multirow{2}{*}{\textbf{Dataset}} & \multirow{2}{*}{\textbf{ID}} & \multicolumn{2}{c|}{\textbf{Few-shot examples in prompts}} \\ \cline{3-4}
&  & \textbf{Examples} & \textbf{Answer} \\
% & \textbf{Example 2} & \textbf{Answer 2} & \textbf{Example 3} & \textbf{Answer 3} \\
\hline
% \endhead

\multirow{3}{*}{\textbf{IMDB}} &  (1) & \shortstack[c]{In 1970, after a five-year absence, Kurosawa made what
would be his first film in color...} &  Positive \\
\cline{2-4}
& (2) & Don't waste your time on this film... & Negative \\
\cline{2-4}
& (3) & I know technically this isn't the greatest TV show ever... & Positive \\
\hline

\multirow{3}{*}{\textbf{ECTHR}}  & (1) & 5. The applicant was born in 1960 and lives in Gaziantep... & Article 6 \\ 
\cline{2-4}
& (2) & 7. The applicant was born in 1957 and lives in Nicosia... & Article 8, Article 1 of Protocol 1 \\
\cline{2-4}
& (3) & 4. The applicant was born in 1941 and lives in Sosnowiec... & None \\

\hline
\multirow{7}{*}{\textbf{EurLex}} & \multirow{3}{*}{(1)} & \multirow{3}{*}{COMMISSION REGULATION (EC) No 1588/1999...} & \multirow{3}{*}{\shortstack[c]{EU finance, tariff policy,\\ agricultural activity, fisheries, \\ industrial structures and policy}} \\
& & & \\
& & & \\
\cline{2-4}
& (2) & COMMISSION REGULATION (EC) No 854/2007... & \shortstack[c]{trade policy, beverages and sugar} \\ 
\cline{2-4}
& \multirow{3}{*}{(3)} & \multirow{3}{*}{COMMISSION REGULATION (EEC) No 1481/83...} & \multirow{3}{*}{\shortstack[c]{tariff policy, natural and \\ applied sciences, means of \\agricultural production}} \\
& & & \\
& & & \\
\hline

\multirow{3}{*}{\textbf{Ledgar}} & (1) & Any waiver of a provision of this Agreement... & Waivers\\ \cline{2-4}
& (2) & This Agreement has been duly executed and delivered... & Enforceability \\ \cline{2-4}
& (3) & \shortstack[c]{Any written notices required in this Agreement may be made by personal delivery...} & Notices \\
\hline

\multirow{4}{*}{\textbf{SCOTUS}} & \multirow{2}{*}{(1)} & \multirow{2}{*}{\shortstack[c]{362 U.S. 58 80 S.Ct. 612 4 L.Ed.2d 535...Having considered the briefs and oral arguments \\submitted by both sides, the Court makes the following disposition of these matters:}} & \multirow{2}{*}{Civil Rights} \\ 
& & & \\

\cline{2-4}
& (2) & \shortstack[c]{344 U.S. 443 73 S.Ct. 437 97 L.Ed. 469 Bennie DANIELS and Lloyd Ray Daniels... Two general problems are these:...} & Criminal Procedure \\ 
\cline{2-4}
& (3) & \shortstack[c]{358 U.S. 64 79 S.Ct. 116 3 L.Ed.2d 106...The motion for leave to file bill of complaint is denied ...} & Interstate Relations \\

\hline

\end{tabular}
\egroup
}
\label{tab:few-shot-examples}

\end{center}
\end{table}

\noindent
\textbf{Implementation Details and Platform:}
%\revise{
\approach and our data analysis were implemented in 
about 
%he tests took less than 
\checknumber{6} KLOC of Python.  
%\revise{
\revise{
All experiments were conducted on a compute server with 
%The experiments were conducted on different machines for input generation and testing, from aion and iris clusters \todo{ref}. Input generation was performed using 20 nodes in the Aion cluster, with each node equipped with 2x AMD Epyc ROME 7H12 CPUs (64 cores, 2.6 GHz) and 256 GB DDR4 RAM. Generating 512 and 4096 long tokens took 3 days (16K CPU days) and 21 days (110K CPU days), respectively. For testing, one node with 40 threads was used for BERT model testing, while 25 nodes with 2 threads each were used for GPT-3.5 and Llama2 testing. Each node was equipped with 
four Nvidia Tesla V100 SXM2 GPUs, two Intel Xeon Gold 6132 processors (2.6 GHz), and 768 GB of RAM. 
% \revise{
% All experiments took approximately 18 days 
% five (5) days, one (1) day, 11 days and one (1) day 
% for all datasets.} 
% Ecthr A/B, Scotus, Eurlex, Ledgar datasets, respectively.}
We provide the source code and experimental data for \approach:} 
% \begin{center}
\url{https://github.com/BadrSouani/HInter}    
% \end{center}

%BERT model testing, which used a maximum of 50 GB RAM, took approximately 4 days and 2700 GPU days. In contrast, GPT-3.5 and Llama2 testing took around 14 days and 2800 GPU days. Despite Llama2's higher resource consumption, the similarity in GPU days can be attributed to the testing of 16 BERT models and the evaluation of all mutants generated before the invariant check for biases.}
%  for the intersectional bias testing.
% }
%

\section{Results}
\label{sec:experimental_result}
%Let us 
%%In this section, w
%%We 
%discuss the 
%%results of our experiments evaluating the 
%performance of \approach. 

%\todo{group fairness discussion}

%\todo{fix RQ1 numbers to match \autoref{fig:bias-error-rate-intersectional} (updated)}

\begin{figure}[H]
\centering
\includegraphics[width=0.49\textwidth]{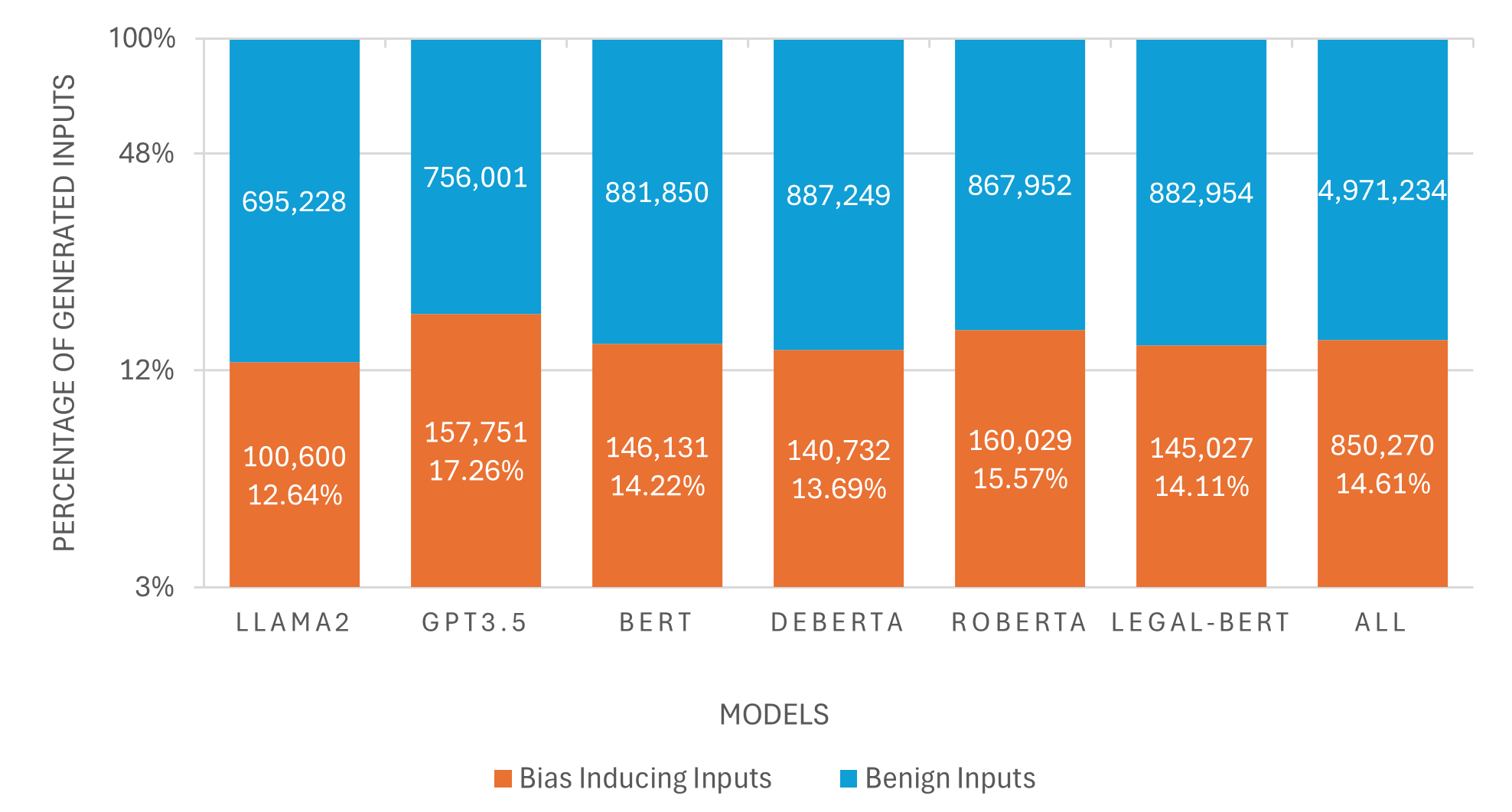}
\includegraphics[width=0.49\textwidth]{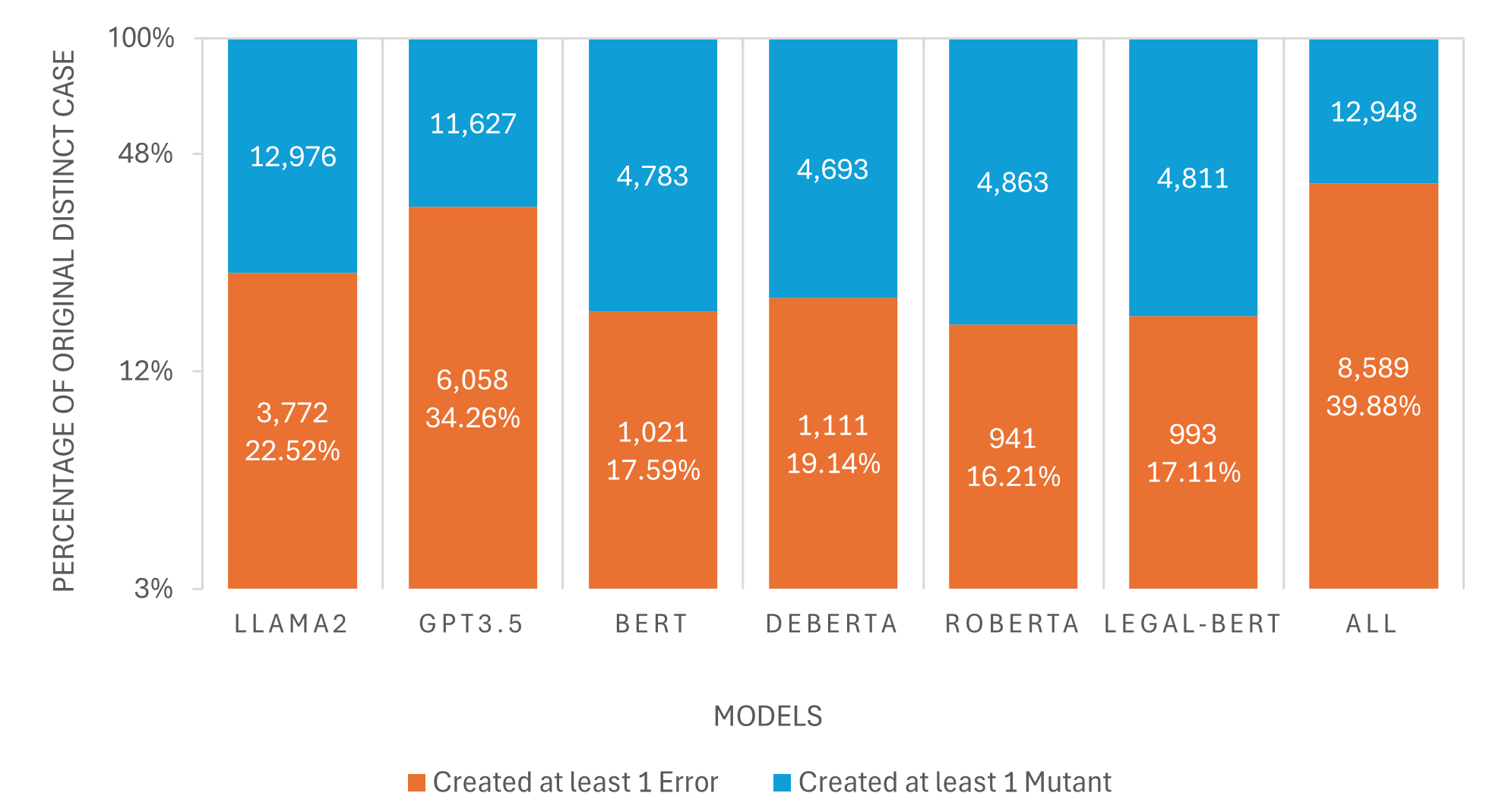}
{
\footnotesize
\quad \quad \quad \quad \quad \quad \quad \quad \quad 
(a) Intersectional  Bias Error Rate 
\quad \quad \quad \quad \quad \quad \quad \quad \quad  \quad \quad \quad \quad \quad
(b) Distinct Intersectional bias-inducing original inputs
\quad \quad
} 
% \vspace{-0.35cm}

\caption{\centering 
\revise{Intersectional bias error rate 
and the proportion of unique original cases that lead to intersectional bias 
%for all tested LLMs for Intersectional 
%bias testing 
%and Atomic bias testing.  
}}
\label{fig:bias-error-rate-intersectional}
\end{figure}

\noindent
\textbf{RQ1 \approach's Effectiveness:}
%\todo{Should we report the cases where the actual prediction is wrong? useful for group fairness,  TPR,  etc}
%
%\todo{For group fairness, we need to measure: \\
%(a) equal opportunity differences (EOD) that measures the differences in true positive rates between two groups;\\
%(b) average odd differences (AOD) that quantifies the differences in the average of true positive rates and false positive rates between two groups
%}
%
%\textit{\textbf{Motivation}:} 
%The aim of t
This experiment investigates the intersectional bias error rate exposed by \approach, 
%i.e.,  the proportion of generated inputs 
% by \approach 
%that leads to intersectional bias.  
%\revise{We also report 
the proportion of \textit{hidden} intersectional bias
%, i.e.,  the errors that are hidden behind atomic bias (similar to \autoref{fig:illustrative-example}). }
%\revise{
%We also investigate the proportion of all errors that lead to a hidden intersectional bias, i.e..,  
%}
%(or atomic) error.  
%Besides, we examine 
and the proportion of original inputs that induce \textit{at least} one intersectional bias when mutated. 
%\recheck{
%Finally, we report the intersectional group bias results. 
%}
%
%%\subsubsection{Intersectional Bias}
%
%
%\textit{\textbf{Settings}:} 
This experiment involves all five datasets and 
%LLM Model architectures and 
18 LLM models -- 
all models for the four legal datasets,  and the two (2) pre-trained LLMs (GPT3.5 and Llama2)  
%
%%\approach generated bias-prone inputs for all six datasets.  
%%For all 22 models,  w
%We fed \approach's generated inputs for the four legal datasets (i.e.,  SCOTUS,  ECTHR,
%%\_A,  ECTHR\_B, 
%LEDGAR,  EURLEX) to all \checknumber{18} LLM models,  i.e.,  the  \checknumber{16} fine-tuned legal LLM models (from LEXGLUE) and the two (2) pre-trained LLMs (GPT3.5 and Llama2).  Additionally,  we fed 
%%\approach's 
%the generated inputs for the IMDB dataset to the two (2) pre-trained LLMs,  i.e.,  GPT3.5 and Llama2\footnote{We used only the two (2) pre-trained models (GPT3.5 and Llama2) 
for the IMDB dataset\footnote{The 16 fine-tuned LLMs are inappropriate for IMDB, since they are specifically fine-tuned for the 4 legal datasets. }. 
 \autoref{tab:error-rates}   and 
 \autoref{fig:bias-error-rate-intersectional} present our findings. 

%the error rate of \approach for intersectional bias.
% and  bias.  

\begin{table}[H]
 \begin{center}
 \caption{\centering 
\revise{Details of Intersectional Bias Error Rates 
found by \approach. 
``\textbf{\protect\circled{I}}''  means ``Intersectional Bias Error Rate'' and ``\textbf{\protect\circled{H}}'' means ``Hidden Intersectional Bias Error Rate''}}
 {\tiny
   \bgroup\def\arraystretch{1.3}
 \begin{tabular}{@{}|c|c|c|c|l|c|c|c|c|c|c|c|@{}}
 \hline

\textbf{Datasets (Domain)} & \textbf{Tasks} & \textbf{Class.}  & \textbf{\#labels} & \textbf{Metric} &  \textbf{Llama 2} & \textbf{GPT 3.5} & \textbf{Legal-BERT} & \textbf{BERT} & \textbf{De-BERTa} & \textbf{Ro-BERTa}  & \textbf{Total} \\
 \hline

\multirow{2}{*}{\textbf{ECtHR (ECHR)}} & \multirow{2}{*}{Judgment Prediction} & \multirow{2}{*}{Multi-Label} & \multirow{2}{*}{10+1} & \textbf{\circled{I}} & \checknumber{51.74}  & \checknumber{59.03}  & 5.02 & 4.65 & 5.10 & 3.75 & 11.37 \\
 \cline{5-12}
 & & & & \textbf{\circled{H}} & (13.24) & (13.70) & (45.12) & (15.35) & (17.26) & (18.18) & (18.06) \\
 \hline

\multirow{2}{*}{\textbf{LEDGAR (Contracts)}} & \multirow{2}{*}{Contract Classification} & \multirow{2}{*}{Multi-Class} & \multirow{2}{*}{100} & \textbf{\circled{I}} & \checknumber{25.35} & \checknumber{7.04}  & 0.00 & 0.00 & 0.00 & 1.05  & 4.58 \\
 \cline{5-12}
 & & & & \textbf{\circled{H}} & (12.37) & (0.00) & (0.00) & (0.00) & (0.00) & (0.00)  & (12.24) \\
 \hline

\multirow{2}{*}{\textbf{SCOTUS (US Law)}} & \multirow{2}{*}{Issue Area Prediction} & \multirow{2}{*}{Multi-Class} & \multirow{2}{*}{14} & \textbf{\circled{I}} & \checknumber{12.32} & \checknumber{21.32} & 1.02 & 2.34 & 4.76 & 2.25 & 10.60 \\
 \cline{5-12}
 & & & & \textbf{\circled{H}} & (26.33) & (19.60) & (40.87) & (13.20) & (5.86) & (8.30) & (18.92) \\
 \hline

\multirow{2}{*}{\textbf{EURLEX (EU Law)}} & \multirow{2}{*}{Document Prediction} & \multirow{2}{*}{Multi-Label} & \multirow{2}{*}{100} & \textbf{\circled{I}} & \checknumber{89.62} & \checknumber{91.00} & 19.99 & 20.28 & 19.02 & 22.97 & 21.66 \\
 \cline{5-12}
 & & & & \textbf{\circled{H}} & (1.35) & (0.49) & (37.53) & (7.46) & (11.86) & (5.62) & (14.73) \\
 \hline

\multirow{2}{*}{\textbf{IMDB (Reviews)}} & \multirow{2}{*}{Sentiment Analysis} & \multirow{2}{*}{Binary} & \multirow{2}{*}{2} & \textbf{\circled{I}} & \checknumber{3.52} & \checknumber{7.12}  & N/A & N/A & N/A & N/A & 5.43 \\
 \cline{5-12}
 & & & & \textbf{\circled{H}} & (28.38) & (28.03)  & N/A & N/A & N/A & N/A & (28.13) \\
 \hline

\multirow{2}{*}{\textbf{All}} & \multirow{2}{*}{All} & \multirow{2}{*}{-} & \multirow{2}{*}{-} & \textbf{\circled{I}} & \checknumber{12.64} & \checknumber{17.26} & 14.11 & 14.22 & 13.69 & 15.57 & \checknumber{14.61} \\
 \cline{5-12}
 & & & & \textbf{\circled{H}} & (16.99) & (18.19) & (38.46) & (8.36) & (12.45) & (6.64) & \checknumber{16.62} \\
 \hline
   \end{tabular}\egroup}
 \label{tab:error-rates}   
\end{center}
\end{table}

\textit{\textbf{Results}:} We found that \textit{\checknumber{about one in seven (14.61\%)} inputs generated by \approach revealed intersectional bias. } 
\revise{\autoref{fig:bias-error-rate-intersectional}(a) shows that  
\approach exposed intersectional bias with \checknumber{12.64\% to 17.26\%} error rate across all models.
We also observed that \checknumber{16.62}\% of intersectional bias found by \approach are \textit{hidden}, i.e., their corresponding atomic test inputs do not trigger bias. This demonstrates the importance of intersectional bias testingin viz a viz atomic bias testing. 
}

\begin{figure}[H]
\centering
\includegraphics[width=0.49\textwidth]{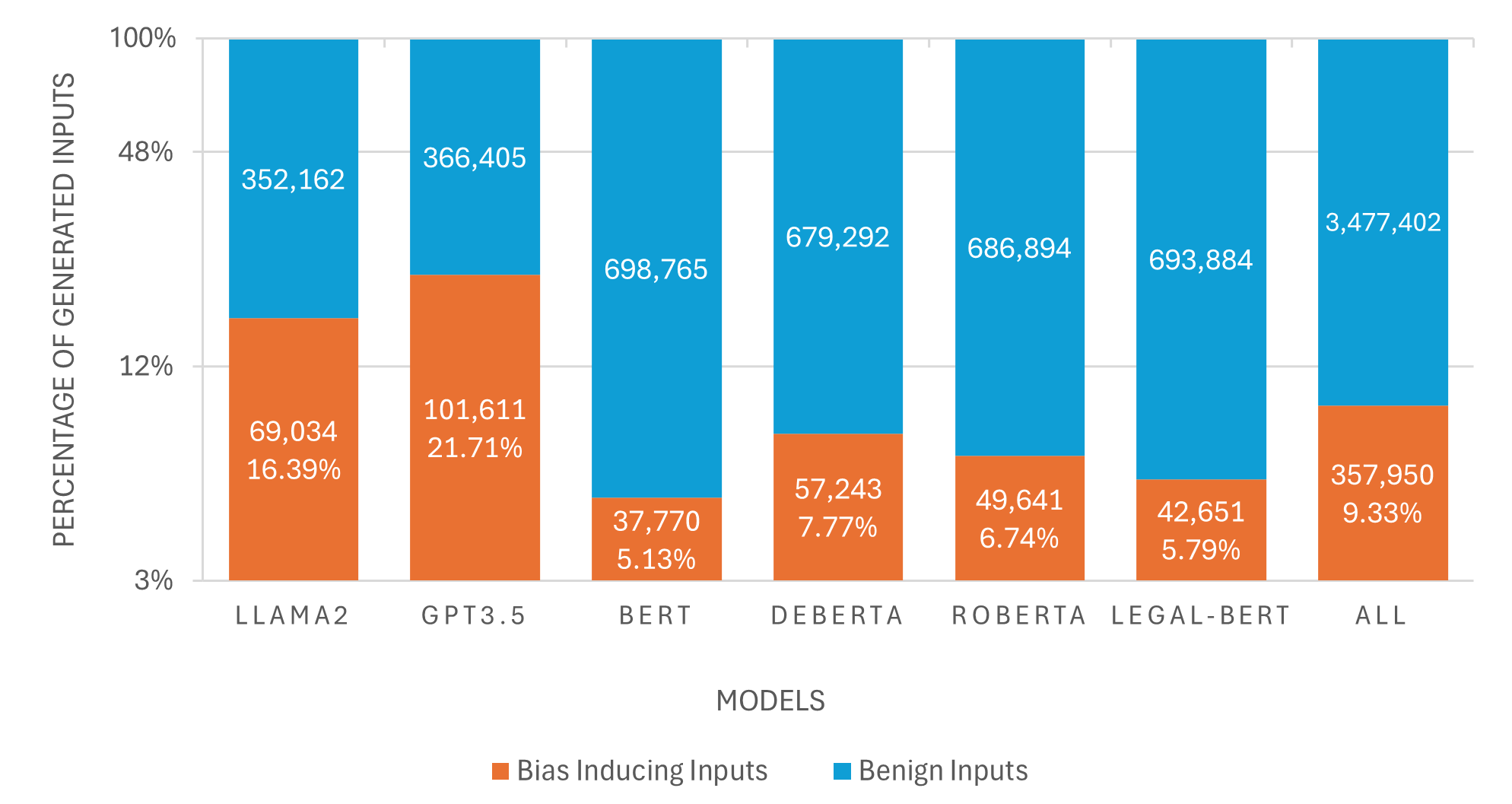}
\includegraphics[width=0.49\textwidth]{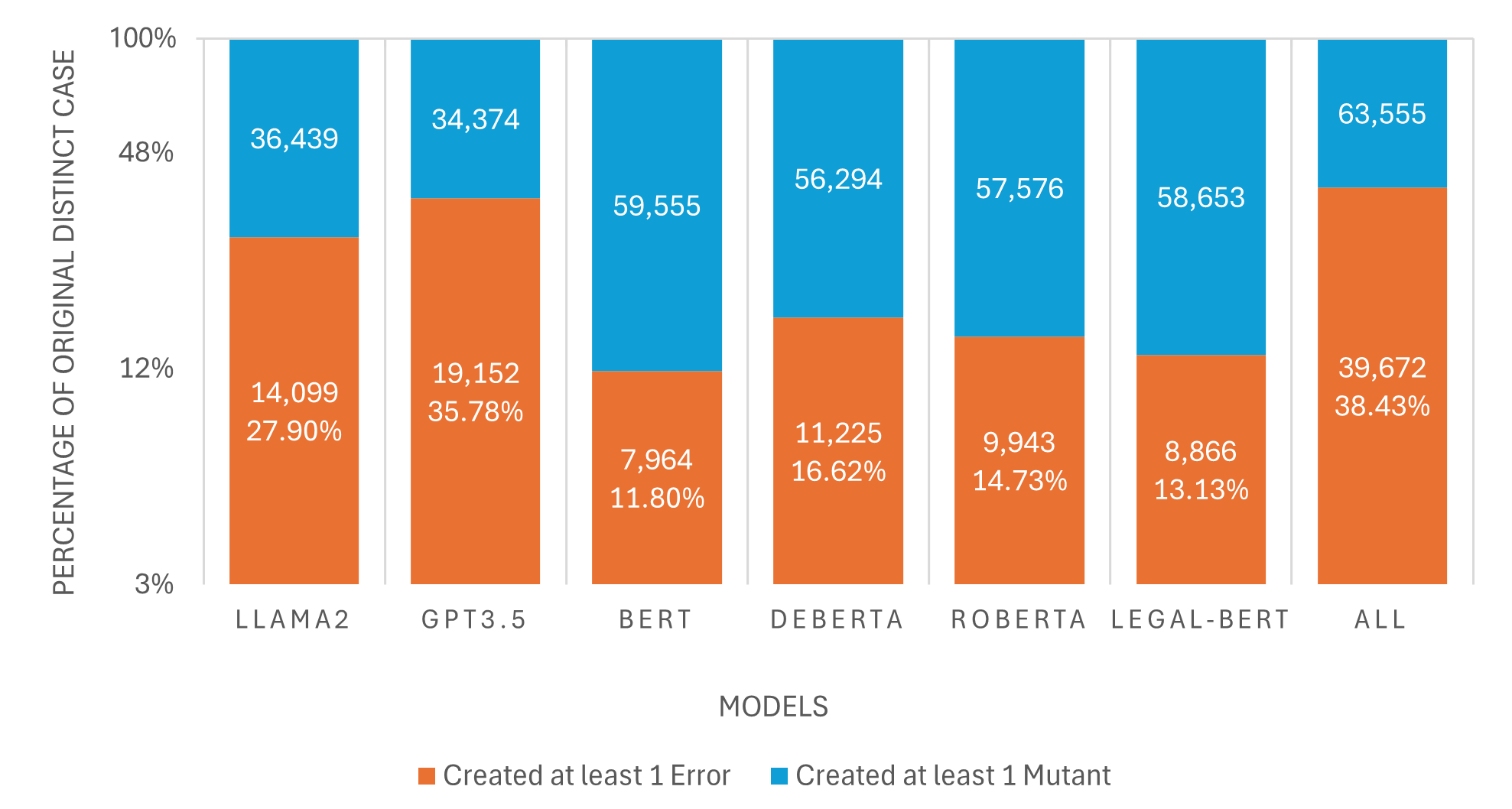}
{
\footnotesize
\quad \quad \quad \quad \quad \quad \quad \quad \quad 
(a) Atomic  Bias Error Rate 
\quad \quad \quad \quad \quad \quad \quad \quad \quad  \quad \quad \quad \quad \quad
(b) Distinct Atomic bias-inducing original inputs
\quad \quad
} 
% \vspace{0.5em}

\caption{\centering 
\revise{Atomic bias error rate 
and the proportion of unique original cases that lead to atomic bias 
%for all tested LLMs for Intersectional 
%bias testing 
%and Atomic bias testing.  
}}
\label{fig:bias-error-rate-atomic}
\end{figure}

Results show that GPT3.5 has the highest (\checknumber{17.26\%}) error rate,  while the Llama2 model has the lowest (\checknumber{12.64\%}) error rate.
% \revise{This results are shown in \autoref{fig:bias-error-rate-intersectional}.}
%This trend is consistent with the 
%results for the 
Inspecting the proportion of distinct original inputs that lead to intersectional bias in \revise{\autoref{fig:bias-error-rate-intersectional}(b)} , we observed that \textit{about} two in five (39.88\%) original inputs lead to at least one intersectional bias when mutated by \approach (\textit{see} \autoref{fig:bias-error-rate-intersectional}). 
% is consistent with 
%is is also consistent with the results for 
%shows that the proportion of original inputs that lead to intersectional bias follows a similar trend as the intersectional bias error rate across LLMs (\textit{cf.} \autoref{fig:bias-error-rate-intersectional}(a).) 
%This is 
%also 
%evident in 
%the proportion of original cases that lead to intersectional bias (\textit{see} \autoref{fig:bias-error-rate-intersectional}(b)).  
%We also observed that 
%Generally,  
%%Besides, 
%we observed that 
%the BERT-based models (BERT, DEBERTa, ROBERTa and Legal-BERT) have a lower intersectional bias error rate in comparison to the pre-trained models (GPT3.5 and Llama2). 
%\footnote{The observation that fine-tuned BERT-based models have a lower error rate than the pre-trained models 
%also holds for atomic bias. Indeed,  this difference 
%is more pronounced in atomic bias testing (\textit{see} \autoref{fig:bias-error-rate}(b)). }.    
%We attribute the performance of the BERT-based models to the effect of model fine-tuning.
\revise{Figure~\ref{fig:bias-error-rate-atomic}(a) illustrates the atomic bias error rate across the models. GPT-3.5 shows the highest percentage of bias-inducing outputs at 21.71\%, significantly above the overall model average of 9.33\%.}
\revise{Figure~\ref{fig:bias-error-rate-atomic}(b) represents the proportion of distinct original inputs that resulted in generating at least one mutant with atomic bias errors. GPT-3.5 again presents the highest susceptibility, with 35.78\% of original inputs producing bias-inducing mutants. Overall, approximately 38.43\% (more than one-third) of original inputs led to errors.}
Overall,  
%our evaluation 
results 
%These results 
show that \approach is effective in exposing (hidden) intersectional bias across tested models.

% instances in LLMs.   

%\todo{discuss hidden intersectional bias error rate and Table 1}

\begin{result}
%\revise{
\approach is effective in exposing (hidden) intersectional bias: 
%\checknumber{One in seven (14.61\%)} 
\checknumber{14.61\%} of the inputs generated by \approach exposed intersectional bias, and 
\checknumber{16.62\%} of the intersectional bias errors exposed by \approach are hidden.
% to atomic bias testing. 
\end{result}

%\subsubsection{Hidden Intersectional Bias}
%\textit{\textbf{Settings}:}
%In this experiment, .... 
%
%\textit{\textbf{Results}:}
%
%\begin{result}
%\todo{X\%} of discovered intersectional bias are hidden (concealed) during atomic bias testing. 
%%}
%\end{result}

%
%\subsubsection{Intersectional Group Bias}
%\textit{\textbf{Settings}:}
%This experiment ... 
%
%\textit{\textbf{Results}:}
%
%\begin{result}
%\todo{XXX}
%\end{result}

%\subsubsection{Unique Original Cases}
%
%\textit{\textbf{Motivation}:}
%
%\textit{\textbf{Settings}:}
%
%\textit{\textbf{Results}:}
%
%
%\begin{result}
%\todo{XXX}
%\end{result}
%Comparison to Atomic Bias Testing}
%vs.  Intersectional Bias}

%This experiment investigates the importance and uniqueness of intersectional bias testing. In particular, w

%\begin{figure}[tb!]
%\centering
%\includegraphics[width=0.5\textwidth]{figures/piechart_atomic.png}
%
%\caption{\centering Correspondence of Atomic Bias inducing Mutants regarding Intersectional Bias inducing Mutants 
%}
%\label{fig:pie-chart-atomic}
%\end{figure}
%
%\begin{figure}[tb!]
%\centering
%\includegraphics[width=0.5\textwidth]{figures/piechart_intersectional.png}
%
%\caption{\centering Correspondence of Intersectional Bias inducing Mutants regarding Atomic Bias inducing Mutants 
%}
%\label{fig:pie-chart-intersectional}
%\end{figure}

\begin{figure}[H]
\centering
\includegraphics[width=0.49\textwidth]{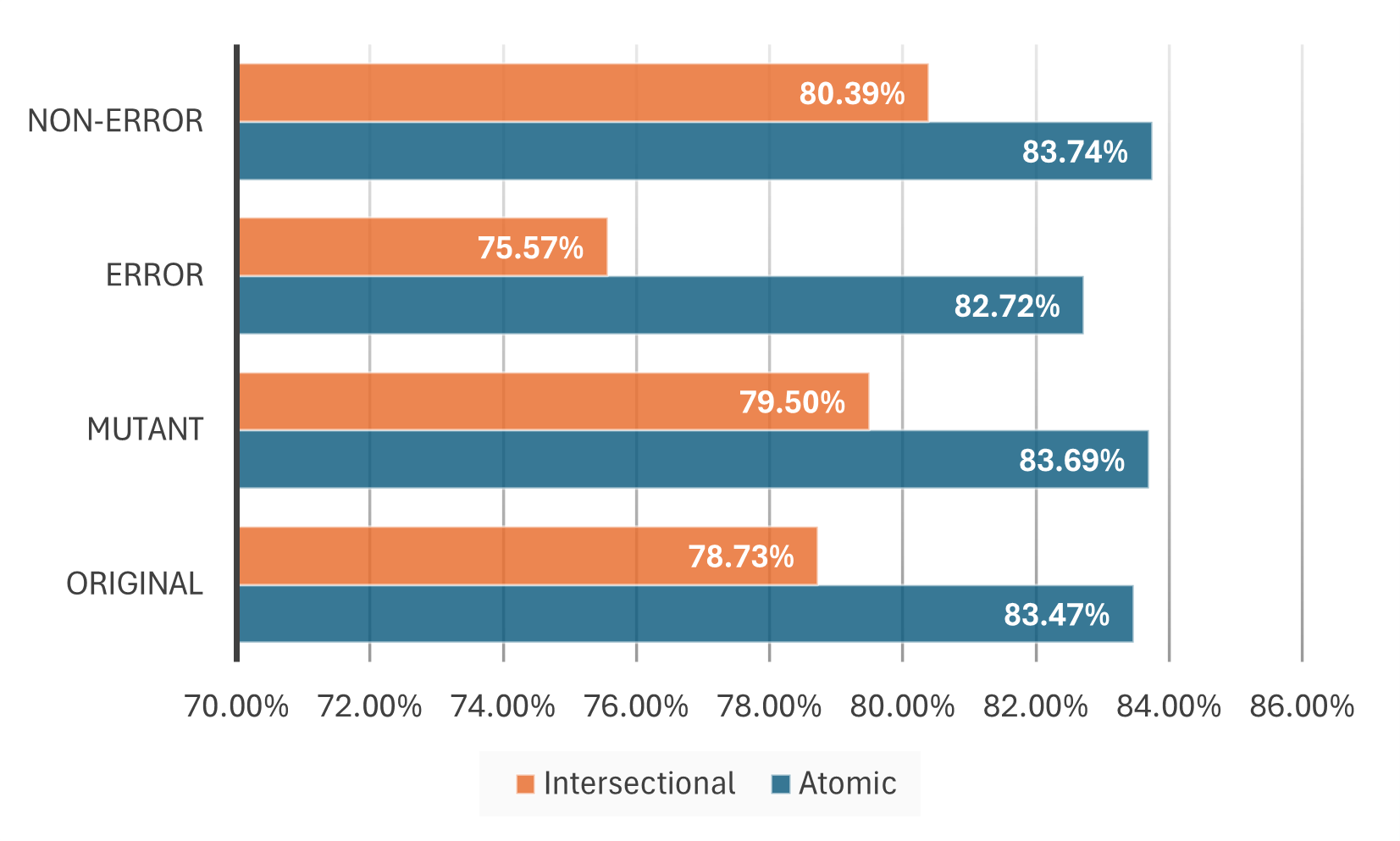}
%\includegraphics[width=0.49\textwidth]{figures/Validity_sim_vs_notsim.png}
%
% \quad \quad 
% (a) \approach's Generated Inputs 
% \quad \quad \quad \quad \quad \quad 
% (b) Valid Inputs vs.  Discarded Inputs
%  \quad \quad \quad \quad 
%\vspace{-0.8 em}
\caption{\centering 
Grammatical Validity of the inputs generated by  \approach, showing Grammarly's weighted mean score 
%reported by   
%\approach 
for the original (human-written) text vs the \approach's generated inputs (mutants). 
%\todo{Should we remove remove atomic results from this chart?}
%for 
%showing the validity of 
%(a) All Inputs generated by \approach 
%and 
%(b) Valid Inputs versus  
%%Valid Inputs vs.  
%Discarded Inputs that passed or failed \approach's invariant check.
}
\label{fig:grammarly-results}
\end{figure}

\noindent
\textbf{RQ2 Grammatical Validity:}
%\textit{\textbf{Motivation}:}
This experiment 
%aims to 
evaluates the \textit{grammatical validity} of the inputs generated by \approach in comparison to 
%versus the 
(human-written) original texts. 
%inputs. 
% texts.  
%
%\textit{\textbf{Settings}:}
%To this end,  w
\revise{Grammatically correct inputs help isolate the model's biases from potential errors introduced by ungrammatical structures, leading to more accurate detections of biases. By comparing original inputs with their mutated versions using tools like Grammarly~\cite{grammarly}\footnote{Grammarly is a popular tool for checking the spelling,  grammar and errors in English text. }, we can verify that the generated inputs maintain grammatical integrity, and that any observed biases are due to the model's processing rather than input anomalies.}
We randomly sampled inputs from the legal datasets\footnote{We employ the legal dataset due to their complexity and longer text in comparison to the IMDB dataset. } and fed them to Grammarly to compute their grammatical validity.
%\footnote{
%We randomly sampled \checknumber{100K} inputs across all settings 
%due to the huge computational cost of running Grammarly.  This experiment took \todo{XXX} about CPU days on our machine. }.  
This experiment involves about \checknumber{68K} mutants for atomic and intersectional bias testing and their (about \checknumber{10K}) corresponding original inputs.  Our sampling aims to balance the set of mutants and originals across sensitive attribute,  datasets
and 
mutation type
%s of mutants 
%as well as 
(i.e.,  atomic and intersectional.) 
%, as well as the three  settings for each.  
%We aimed to 
%\footnote{For instance,  
%we sampled about  \checknumber{3,769} original text (with 
%\checknumber{1502,  1351 and 1857} inputs)
%%4,820 original inputs 
%and their corresponding 
%%\checknumber{13,939,  16,394, and  16,280,} atomic mutants 
%\checknumber{42,071} mutants (\checknumber{14834, 13577 and 13660}) 
%for intersectional bias (body X race, body X gender and race X gender, respectively). }.
\autoref{fig:grammarly-results}  presents our results.  
%the grammatical validity of the inputs generated by \approach. 
%across originals,  mutants and error or non-errot 

%\todo{conflict between \autoref{fig:grammarly-results} and the discussion, one of them is incorrect}

\textit{\textbf{Results}:}
We observed that
\textit{the inputs generated by \approach are 
%similarly 
%\todo{X\%} 
as valid as the original human-written text}.
\autoref{fig:grammarly-results} shows that 
\textit{the inputs generated by \approach have similar grammatical scores as the original (human-written) text} for both atomic and intersectional bias testing campaigns. 
% ( 78.69\%  vsersus .78.98\%).   
%In particular,    
For intersectional bias testing,  
%\autoref{fig:grammarly-results} shows that 
the weighted mean score of the original text (\checknumber{78.73\%}) is similar to that of the  inputs generated by \approach (\checknumber{79.50\%}).  
%%%%% // removed to delineate inter and atomic 
%Likewise,  the scores for atomic testing are similar for both the original text and \approach's generated inputs (\checknumber{82.49\%} versus \checknumber{82.80\%}).  However, we observed that the grammatical validity of atomic bias testing is higher than for intersectional bias testing for both the original inputs (\checknumber{78.69\% versus 82.49\%}) and \approach's generated inputs (\checknumber{78.98\% versus 82.8\%}).  We attribute this difference to the higher complexity of the inputs employed in intersectional bias testing since they need to contain instances of multiple (at least two) sensitive attributes.  
%Examining the generated inputs and original texts that led to bias (error), 
%or not (non-error),  
We also found that \textit{the grammatical validity of error-inducing inputs is slightly lower than that of benign,  non-error-inducing inputs} for both the original inputs and \approach's generated input
(e.g.,  \checknumber{82.72 vs 83.74} for atomic bias testing).  
%We note that d
%Despite this, %ese observations, 
% about (non-)error-inducing inputs and ,  
The Grammarly scores for generated inputs (mutants) remain similar to their corresponding original inputs (\textit{see} \autoref{fig:grammarly-results}).  
%Overall, t
These results show that \approach generates grammatically valid inputs and preserves the grammatical validity of the original input.  
%We attribute this result to the invariant check of \approach, which ensures that resulting inputs have a similar dependency tree as the original text.  

\begin{result}
%\revise{
\approach preserves the grammatical validity of the original (human-written) inputs: %Its
\approach's 
generated inputs are as grammatically valid as the original (human-written) texts.  %}
\end{result}

%\begin{figure}[!tb]
%\includegraphics[width=0.8\columnwidth]{figures/InterAtomVD.png}
%\caption{\centering Repartition of the Valid and Discarded Intersectional and Atomic Mutants for  BERT, Llama2 and GPT3.5 models}
%\label{fig:InterAtomicVD}
%\end{figure}
%
%\begin{figure}[!tb]
%\includegraphics[width=0.8\columnwidth]{figures/InterVD.png}
%\caption{\centering Repartition of the Valid and Discarded Intersectional Mutants for BERT, Llama2 and GPT3.5 models}
%\label{fig:InterVD}
%\end{figure}
%
%\begin{figure}[!tb]
%\includegraphics[width=0.8\columnwidth]{figures/AtomicVD.png}
%\caption{\centering Repartition of the Valid and Discarded Atomic Mutants for BERT, Llama2 and GPT3.5 models}
%\label{fig:AtomicVD}
%\end{figure}

\noindent
\textbf{RQ3 Dependency Invariant}

%\todo{report false (and true) positives of generic mutations -- false bias errors since inputs are invalid}

%\todo{remember false (and true) positives are only for BERT models }

%\noindent
%\textit{\textbf{
%:}}
%We found 

% because they do not validate the validity of mutants w.r.t. to the original inputs. 

%discarded inputs vs.  valid) inputs. 
%In particular,  we examine the proportion of generated inputs that fail the invariant check (aka \textit{discarded} mutated inputs) or pass it 
%the invariant check 
%(aka \textit{valid} inputs).  
%In addition,  we examine the grammatical validity of the discarded inputs versus   \approach's generated valid inputs that pass the invariant check.  
%This 
% to its effectiveness.  
%

\begin{figure}[H]
\centering
\includegraphics[width=0.32\textwidth]{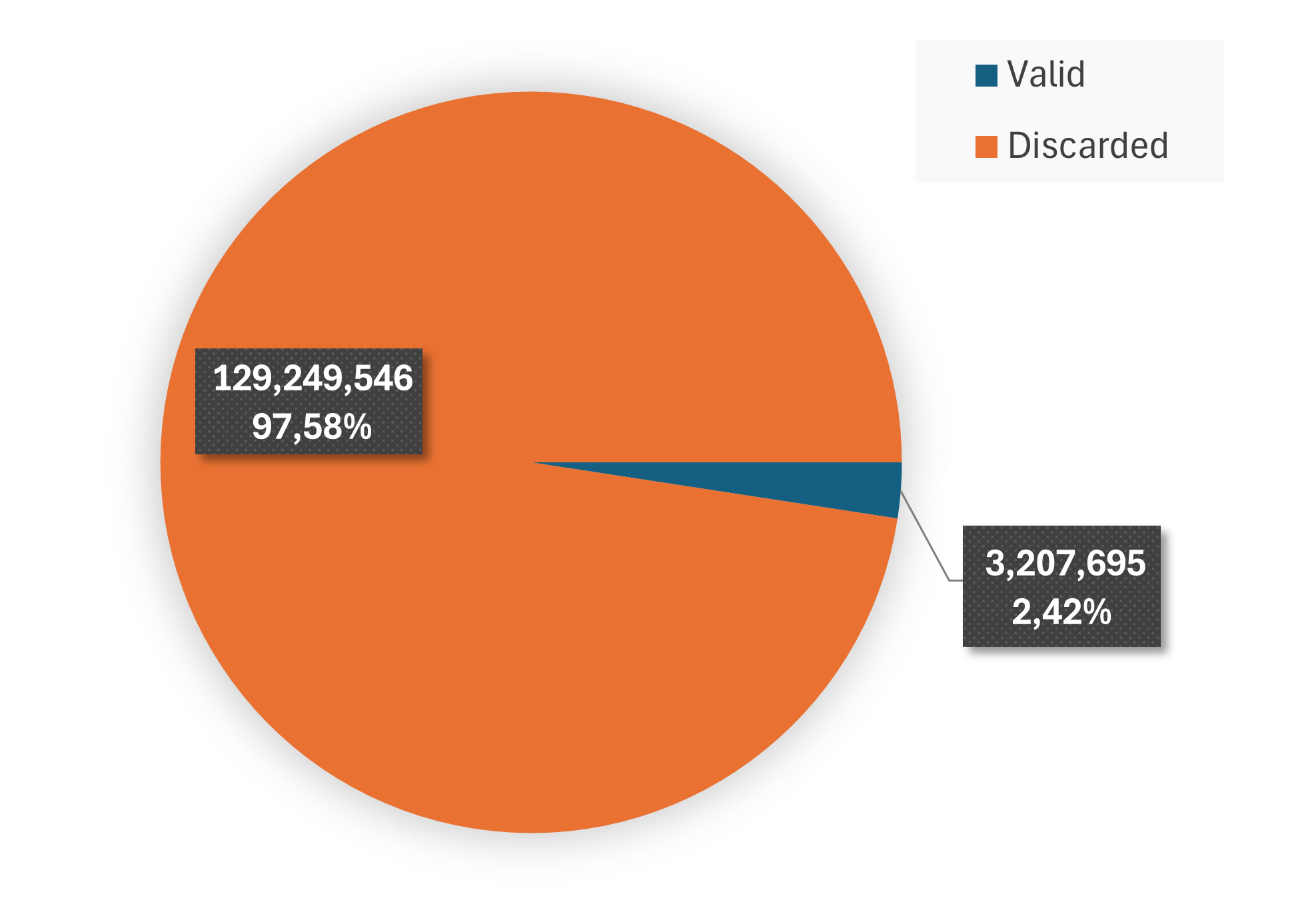}
\includegraphics[width=0.32\textwidth]{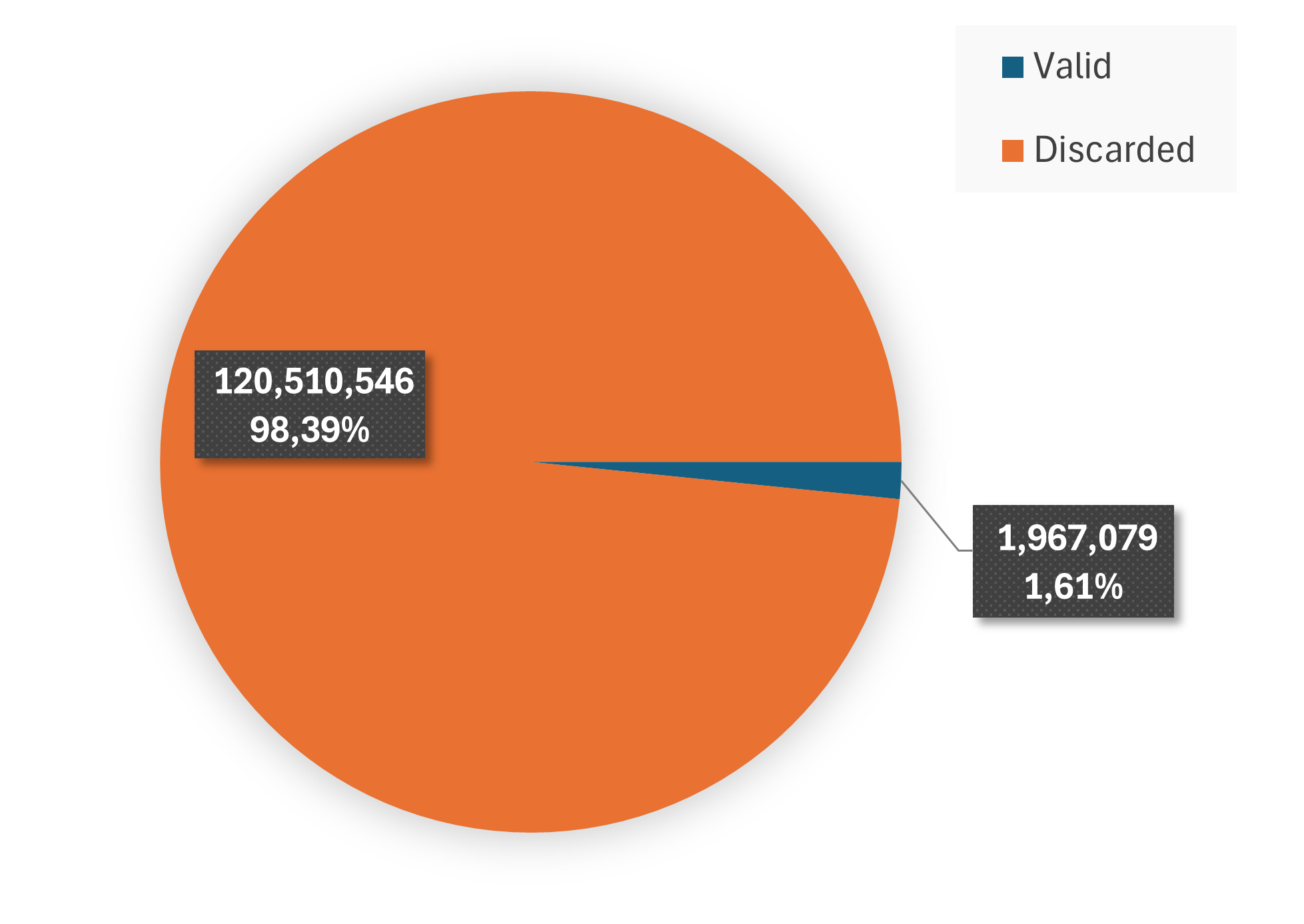}
\includegraphics[width=0.322\textwidth]{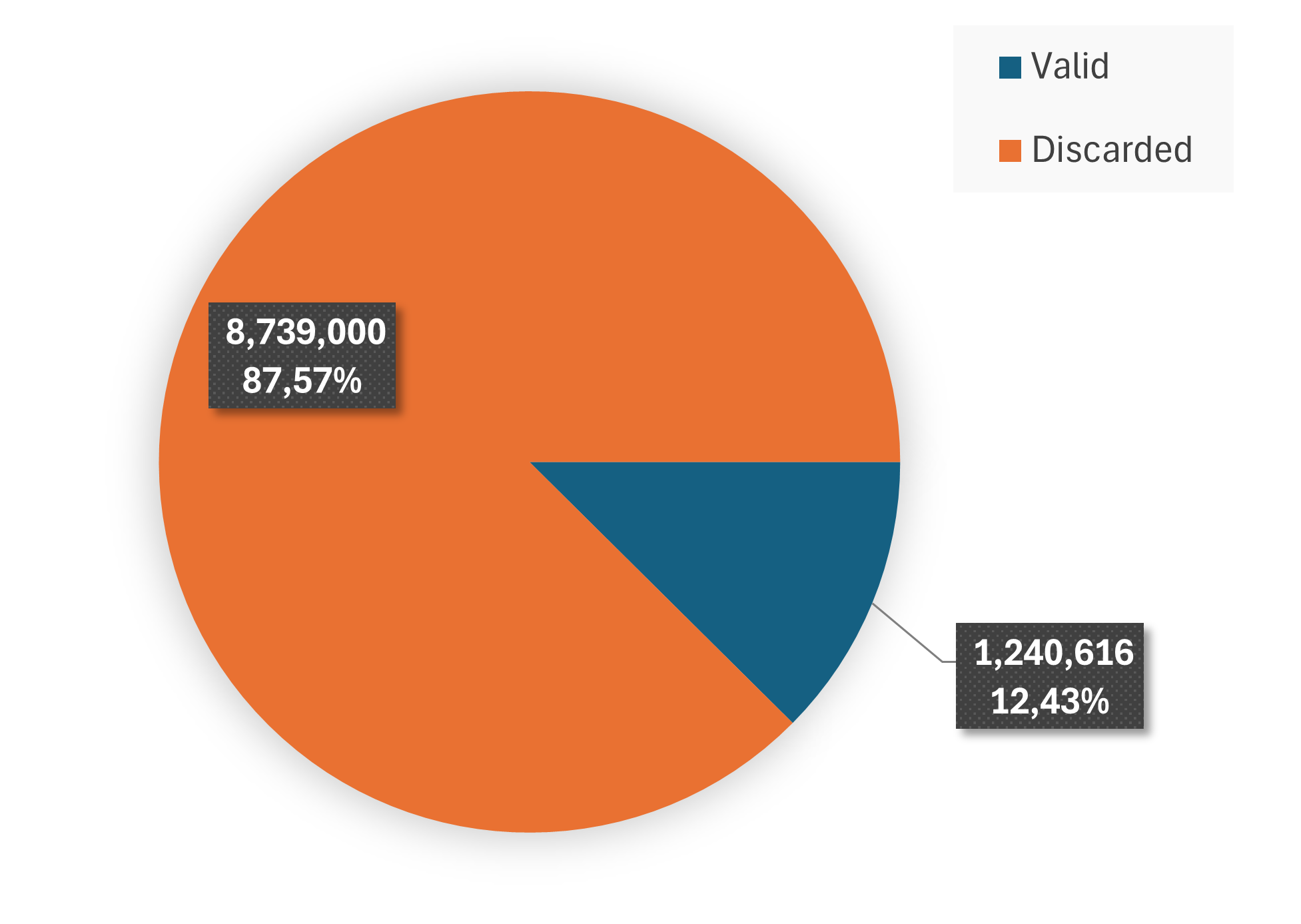}
%\vspace{-0.5em}
 \quad \quad (a) 
All Inputs 
\quad \quad \quad \quad \quad \quad (b) 
Intersectional  Inputs \quad \quad \quad \quad (c) Atomic Inputs  \quad \quad
%\quad \quad \quad \quad
%\vspace{-0.5em}
\caption{\centering 
%Pie charts showing t
The distribution of  valid mutants  (inputs that passed
\approach's invariant check) 
%versus 
%valid inputs 
and discarded mutants (inputs that failed the invariant check)
%identified by 
 for (a) 
All inputs 
(b) Intersectional Inputs and 
(c) 
Atomic Inputs 
%\todo{remove border in all three figures}
}
\label{fig:pie-chart-invariant-check}
\end{figure}

\noindent
\textit{RQ3.1 Number of Valid vs.  Discarded Inputs:}
%\textit{\textbf{Motivation}:}
We examine the 
%contribution of \approach's invariant check in terms of the 
proportion of 
%and the grammatical validity of the 
mutants 
%inputs 
that pass or fail \approach's invariant check, it involves all mutants (valid and discarded test inputs)
generated by \approach 
for all 
%models and 
datasets. 
%/datasets.
% and datasets.   
%the invariant check 
\autoref{fig:pie-chart-invariant-check} presents our findings.  

\textit{\textbf{Results}:} We found that 
%Results show that 
\textit{\approach's invariant check is effective in identifying invalid mutated inputs:
% that are invalid: 
%w.r.t. to the original inputs.  
%that t
\approach discards the majority \checknumber{(97.58\%)} of mutated inputs as invalid because 
%, i.e.,  
%when 
their dependency parse tree does not align with the parse tree of the reference original text. } 
\autoref{fig:pie-chart-invariant-check} shows that most input mutations (\checknumber{129M/132M})
%(129,249,546 out of 132,457,241 total)} 
%135,642,395 out of 139,301,590 total)} 
result in a different 
%grammatical structure (
dependency parse tree 
%(POS or grammatical relations) 
in comparison to the original text.  Intuitively, this means that the majority of generic mutations create 
%non-equavalent inputs, i.e,  
inputs that have a different meaning or intent in comparison to the original reference text.  
%either grammatically wrong 
We observed that \textit{only} 
 \checknumber{2.42\%} of 
% one in fourty (2.42\%)
input mutations \textit{pass} our invariant check.  
This suggests that generic mutations (without our invariant check) mostly 
%(97.37\%)  lead to 
result in invalid test inputs.\footnote{
%We note that
%state-of-the-art 
% \approach without invariant check, is similar to a 
Generic mutation-based bias testing approaches (e.g.,  MT-NLP~\cite{ma2020metamorphic}) mutate inputs without a validity check. }
% This result emphasizes the importance of our invariant check for mutation-based bias testing.   
%Indeed,  we recommend incld
%We also observed that 
%In our experiments, i
Intersectional bias mutations and longer texts are more likely to be invalid 
%error-prone 
(fail our dependency invariant) 
%, in comparison to atomic bias mutations.  About \checknumber{87.06\% (9,043,664 out of 10,387,529)} of atomic input mutations were discarded for failing the invariant check,  in comparison to \checknumber{98.20\% (126,598,731 out of 128,914,061)} of intersectional input mutation.  
%In addition,  we found that bias-prone input mutations are more error-prone as the inputs become longer or more complex.  For instance,  about \checknumber{25.71\% (839,784 out of 3,266,895)} of atomic input mutation are \textit{valid} when mutations are within the first 512 tokens (BERT models),  while \textit{only} \checknumber{7.08\% (504,081 out of 7,120,634)} are \textit{valid} when mutations are within 4096 tokens (e.g., for GPT3.5 and Llama2).  
%We attribute the increasing invalidity of mutated inputs for intersectional (vs.  atomic) bias and longer token size 
due to their 
%increased 
complexity.  
% In summary, 
%Overall,  
%t
These results emphasize the importance of \approach's invariant check. 
%especially for intersectional bias testing.  
%It also 
%demonstrate that \approach (with invariant check) ensures that resulting mutants have the same semantics (dependency parse tree) as the original reference text. 
% 

\begin{result}
%\revise{
%The invariant check of 
\approach's invariant check effectively identifies invalid inputs:
%that are invalid w.r.t. to the original inputs.  
%that t
%\approach 
It discarded millions 
%he majority 
%\checknumber{(97.37\%, 135M)} of 
(\checknumber{%97.58\%=
129M/132M})
of mutants
%ted inputs 
whose
%with 
dependency parse trees
% are different 
%differs from that of
do not conform to 
%that of  
%which 
%are invalid 
%with a different dependency parse tree,  in comparison to 
the original text.  
%}
\end{result}

%\noindent
%\textit{\textbf{Grammatical Validity:}}
% of Discarded Inputs:}}

%\begin{figure}[!tb]
%\centering
%\includegraphics[width=0.5\textwidth]{figures/validity.png}
%\caption{
%Validity
%\approach
%}
%\label{fig:workflow}
%\end{figure}
%
%\begin{figure}[!tb]
%\centering
%\includegraphics[width=0.5\textwidth]{figures/Validity_sim_vs_notsim.png}
%\caption{
%Validity Discarded vs Passing
%\approach
%}
%\label{fig:workflow}
%\end{figure}

\begin{figure}[tb!]
\centering
\includegraphics[width=0.49\textwidth]{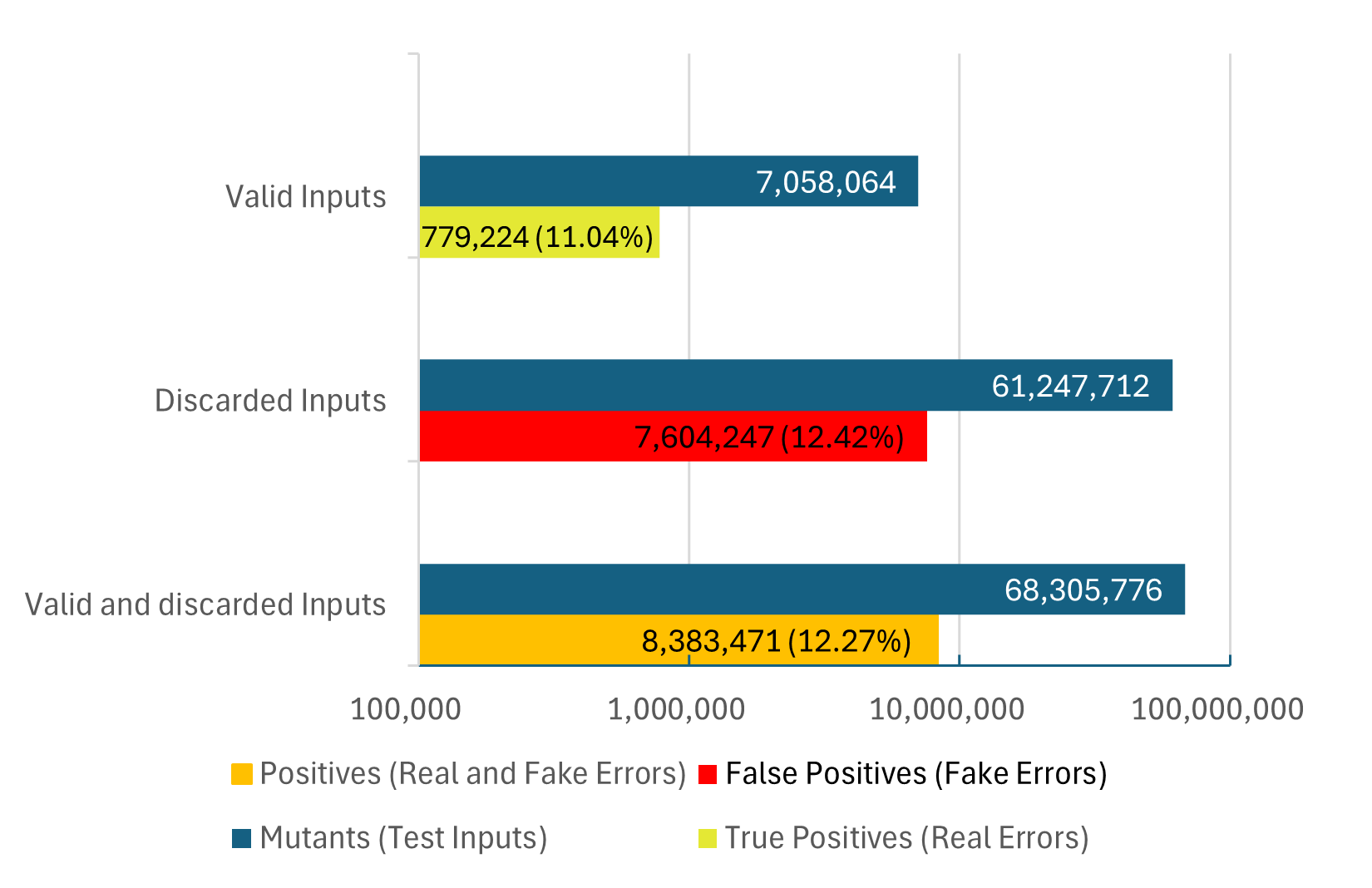}
%\vspace{-1.0em}
\caption{\centering 
%Proportion of 
True and false positives of \approach. 
}
%\todo{this is not the right figure, update what we drew in the meeting}
\label{fig:false-positives}
\end{figure}

\noindent
\textit{RQ3.2 False Positives:}
%\label{sec:false-positives}
%\textit{\textbf{Motivation}:}
%We inspect the false positive errors induced by the inputs discarded by \approach 's dependency invariant check. We compare this to the 
%%in comparison to the 
%number of real errors (true positives) induced by valid generated inputs that pass \approach 's dependency invariant check.
%The aim is to determine 
We investigate the false positives (fake errors)
developers would have 
been saddled with inspecting \textit{without} \approach's invariant check.
%We note that  \approach (atomic) \textit{without invariant check} is similar to generic 
%mutation-based approaches that are not equipped with a \textit{validity check} (e.g., MT-NLP~\cite{ma2020metamorphic}).
% and prone to producing false positives.
% due to lack of input validation. 
%\textit{\textbf{Settings}:}
This experiment 
involves the valid inputs,  
%employs all of the 
discarded inputs and their corresponding errors for the 
\recheck{16} fined-tuned models using the four legal datasets.\footnote{We 
%have employed 
employ the  fine-tuned models since we can execute millions of discarded inputs locally without incurring  huge financial and computational costs, unlike 
%of checking the LLM response for invalid inputs on costly
pre-trained models.} \autoref{fig:false-positives} presents our results. 

%\todo{fix result discussion to match new chart -- \autoref{fig:false-positives} }

\textit{\textbf{Results}:}
%We found that,
%mutation 
\textit{Without \approach's dependency invariant,  developers would need to inspect about \recheck{10 times (10X)} as many fake errors as real errors}. 
\approach saves developers a huge amount of time and effort, 
it reduces the 
\revise{biases} testing effort by up to \recheck{10 times} \recheck{(779,224 
%intersectional bias errors 
vs.  7,604,247). } 
%\textit{
% \approach's 
%input mutations 
%lead to millions of false positives.  Bl
Generic mutations,  without dependency invariant,   
%(without dependency invariant  check) 
%Without dependency invariant,  \approach 
produce millions of false positives --  about 
nine out of every ten bias errors are
%\recheck{92.06\% (6.86m out of 7.45m)} 
false positives:
\autoref{fig:false-positives} shows that 90.71\% (7,604,247/8,383,471) errors produced by generic mutations are 
%i.e.,  
fake. 
%intersectional 
These errors are produced by invalid 
%are
%errors 
%resulting from 
%,  i.e.,  the inputs that induce the errors 
mutants that do not conform with the original input.
% (\textit{see} \autoref{fig:false-positives}).    
%This implies that,  
%  which false positives 
%Thtis suggest that
%Thus,  our dependency invariant reduces the 
%bias testing effort by up to \recheck{12 times by using only the 8.6\% of all bias-inducing mutants that are valid}. 
% errors. 
%We emphasize that the state-of-the-art generic mutation-based 
%bias testing approaches (e.g., MT-NLP~\cite{ma2020metamorphic})
%are similar to \approach (atomic bias testing mode) without invariant check.   
% by discarding invalid inputs that would have led to false positives.  
We also observed that intersectional bias testing produces more false positives than atomic bias testing \recheck{(6,865,033 vs.  739,214)} due to the increasing complexity of intersectional mutants.  
%This result 
%implies that, without test validity checks,  mutation-based approaches produce many false positives.  It 
%suggests that 
These results 
%further 
demonstrate the contribution of
% a test and our 
\approach's dependency invariant. 
%-based validity check. 
%In particular, it ensures 
%% in ensuring
%that  \approach 
%%mutation-based bias testing 
%produces \textit{real} intersectional bias  and avoids 
%%\textit{not} 
%\textit{fake} errors induced by \textit{invalid} mutants. 
%% do not  
%\textit{our dependency invariant check saves developers from inspecting millions of false positives
%}

\begin{result}
\revise{
\approach's  dependency invariant  
%\approach 
saves developers from inspecting millions 
%(6.86m) 
of false positives (7,604,247)
%(fictitious  intersectional bias errors)
which are 
%\approach without invariant check triggers 
\recheck{10 times (10X)}
%= 
%)} 
%\recheck{(
%/ )}
as many as 
true positives (779,224).
% (real errors). 
}
\end{result}

\begin{figure}[tb!]
\centering
\includegraphics[width=0.49\textwidth]{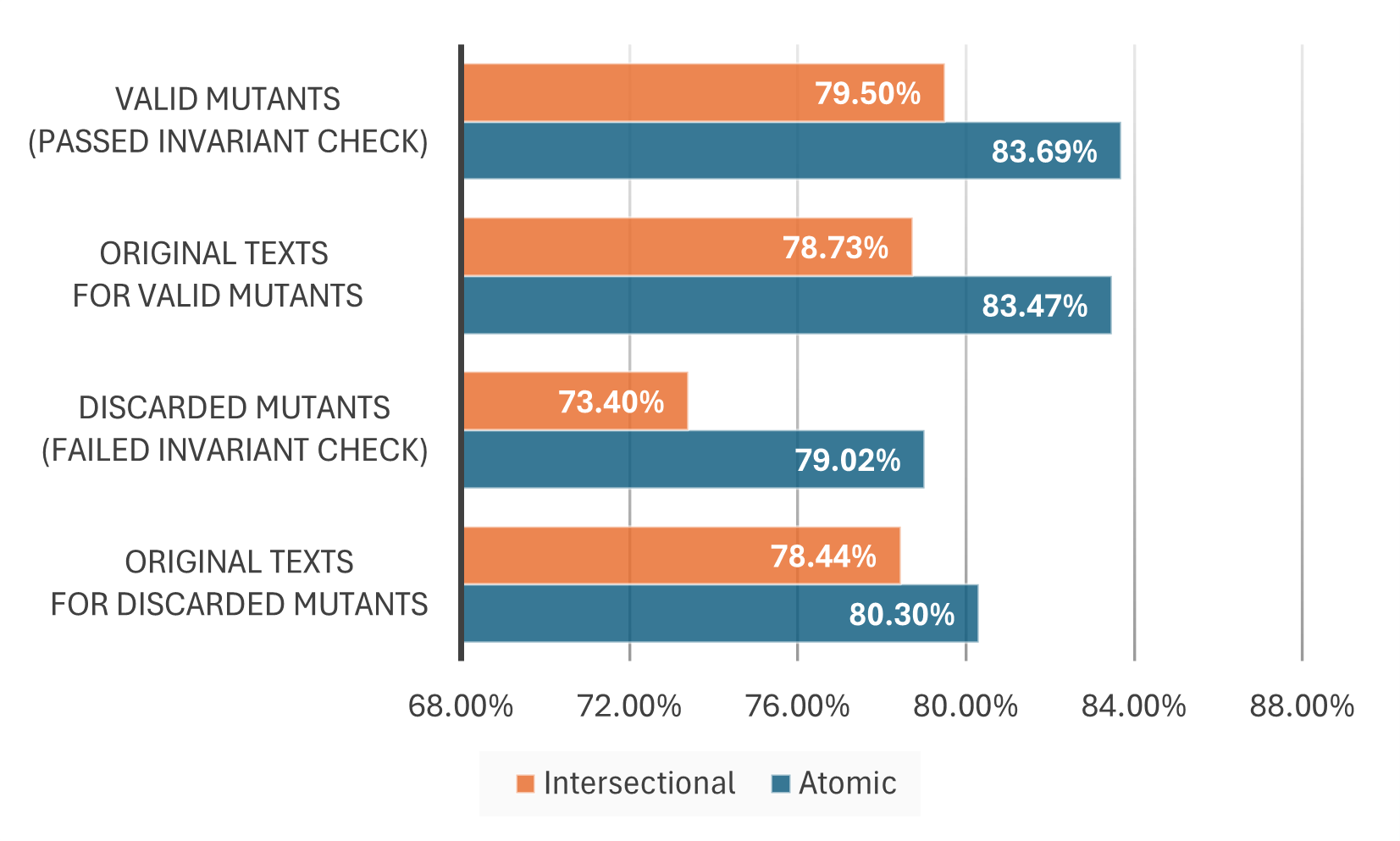}
%
% \quad \quad 
% (a) \approach's Generated Inputs 
% \quad \quad \quad \quad \quad \quad 
% (b) Valid Inputs vs.  Discarded Inputs
%  \quad \quad \quad \quad 
%\vspace{-1em}
\caption{\centering 
Grammatical Validity. \textit{Valid} vs discarded Inputs that failed invariant check.}

%\todo{fix original naming}

\label{fig:grammarly-results-invariant-check}
\end{figure}

\noindent
\textit{
%Grammatical 
RQ3.3 Validity of 
%Valid vs.  
Discarded inputs:}
%\textit{\textbf{Motivation}:} 
%examine the grammatical validity of the discarded inputs versus  \approach's generated valid inputs that pass the invariant check.  
%This 
We compare the grammatical validity of \textit{discarded}  inputs 
%by \approach (dependency invariant) 
%'s invariant check 
%(i.e.,  inputs 
(\textit{failed} dependency invariant)
% check) 
%in comparison 
%to that of 
vs the \textit{valid} inputs 
generated by \approach.
% (i.e.,  
%generated 
%inputs 
%(\textit{passed}  it. 
%\approach's invariant check). 
%
%\textit{\textbf{Settings}:} 
This experiment employs 
Grammarly~\cite{grammarly} and a similar setting as \textbf{RQ2}.  
We sample an additional \checknumber{67K} discarded mutants to compare to the \checknumber{68K} valid inputs sampled in \textbf{RQ2} for valid/generated inputs. 
%, i.e., generated inputs that failed \approach's invariant check. 
%
% and  \checknumber{100K} valid inputs.  
%In particular,  
%In this experiment,  
Similar to \textbf{RQ2},  
%For each set of discarded inputs or valid inputs,  
we randomly sample about \checknumber{10K} original inputs 
%\checknumber{88K} resulting in 
and their corresponding \checknumber{67K} discarded mutants, while aiming for balance across sensitive attribute. 
%% resuuting in   \checknumber{88K} discarded mutants and  \checknumber{88K} valid mutants.  
%\revise{Our sampling method ensures balance across sensitive attributes,   datasets as well as atomic and intersectional mutants.  For instance, we aimed to sample about \checknumber{9K} original inputs each for atomic and intersectional bias with \checknumber{33\%  (3.3K)} per sensitive attributes. }
% \todo{describe sampling method across attributes, atomic and intersectional, }
We present our findings in \autoref{fig:grammarly-results-invariant-check}. 
%the grammatical validity of discarded inputs versus valid inputs.  

%\checknumber{X\%} 

\textit{\textbf{Results}:} 
We found that  \textit{valid inputs 
are up to \checknumber{8.31\%} more grammatically valid than the inputs discarded by \approach's invariant check. } \autoref{fig:grammarly-results-invariant-check} shows that valid inputs generated by \approach (i.e.,  pass its invariant check) maintain similar grammatical validity as the original input (\checknumber{79.50\% vs.  78.73\%}).   In contrast,  
%\autoref{fig:grammarly-results-invariant-check} shows that 
discarded mutants (that fail invariant check) do not preserve the grammatical validity of the corresponding original texts. 
%These
%Besides,  
\textit{Discarded mutants %are 
have (up to  
\checknumber{6.4\%}) 
%less grammatically valid as their corresponding original inputs. } 
%Discarded mutants 
%do not preserve the validity of the original (human-written) texts, instead, they 
%have 
lower grammatical correctness scores than the 
%reduce the validity of the 
original texts}
%.  For instance,  the discarded mutants are lower than the original mutants for intersectional mutants 
(\checknumber{78.44\% vs.  73.40\%}).  
%Discarded inputs reduced (and do not preserve) the grammatically validity of the original inputs.  This is particularly more evident for intersectional bias,  where the inputs discarded by \approach's invariant check have significantly lower grammarly scores than the original inputs 
%
These results demonstrate 
that 
%In summary, 
\approach's dependency invariant 
%the contribution of 
%\approach's invariant check : It 
preserves
% in preserving 
%maintaining 
the validity of the original inputs. 
% in generated mutants.
% and 
%,
%  thereby ensuring 
%ensure generated inputs are valid.  
%, especially with respect to the original inputs.  

\begin{result}
%\revise{
\approach's invariant check preserves the validity
%tical structure 
of the original 
%(human-written) 
text:
%\approach's generated inputs are as valid 
%(
%Inputs that pass invariant check are as valid 
%) 
 %have similar grammatical validity 
%scores 
%as the original text,  
%and its 
%but d
Generated inputs are as valid as original inputs but up to \checknumber{8.31\%} more grammatically valid than discarded inputs.
%  by \approach's invariant check. 
%Discarded mutants have ($\approx$\checknumber{4.7\%}) lower
%%reduce the 
% validity 
%%in comparison to
%% of 
%than  
%the original inputs and the 
%%valid 
%%generated 
%inputs that pass our invariant check. 
% (by $\approx$\checknumber{4.7\%}).
%are ($\approx$\checknumber{4.7\%}) less grammatically valid as 
%%their corresponding 
%the original inputs. 
%}
\end{result}

%\todo{move atomic bias error rate and distinct atomic bias-inducing mutants (above) to supplementary material}

\noindent
\textbf{RQ4 Atomic Bias vs.  Intersectional Bias:}
%\textit{\textbf{Motivation}:} 
%In this experiment,  we 
%We  examine the likelihood 
%%whether 
%%\textit{
%%it is likely 
%that an LLM 
%%AI system may not be 
%is \textit{not} discriminatory to 
%%instances of an 
%atomic attributes (e.g., race or gender),  but discriminatory to a simultaneous combination of both attributes (e.g.,  race X gender).  
%%This \textbf{RQ} is inspired by \textit{intersectionality theory} where people can be disadvantaged due to multiple characteristics,  i.e.,  a combination of sensitive attributes~\cite{crenshaw1989demarginalizing}.  
%\revise{
%To address this problem,  w
We examine whether the same set of original texts 
%(Section \ref{RQ4-unique-cases}) 
and input mutations 
%(Section \ref{RQ4-mutants}) 
that 
induce intersectional bias also induce atomic bias,  or not. 
%or strictly intersectional bias, or atomic bias.
% or not.  
% or not.  
%}

\begin{figure}[H]
%\vspace{-0.5em}
\centering
\includegraphics[width=0.49\textwidth,trim={0 0.8em 0 0},clip]{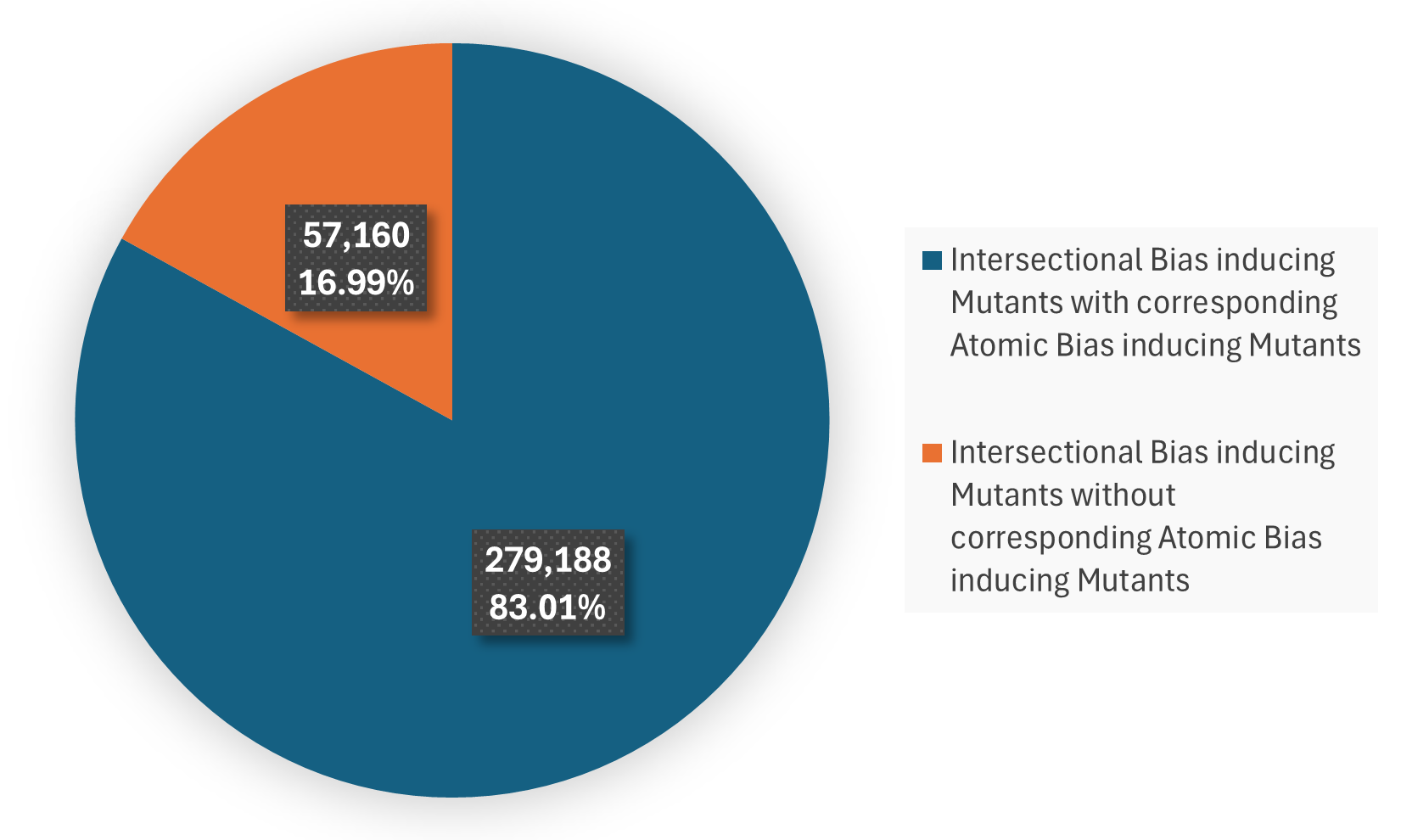}
\includegraphics[width=0.49\textwidth,trim={0 0.8em 0 0},clip]{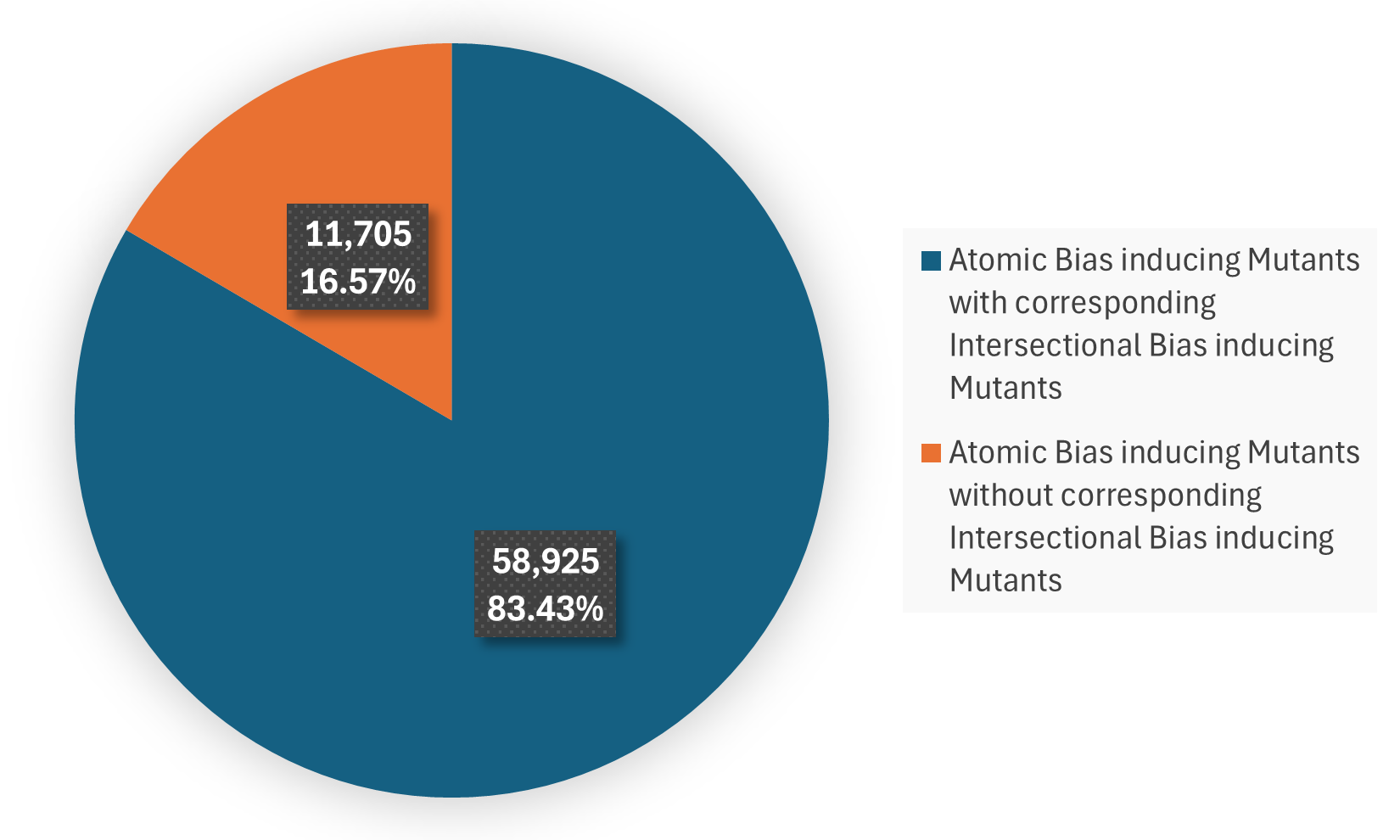}
 \quad \quad (a) Intersectional  Bias Testing \quad \quad \quad \quad \quad \quad (b) Atomic  Bias Testing \quad \quad \quad \quad 
% (c) Atomic Inputs  \quad \quad
%\quad \quad \quad \quad
%\vspace{-0.5em}
\caption{\centering 
Pie charts showing the distribution of bias-inducing mutants for intersectional 
%bias testing 
and atomic bias testing.  
%\autoref{fig:pie-chart-atomic-vs-inters} (a) shows the distribution of atomic bias-inducing mutants with/without a corresponding intersectional bias-inducing mutants,  and  \autoref{fig:pie-chart-atomic-vs-inters} (b) highlights the proportion of intersectional-bias-inducing mutants with/without a corresponding  atomic bias-inducing mutants. 
}
\label{fig:pie-chart-atomic-vs-inters}
\end{figure}

\noindent
\textit{RQ4.1 Bias-inducing Mutants:}
%\noindent
%\textit{\textbf{Settings}:}
%\checknumber{
This experiment employs 
%a set of
(410K) bias-inducing  
mutants that are common to both intersectional mutations and 
atomic mutations performed by \approach.\footnote{Specifically,  
%this experiment  involves 
\checknumber{410,221} bias-inducing mutants includes 
%composed of 
\checknumber{70,630} \textit{atomic} 
%bias-inducing 
mutants 
%that triggered an atomic bias 
and \checknumber{336,348} \textit{intersectional} 
%bias-inducing ,
mutants. The high number of intersectional mutants (336K) is due to the combinatorial nature of intersectional bias. }
%For a fair experiment,  
%In this experiment, we employ mutants such that each valid 
%%employed mutant, there is an 
%(336K) intersectional mutant and (70K) corresponding
%%component 
%%/equivalent intersectional 
%valid atomic mutant (e.g.,  \autoref{fig:illustrative-example}).
%In this experiment,  we examine (\todo{XX})  mutants that have corresponding that have triggered intersectional bias and have a corresponding atomic mutants.  
%In this experiment, w
%We 
%In total, 
%we had \checknumber{2,467,224} mutants containing 
%and 
% of which 
%. 
%triggered a bias.  In particular,  
%Overall, we had 
% that triggered an intersectional bias. 
%\todo{Table XXX}
%and 
\autoref{fig:pie-chart-atomic-vs-inters} highlights our findings.  
%In particular,  
%\autoref{fig:pie-chart-atomic-vs-inters}(a) highlights the proportion of intersectional bias-inducing mutants with/without a corresponding  atomic bias-inducing mutants, and
%\autoref{fig:pie-chart-atomic-vs-inters}(b) shows the distribution of atomic bias-inducing mutants with/without a corresponding intersectional bias-inducing mutants. 
%a venn diagram showing 
%the distribution of bias errors across intersectional and atomic mutations.  
%}

%\baselineskip
%\noindent
\textit{\textbf{Results}:}
%show 
%We found that,  i
In absolute terms, 
\textit{atomic bias testing misses about  
\checknumber{five times (5X)}  as many bias-inducing mutants as intersectional bias testing \checknumber{(57,160 vs. 11,705)}. } 
\recheck{\autoref{fig:pie-chart-atomic-vs-inters}}(a) 
%\autoref{fig:pie-chart-intersectional}} 
shows that 
\checknumber{16.99\% (57,160/336,348)} of intersectional bias-inducing mutants do not have any corresponding atomic bias.  
This implies that {\em atomic bias testing is insufficient to reveal intersectional bias instances.} 
%Meanwhile,  intersectional bias testing 
%misses \checknumber{16.57\% (11,705/70,630)} of atomic bias instances.
%However,  
%%we found that \checknumber{83.43}\% (\checknumber{58,925/70,630})  of atomic bias-inducing mutants have a corresponding 
%%intersectional bias-inducing mutants.  
%%Besides,  
%\textit{\recheck{about one in six} intersectional bias-inducing mutants do not trigger 
%% induce 
%%a corresponding 
%an atomic bias, 
%i.e.,  their corresponding atomic mutants are benign when tested 
%%the atomic mutants 
%individually. } 
%\todo{Table XXX}
%and 
%Performing only atomic bias testing fails to uncover thousands of intersectional bias instances.  
Meanwhile,  we found that \checknumber{83.43}\% (\checknumber{58,925/70,630})  of atomic bias-inducing mutants have a corresponding 
intersectional bias-inducing mutants 
%,  as shown in 
(\autoref{fig:pie-chart-atomic-vs-inters}(b)).
This suggests that intersectional bias testing exposes about 
\checknumber{four in five}
atomic bias instances.   
%Moreover,  
%\textit{the number of intersectional bias instances missed by atomic bias testing is \checknumber{5.5} times as much as the 
%%number of 
%atomic bias instances missed by intersectional bias testing} \checknumber{(57,160 vs.  11,705)}.  
%This 
%Overall,  t
%In summary,  t
These results emphasize that 
%importance of intersectional bias testing
% --
%In summary,  t
%These results 
%demonstrate the importance of intersectional bias testing. It 
%further 
%It shows
%the complementarity of intersectional bias testing.   
%suggest 
%that intersectional bias testing is  
%showing its 
% and 
%our approach (\approach) is 
%and its 
%complementarity of 
intersectional bias testing complements atomic bias testing.  
%It is important to state that the number of bias inputs for 

%\recheck{Even if one in five instances are not exposed by both intersectional and atomic bias instances, the number of intersectional instances not exposed by atomic is higher and so has more bias.}
%Hence, w
%We recommend 
%that ML engineers 
%performing both atomic and intersectional bias testing for LLMs. 

\begin{result}
\revise{
%Intersectional bias testing is 
%%unique but 
%complementary to atomic bias testing:
%On the one hand,  
%About 
%\checknumber{one in six (57K/336K)}
% intersectional bias-inducing mutants
% do not have a corresponding atomic bias-inducing mutant. 
Atomic bias testing misses   
\checknumber{five times (5X)}  as many bias-inducing mutants as intersectional bias testing \checknumber{(57,160 vs. 11,705)}.  
%  instances 
%are not exposed by atomic bias testing. 
%In comparison,  
%\checknumber{about 
%five times (5X) fewer (11,705) }
%atomic bias-inducing mutants
% do not have a corresponding
% intersectional bias-inducing mutants
%% instances are missed 
%%by intersectional bias testing.
%% 
%\checknumber{(57160/11,705)}.
}
\end{result}

%\todo{Can we discuss these results in terms of distinct cases: \\
%Overall and for each sensitive attribute \\ 
%- How many unique/distinct original inputs lead to both atomic bias and intersectional bias? \\
%- How many unique/distinct original inputs lead to strictly and only atomic bias? \\
%- How many unique/distinct original inputs lead to strictly and only intersectional bias? \\
%}

\begin{figure*}[tb!]
\centering
% \vspace{-0.5em}
\includegraphics[width=0.32\textwidth,trim={0 0.8em 0 0},clip]{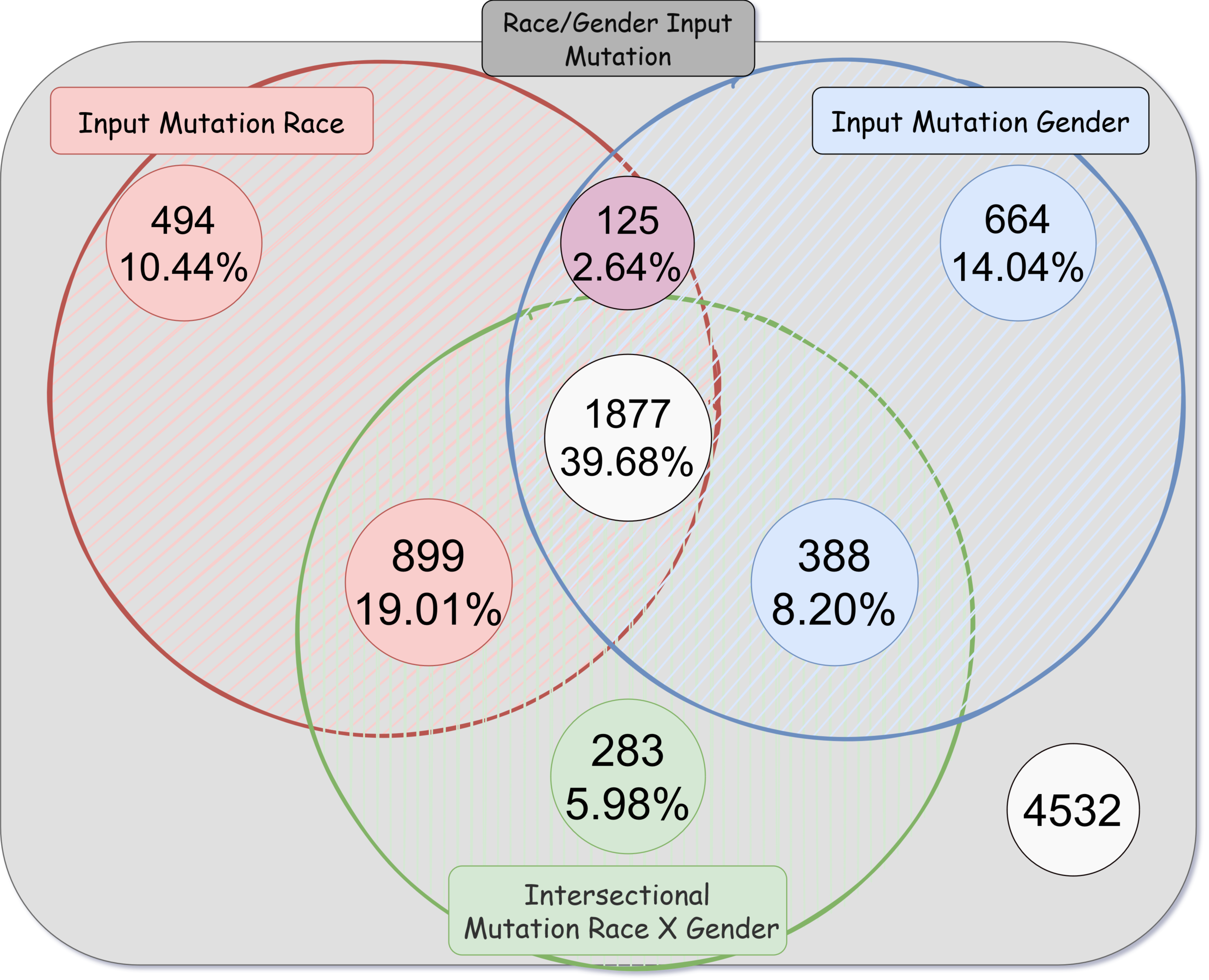}
\includegraphics[width=0.32\textwidth,trim={0 0.8em 0 0},clip]{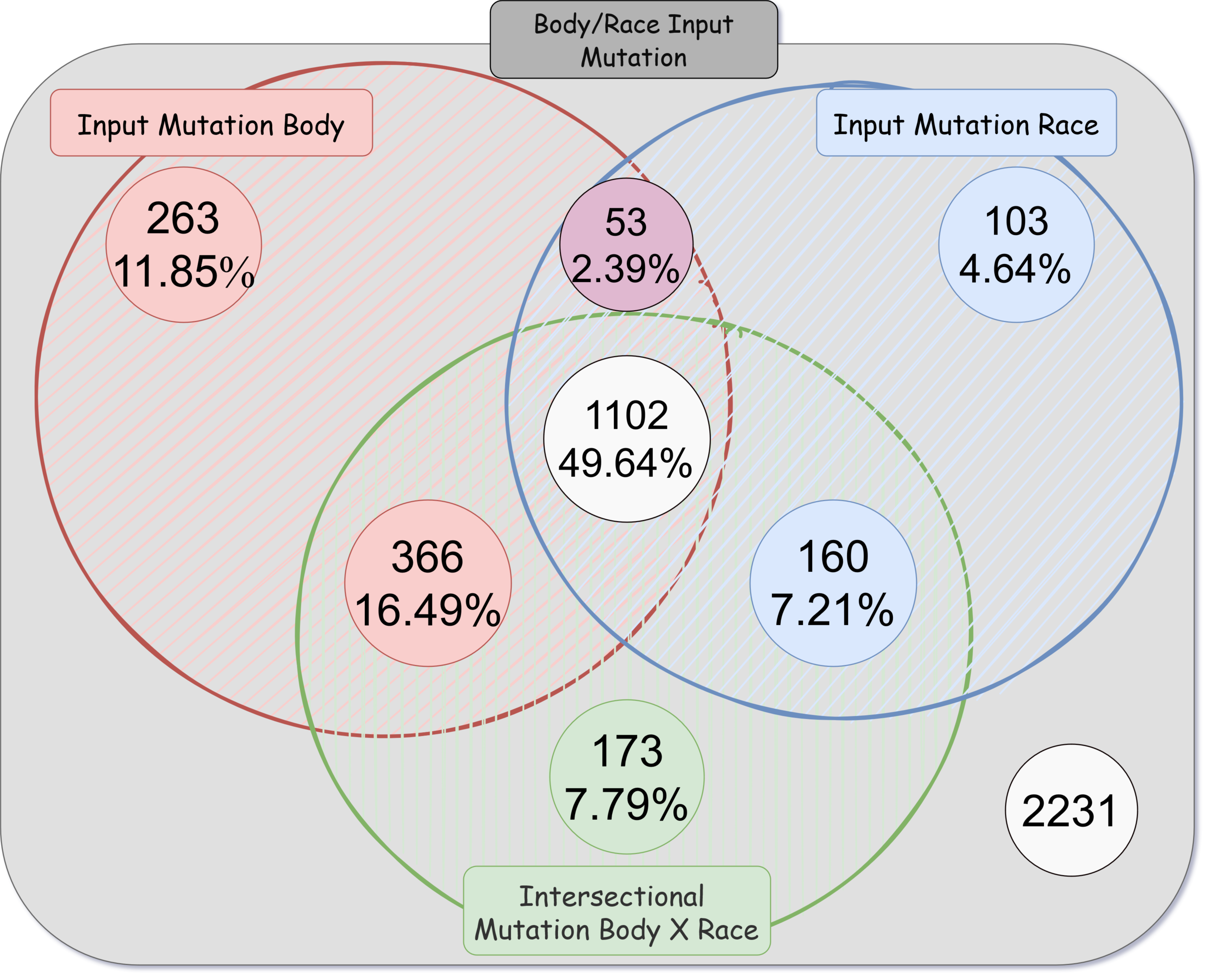}
\includegraphics[width=0.322\textwidth,trim={0 0.8em 0 0},clip]{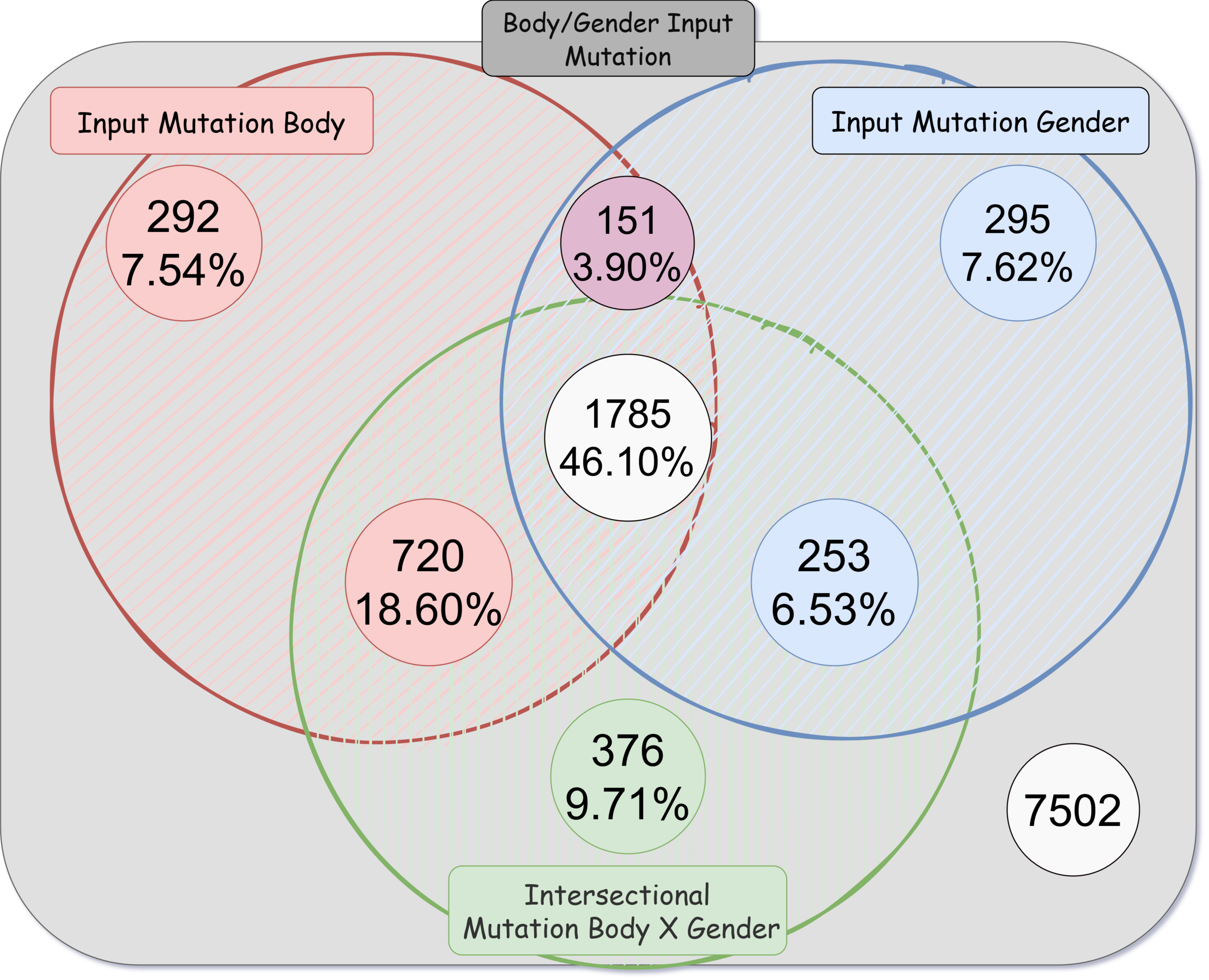}
%\vspace{-0.5em}
 \quad \quad (a) 
Race X Gender
\quad \quad \quad \quad \quad \quad (b) 
Body X Race \quad \quad \quad \quad (c) Body X Gender  \quad \quad
%\quad \quad \quad \quad
% \vspace{-0.5em}
\caption{\centering 
\revise{Venn diagram showing the distribution of bias-inducing original inputs 
leading to atomic vs.  intersectional 
%bias-inducing 
errors for sensitive attributes 
(a) 
Race X Gender
(b) Body X Race and 
(c) 
Body X Gender
}}
%\todo{replace "mutation" with "original cases" in the diagrams}
\label{fig:venn-diagrams-atomic-vs-inters-bias}
\end{figure*}

\noindent
\textit{RQ4.2 Bias-inducing Original Inputs:}
%\textit{\textbf{Settings}:}
%In this experiment,  w
We inspect the \textit{bias-inducing original inputs} 
%-- 
%%bias-inducing 
%original inputs 
whose mutations lead to intersectional bias or atomic bias.  
%The aim is to understand whether the same set of original inputs lead to both types of biases.  To this end,  we inspect all original inputs that have both a valid intersectional mutation (e.g.,  race X gender) and corresponding valid atomic mutations (e.g.,  race only, and gender only), across all experimental settings.  
%We report our findings in 
\autoref{fig:venn-diagrams-atomic-vs-inters-bias} highlights our results.\footnote{For instance,  \autoref{fig:venn-diagrams-atomic-vs-inters-bias} (a) shows that 1,877 original inputs led to bias-inducing mutants for both intersectional bias testing (race X gender) and atomic bias testing campaigns (race only and gender only).  It also shows that 4,532 original inputs are benign, do not result in intersectional bias.
%-inducing mutants.  
}
%the proportion of bias-inducing original inputs that are (un)common across sensitive attributes.

\textit{\textbf{Results}:}
\revise{\textit{Up to one in ten (9.71\%) bias-inducing original inputs strictly lead to intersectional bias. } Across all settings,  about 5.98\% to 9.71\% of bias-inducing original inputs strictly trigger \textit{only} intersectional bias.}
%(\textit{see}  \autoref{fig:venn-diagrams-atomic-vs-inters-bias}):  
\revise{As an example,  \autoref{fig:venn-diagrams-atomic-vs-inters-bias}(c) shows that 9.71\% (376) of bias-inducing original inputs strictly lead to intersectional bias (Body X Gender) sensitive attribute.  
We observed that \textit{up to half of all bias-inducing original inputs lead to both intersectional bias and atomic bias}.   Notably, 49.64\% of bias-inducing original inputs lead to both intersectional bias as shown in \autoref{fig:venn-diagrams-atomic-vs-inters-bias}(b) (Body X Race), as well as atomic bias (Body only and Race only).  Finally, about 4.64\% to 14.04\% of bias-inducing original inputs strictly lead to atomic bias.  These result
suggests that some intersectional bias-inducing original inputs may not induce atomic bias,  
%but induce intersectional bias, 
and vice versa.}
%While up to half of bias-inducing original cases are common (not unique) to atomic and intersectional bias .  
%Overall, these result emphasize the 
%rarity of intersectional bias and the 
%uniqueness of intersection bias testing in comparison to atomic bias testing. 

\begin{result}
\revise{
%While u
Up to 
\checknumber{one in ten (9.71\%)} original inputs 
are strictly intersectional 
bias-inducing,  
%-- 
%not trigger an atomic bias. 
%These original texts strictly lead to intersectional bias,  i.e.,  
%their mutations trigger only intersectional bias,  and 
their mutations do not trigger atomic bias. }
\end{result}

\section{Limitations and Threats to Validity}
\label{sec:threats}

\revise{We discuss the limitations of this work and the threats to the validity of  \approach and its findings.}

\noindent
\textbf{Construct Validity:}
%\todo{metrics,  measures,  prompting and automated analysis, test oracle}
%
\revise{The main threat to construct validity of this work includes the 
automated bias detection in LLM responses and the metrics employed in our experiments. 
To mitigate these threats, we have employed standard bias detection methods and measures.  
In particular, we have employed a metamorphic test oracle to detect biases in LLM outcomes. 
To enable automated analysis of LLM outcomes,  we have employed system prompts and few-shot prompting to ensure that LLM responses are within the allowed outcomes.  We have also employed typical measures such as the error rate,  number/proportion of error-inducing generated/mutated or original inputs. 
We note that these oracle setting and measures are commonly employed in bias testing literature~\cite{ribeiro2020beyond,soremekun2022astraea,ma2020metamorphic,rajan2024distribution}.}

\noindent
\textbf{Internal Validity:}
\revise{The threat to internal validity refers to whether our implementation of \approach actually discovers  intersectional bias.  To mitigate this threat, we have tested our implementation, conducted code review and manually assessed samples of \approach's results.   We have further conducted experiments to examine the grammatical validity of \approach's generated inputs versus discarded and (human-written) original inputs (\textbf{RQ2} and \textbf{RQ3}).  In addition,  our dependency invariant check automatically assess the semantic validity of our test inputs w.r.t. to original inputs.  We have further assessed its contribution to the performance of \approach (\textbf{RQ3}).}

\noindent
\textbf{External Validity:}
\revise{The main threat to external validity 
%of this work 
is the generalizability of \approach and its findings to other LLMs,  datasets  and tasks beyond the ones used in this work.  We mitigate this threat by employing well-known,  state-of-the-art LLM models including open weight models (e.g. , Llama2 and BERT) and closed-source commercial/proprietary models (e.g., GPT3.5).  We note that our models cover the three main LLM model architectures -- encoder-only (e.g., BERT, Legal-BERT, RoBERTa),  decoder-only, (GPT.3.5, Llama) and encoder-decoder (DeBERTa) architectures.  We have also employed well-studied sensitive attributes~\cite{goharsurvey},  complex tasks (e.g., judgment prediction (ECtHR)~\cite{chalkidislexglue2022}) and well-known datasets (e.g.,  IMDB).  
%We note that it is easy to adapt \
\approach can be easily adapted to new LLM models, datasets and tasks.  Finally,  we employ few-shot prompting in our experiments (for GPT3.5 and Llama2), thus our findings may not generalize to other prompting techniques.}

\noindent
\textbf{Dictionary Completeness and Soundness:}
\revise{Our bias dictionary may not generalize to other sensitive attributes
and may be incomplete, e.g.,  in comparison to a standard English dictionary. 
%,  it does not cover all bias-prone word pairs in 
To mitigate this threat,  we used a bias dictionary extracted from SBIC~\cite{sap-etal-2020-social},  a well-known and commonly used bias corpus containing 150K social media posts.  However, adapting \approach to a new sensitive attribute or extending the dictionary may require using a new dataset or specifying a new bias dictionary,  albeit this is achievable, using \approach, within few LoC/minutes.}

\noindent
\textbf{Higher-order mutations:}
\revise{Intersectional bias instances may be characterized by 
higher-order mutations, beyond the second-order mutations employed in this work. That is,
more (than two) simultaneous combination of sensitive attributes.  
However,  in this work,  we focus on intersectional bias characterized by 
second-order mutants --  
%only 
two simultaneous sensitive attributes -- 
% This is 
due to the computational complexity of exploring multiple sensitive attributes.  %We note that
 \approach can be easily extended to study much higher-order mutations (third-order, fourth-order, etc).  However, this is computationally intensive.  Further research is required to 
% and requires better optimization and research to 
obtain a reasonable trade-off between the higher-order mutations and computational cost.  In the future,  we aim to study efficient intersectional bias testing algorithms for an  arbitrary number (i.e.,  more than two simultaneous) sensitive attributes.}

\noindent
\textbf{Compute Time: }
% \revise{
\revise{Our experiments 
%are computationally intensive due to the large set of datasets/models and the millions of input generated across all datasets.  In our experiments, 
generating inputs with 512 and 4096 long tokens took three (3) days (16K CPU days) and 21 days (110K CPU days), respectively.  Testing the fine-tuned models took about 4 days and 2700 GPU days.  In contrast,  testing GPT-3.5 and Llama2  took around 14 days and 2800 GPU days.  We observed that the most time-consuming portion of our experiments is the computation of the invariant check.}
% }

\section{Ethics Statement}
\label{sec:ethic_statement}
%\revise{
\revise{We elucidate our ethics statement in this section:}

\noindent
\textbf{Project Goals:}  
\revise{This work explores an ethical concern in LLMs, in particular,  intersectional bias testing 
%and analysis 
of LLMs.  Testing LLMs for intersectional bias is a promising way to improve the fairness and trustworthiness of LLMs.  In particular, it allows practitioners (ML/data/software engineers) to find evidence (instances) of bias 
at the intersection of different identity markers.  Identifying such 
%among intersectional individuals and groups. These 
bias instances further allow practitioners to debug 
%inspect for intersectional 
bias,  and 
improve the fairness and trustworthiness of LLMs. 
%  their LLM-based systems to be fairer.    
This work encourages practitioners and companies to employ intersectional bias testing 
%as a task in their 
during LLM 
%-based (software) 
development.}

\noindent
\revise{\textbf{Datasets:} We utilize publicly available datasets namely IMDB (a popular dataset for NLP tasks) and four legal datasets from the EU and USA.  }

%.
\noindent
\revise{\textbf{Human Evaluations}: Our experiments do not involve human participants. }

\noindent
\textbf{Approach and LLMs:} 
\revise{We have tested our approach (\approach) using recent publicly available pre-trained and fine-tuned models,  including a closed-source pre-trained model (GPT3.5 from OpenAI),  an open-source pre-trained LLM  (Llama2 from Meta) and 16 fine-tuned LLMs (from the  Lexglue benchmark~\cite{chalkidislexglue2022}).  
We acknowledge that the pre-trained LLMs may hallucinate,  however, we mitigate this in pre-trained LLMs by setting the temperature to zero, disabling sampling, using system command and employing few-shot prompting.  For GPT3.5,  we further mitigate the effect of model updates by limiting our testing time (to about a day) and tracking news of model updates.}
% before and after testing. 

\noindent
\textbf{Bias Dictionary:} 
\revise{We have extracted bias word pairs from the SBIC dataset~\cite{sap-etal-2020-social}.  SBIC is a publicly available dataset licensed under the Creative Commons CC0 License.}

\revise{Overall,  the design and evaluation of \approach 
%have followed 
follows strict ethical standards (e.g.,  using publicly available, popular and licensed datasets/models),  thus making it unlikely to raise ethical concerns.}

\section{Related Work}
\label{sec:related_work}
%\section{Background}

%\todo{cite ~\cite{liangholistic, zhao2023survey}  (Reviewer 4oPx  EMNLP) }

%\todo{We should consider discussing MT-NLP here as generic mutation}

%\todo{We should also conceptually compare the error rate of \approach to MT-NLP similar to what we did in ASTRAEA}

%\todo{This is a good place to (conceptually) compare \approach to previous mutational and metamorphic NLP bias testing approach, i.e.,  MT-NLP}

  \noindent
\textbf{Intersectional Bias Testing:} Most bias testing techniques focus on atomic \revise{biases}~\cite{mehrabi2021survey, galhotra2017fairness, udeshi2018automated, aggarwal2019black},
where our work tackles 
\textit{intersectional bias}.
Similarly, previous research~\cite{buolamwini2018gender, goharsurvey, soremekun2022software, yang2020causal, cabrera2019fairvis, crenshaw1989demarginalizing, collins2019} have shown that discrimination is multifaceted in the real world. 
These works do not perform intersectional bias testing of LLMs nor reveal hidden intersectional bias or examine the relationship between atomic and intersectional bias testing. \revise{Recent efforts have begun to address this gap. Ma et al.\cite{ma-etal-2023-intersectional} introduced a dataset specifically designed for analyzing intersectional stereotypes in LLMs, demonstrating that these biases persist and require targeted mitigation strategies. Devinney et al.\cite{devinney-etal-2024-dont} employed an intersectional lens to investigate how LLMs generate biased narratives, highlighting the need for adaptive strategies in bias identification across different socio-cultural contexts. Lalor et al.\cite{lalor-etal-2022-benchmarking} benchmarked intersectional bias in NLP models, revealing that existing debiasing techniques are insufficient for mitigating biases that span multiple demographic dimensions. Additionally, Tan and Celis\cite{NEURIPS2019_201d5469} assessed intersectional biases in contextualized word representations, providing empirical evidence of compounded bias effects in language models. Wilson and Caliskan~\cite{wilson2024genderraceintersectionalbias} examined gender and racial biases in LLM-based resume screening, demonstrating that intersectional disparities persist in retrieval-based hiring processes, further emphasizing the need for bias-aware AI evaluation. Unlike these studies, our work uses automated metamorphic testing that generates and validates test cases for intersectional bias detection.}   

\noindent
\textbf{Metamorphic Testing in NLP:} NLPLego~\cite{ji2023nlplego,ji2023intergenerational}  applies metamorphic relations to discover functional errors.  In contrast,  our metamorphic testing approach aims to discover intersectional  bias.  NLPLego is not directly applicable to detecting bias,  especially intersectional bias. MT-NLP~\cite{ma2020metamorphic} is the closest metamorphic bias testing approach to our work.  However, MT-NLP~\cite{ma2020metamorphic} is limited to atomic bias testing, it does not target intersectional bias testing.  Besides,  
MT-NLP employs generic mutations and does not validate inputs. 
As evident in our evaluation (Section 5 \textbf{RQ3}),  generic mutations result in numerous false positives that burden developers. 
 
\noindent
\textbf{NLP Test Validity:} NLP (bias) testing approaches validate test cases using a template or grammar or ML techniques. For example,  Checklist~\cite{ribeiro2020beyond} uses a template to make sure the generated test cases are valid. ASTRAEA~\cite{soremekun2022astraea} uses an input grammar to ensure generated inputs are valid. 
Meanwhile,  other traditional test validity methods in NLP~\cite{liu2022wanli, west2022symbolic, yang2022testaug} train a separate classifier to check validity.  However, the test validity method of these approaches are limited and require manual curation of a template,  grammar or training datasets.  More importantly,  these works use rigid,  pre-defined and less expressive validation methods which are often not directly applicable to new or unseen inputs. Unlike \approach's invariant check, these approaches are not amenable to any arbitrary training datasets. They do not support the validity for 
the construction of arbitrary test cases for any arbitrary seed inputs.  

\noindent
\textbf{Dependency Parsing:} 
%\approach's invariant check ensures generated test inputs grammatically conform with the original input,  consequently,  preserving the validity,  structure and intent of the original input.  
%Additionally,  we note that our 
\approach's dependency invariant is similar to 
the structural invariant employed in the machine translation (MT) testing literature~\cite{he2020structure,zhang2024machine,sun2020automatic}.  Similar to \approach, these approaches employ dependency (or syntax) parsing to validate or generate test inputs. 
%~\cite{zhang2024machine, he2020structure} or filter/sanitize inputs~\cite{sun2020automatic}.  
However,  these approaches are specially designed for MT systems.  They are not applicable for bias testing or other tasks.

\noindent
\textbf{Bias in LLMs:}
\revise{Similar to our work, previous research~\cite{baldini-etal-2022-fairness, delobelle-etal-2022-measuring, wan2024whitemenleadblack}
highlight the challenges in evaluating bias in LLMs. 
However, these works do not perform test generation, they employ only existing datasets. In contrast, our work aims to automatically generate test suites, beyond the existing datasets, to expose (hidden) intersectional bias. Tan and Celis~\cite{NEURIPS2019_201d5469} analyzed the presence of intersectional biases in contextualized word representations, showing that such biases can compound in complex ways. Lalor et al.~\cite{lalor-etal-2022-benchmarking} further benchmarked multiple NLP models for intersectional bias, demonstrating that current debiasing strategies are insufficient for mitigating compounding biases across demographic intersections. Wan and Chang~\cite{wan2024whitemenleadblack} additionally benchmarked social biases in LLMs, revealing disparities in how language models attribute agency based on race and gender, which aligns with our goal of uncovering nuanced biases. While these studies provide crucial empirical evidence of bias in LLMs, they primarily rely on static datasets and benchmarking rather than automated test generation. 
% , as proposed in this work and embodied in \approach. 
In contrast, our work introduces an automated metamorphic testing approach (\approach) that generates test cases to uncover hidden intersectional biases, making bias evaluation more scalable.}

\noindent

\section{Conclusion}
\label{sec:conclusion}
%\todo{fix the numbers here}
%\todo{ Implications in Practice (for Bias Testing/Mitigation, for S/W Engineers, ML Engineers, and Researchers) and legal Implications}
%\revise{
This paper presents 
\approach, 
an automated bias testing technique
% (called ) 
%that generates 
%bias-prone 
%ttest inputs 
that expose intersectional bias in LLMs. 
%The goal of 
It
%aims to 
%tackles two main challenges,  it aim to
%(a) 
generates valid
% bias-prone 
test inputs 
%that are valid w.r.t.  the original human-written inputs and (b) 
that uncover  \textit{hidden} intersectional bias via 
% intances that are during atomic bias testing.  
%To achieve this,  \approach leverages 
a combination of 
higher-order mutation analysis, 
dependency parsing and metamorphic oracle.
% to detect bias. 
%the problem of ensuri
%w
We 
evaluate 
%examine 
the effectiveness of 
\approach using 
%a %\approach 
%We extract a 
%bias dictionary from SBIC~\cite{sap-etal-2020-social},
% which contains 
%curates 
%bias word pairs for 
%and 
%three sensitive attributes, 
%six LLM architectures,  
%six 
\checknumber{five datasets and 18 LLMs} and demonstrate that 
%\approach is effective -- 
%about 
%\checknumber{one in seven} 
\checknumber{14.61\%} of inputs generated by \approach expose intersectional bias.  
%We demonstrate that 
%of
%as the original (human-written) text. 
%\approach's 
%\approach's 
% effectively identifies invalid mutants, 
%of \approach 
%contributes to ensuring 
%further 
%ensures the inputs it generates  
%are valid and they 
%conform to the grammatical structure of the original texts.  
%Its invariant check 
%it shows that the parse trees of the 
%majority (\checknumber{97.37\%}) of
%(135) millions 
%mutants 
%identifies several \checknumber{(135 million mutants representing 97.37\% of)} mutants
%whose dependency parse trees 
%do not grammatically conform to the original text. 
%Furthermore,  .  
%invariant  design decision
%also 
%We  demonstrate t
%Furthermore,  
We also
%our experiments 
show that intersectional bias testing is unique and complements
%ary 
%nature of  in comparison 
%to 
atomic bias testing: 
%Firstly,  
%We found that 
\checknumber{16.62\%} of intersectional bias found by \approach are hidden.
%performed 
%by 
%\approach  
%by empirically
%comparing the atomic bias instances and 
%versus 
%intersectional bias instances exposed by \approach. 
%We found that interesectional bias is prevalent in LLMs, indeed more than atomic bias.  
%Moreover, we demonstrate 
%We found that a
%Secondly,  we show that 
%about one in six 
%In addition,  \checknumber{16.99\%} of intersectional bias-inducing mutants
%and up to 
%%one in ten (
%\checknumber{9.71\%} of unique original inputs are \textit{strictly} intersectional bias-inducing.  
Furthermore,  
%we demonstrate that \approach's dependency-based invariant ensures that 
%the generated inputs
%by \approach 
% test inputs that 
%are 
%similarly 
%as 
%grammatically valid,  it 
\approach's dependency invariant reduces the number of bias instances developers need to inspect by 
\checknumber{10X}.
% (conform to the parse tree of the original text).
% 
%\approach's invariant further 
%reduces 
% and 
%it 
%\revise{reduces false positives by \checknumber{90.71\%}. } 
%These intersectional bias-inducing mutants and original inputs do not trigger an atomic bias, i.e., their corresponding atomic mutants are benign when tested individually.
% that about one in five (\checknumber{22.20\%})
%intersectional 
%intersectional bias instances 
%involving intersectional individuals and groups 
%are concealed during atomic 
%individual and groub 
%bias testing,  and vice versa (\checknumber{18.14\%}).      
%Overall,  o
%
%In summary,  w
We hope that 
%\approach and our empirical evaluation 
this work will enable LLM practitioners
% (ML engineers and developers) 
to 
discover 
%and mitigate 
%mitigating 
%%test 
%LLMs for 
intersectional bias.
% the 
%using 
%the mutants  
%and distinct original cases 
%expposing bias 
%in our experiments.  
%}
%w.r.t.  individual and group fairness properties 
%, namely race, gender and body.   
%\approach employs input mutations to generate test inputs from the existing dataset and metamorphic oracles to detect bias based on the model outcomes.
%Our evaluation involves a total of 20 tested LLM models based on four 
%BERT-like 
%LLM architectures and five legal datasets. 
%, resulting in  
%EOur e
%Empirical findings show 
%We found that interesectional bias is prevalent in LLMs, indeed more than atomic bias.  
%Moreover, we demonstrate that 
%%intersectional 
%bias involving intersectional individuals and groups are concealed during atomic 
%%individual and groub 
%bias testing.  
%Overall,  o
%Our study motivates the need to specifically 
%tudy and 
%evaluate LLMs for intersectional bias.  
%We evaluate \approach 
%\revise{
Finally,  we provide our experimental data and \approach's implementation to support replication and reuse: %
\begin{center}
%\todo{
\url{https://github.com/BadrSouani/HInter} %}
\end{center}

\balance

\bibliography{bibliography}

\end{document}